\documentclass[a4paper,fleqn]{cas-dc}
\usepackage{preprint-format}

\usepackage{amsmath,amsfonts}
\usepackage{algorithmic}
\usepackage{algorithm}
\usepackage{array}
\usepackage[caption=false,font=footnotesize,labelfont=bf,textfont=normal]{subfig}
\usepackage{textcomp}
\usepackage{url}
\usepackage{verbatim}
\usepackage{graphicx}
\graphicspath{{figs/}{figures/}{pictures/}{images/}{./}}
\usepackage{xcolor}
\usepackage{booktabs}
\usepackage{tabu}
\usepackage{enumitem}
\usepackage{lineno}

\usepackage[numbers]{natbib}

\def\tsc#1{\csdef{#1}{\textsc{\lowercase{#1}}\xspace}}
\tsc{WGM}
\tsc{QE}

\begin{document}
\let\WriteBookmarks\relax
\def\floatpagepagefraction{1}
\def\textpagefraction{.001}

\shorttitle{A Lightweight Differentiable Method for Real-Time Skin Rendering and Digitization}

\shortauthors{Marquez-Balduque et al.}

\title [mode = title]{A Lightweight Differentiable Method for Real-Time Skin Rendering and Digitization}

\author[1]{Mario Marquez-Balduque}
\author[2,3,4]{Carlos Rodr\'iguez-Pardo}
\author[1]{Jorge Lopez-Moreno}
\affiliation[1]{organization={Universidad Rey Juan Carlos}, country={Spain}}
\affiliation[2]{organization={Politecnico di Milano}, city={Milan}, country={Italy}}
\affiliation[3]{organization={Euro-Mediterranean Center on Climate Change (CMCC)}, city={Lecce}, country={Italy}}
\affiliation[4]{organization={RFF-CMCC European Institute on Economics and the Environment (EIEE)}, city={Milan}, country={Italy}}

\begin{abstract}
Inverse rendering for skin typically relies on offline path tracing to estimate physical properties, which fail to map directly to the approximate shaders utilized in interactive applications, such as game engines. We bridge this gap by directly estimating subsurface scattering parameters tailored for real-time rendering pipelines. We introduce a differentiable Screen-Space Subsurface Scattering (SSSS) framework that enables the joint optimization of spatially varying base textures and scattering distances. To account for local variations in skin thickness, we estimate these parameters across distinct facial regions, generating a blended profile map ready for production runtime engines. Our pipeline utilizes a low-cost, single-light, single-view capture setup designed for accessible and practical appearance acquisition. Optimization is driven by a masked loss function targeting regions where subsurface scattering effects are dominant. We validate our approach on a real-world dataset of $45$ individuals across the Monk Skin Tone Scale, ensuring demographic generalization. Furthermore, we complement this evaluation with synthetic data, a multi-light setup validating our single-light assumption, and a tilted-view configuration confirming angular consistency. Results demonstrate that our method effectively recovers scattering parameters from lightweight inputs, yielding realistic skin appearance directly compatible with standard real-time shaders.
\end{abstract}





\maketitle

\section{Introduction}
Accurately simulating human skin remains a significant challenge in computer graphics due to its physical complexity and the high sensitivity of the human visual system to facial appearance. As a layered, anisotropic, and spatially varying medium, skin involves intricate subsurface scattering, fine-scale geometric details, and directionally dependent light transport. To meet the computational constraints of interactive applications, the industry often relies on screen-space approximations rather than full volumetric path tracing. However, the perceptual sensitivity of observers ensures that even subtle rendering inaccuracies appear unnatural. Additionally, the global diversity of skin tones demands methods that generalize across the full spectrum of human pigmentation while maintaining both physical plausibility and real-time performance.

Traditional approaches for capturing skin optical properties often require sophisticated hardware, such as hyperspectral sensors or polarimetric imaging systems \cite{Ha2024PolarimetricFaces, Aliaga2023AProperties, Li2024PracticalReflectance}. While these methods yield accurate results, they are typically expensive, restricted to controlled laboratory settings, and produce physical parameters (e.g., extinction coefficients) that do not map directly to real-time shaders. Furthermore, they often struggle to scale across the diverse facial expressions and skin tones encountered in real-world scenarios. To address these limitations, we introduce a low-cost, replicable single-view, single-light capture setup. Our compact configuration facilitates the practical acquisition of skin appearance without requiring specialized environments.

\begin{figure}
    \centering
    \includegraphics[width=\columnwidth]{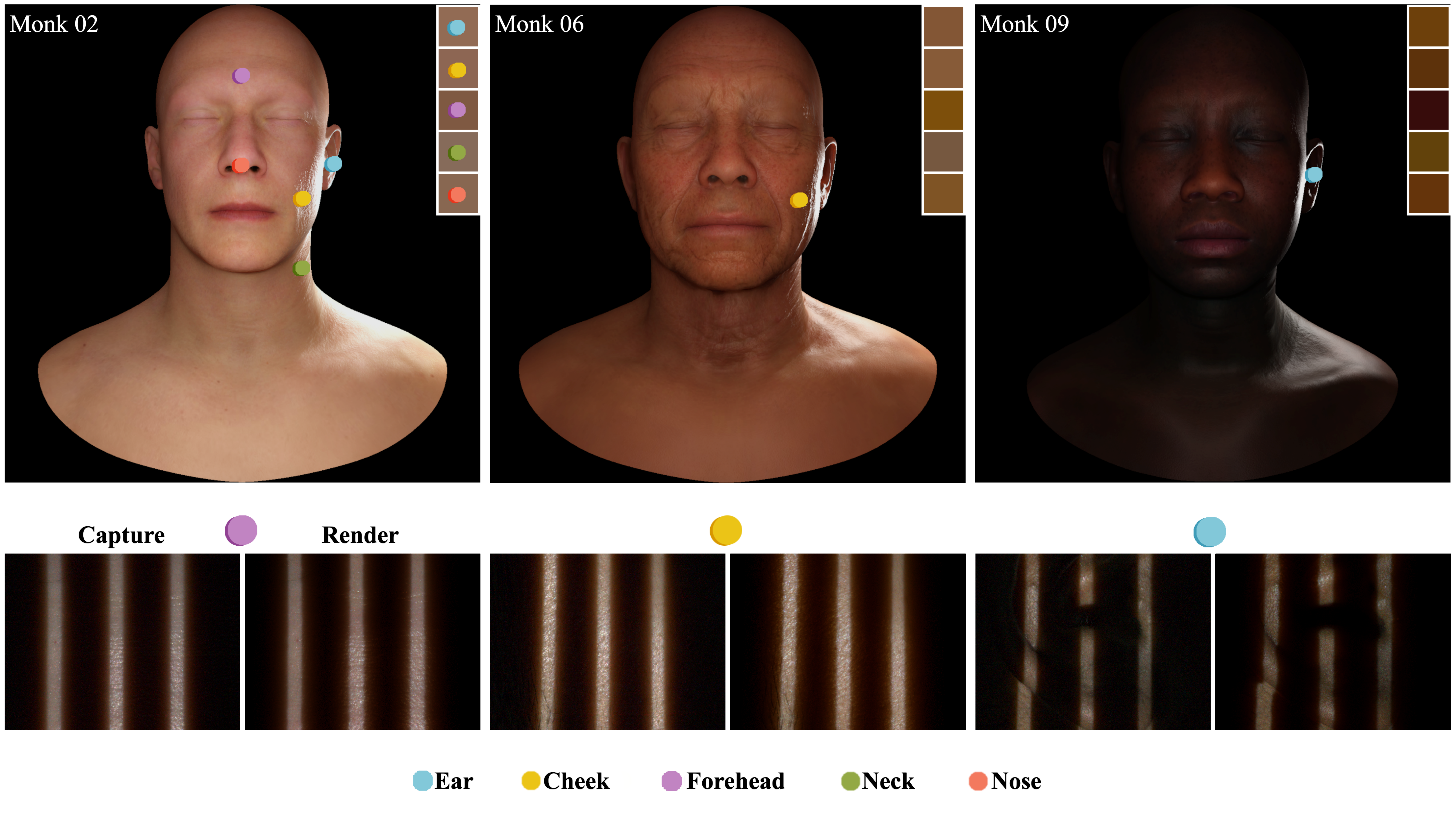}
    \caption{Our lightweight pipeline estimates spatially-varying subsurface scattering from single-image strip-light captures. Shown: rendered avatars across the Monk skin tone scale using optimized per-region scattering parameters.}
    \label{fig:teaser} 
    \vspace{-2em}
\end{figure}

Instead of estimating offline physical parameters like absorption or extinction coefficients through complex path-tracing intermediaries, we directly optimize the specific parameters used in real-time rendering pipelines. We ground our approach in the Bidirectional Subsurface Scattering Reflectance Distribution Function (BSSRDF), specifically utilizing the Burley Normalized Diffusion profile \cite{Burley2015ExtendingScattering}, the standard for real-time physically based skin shading. To enable optimization, we implement a differentiable version of the Screen-Space Subsurface Scattering (SSSS) algorithm incorporating this profile. This model provides control via the scattering distance parameter, which governs the distance light travels beneath the surface. Tuning this parameter manually is difficult due to its strong coupling with material appearance and the lack of direct visual references. Our approach automates this process by directly optimizing scattering distance textures from photographic data, enabling efficient and plausible appearance reconstruction.

\begin{figure}
    \centering
    \includegraphics[width=0.9\columnwidth]{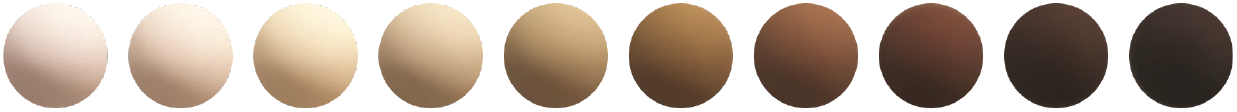}
    \caption{Monk Skin Tone Scale color swatches \cite{Monk2023TheScale}. Image credit: © Google, used under the CC BY 4.0 license.}
    \label{fig:monk}
    \vspace{-2em}
\end{figure}

Prior work frequently underexplores skin tone diversity, focusing primarily on mid-range tones and resulting in biased models that generalize poorly across lighter or darker skin. To address this limitation and drive representative skin modeling, we anchor our framework in the Monk Skin Tone Scale~\cite{Monk2023TheScale} (Fig.~\ref{fig:monk}). Unlike traditional scales designed primarily for dermatological UV response (e.g., Fitzpatrick~\cite{Fitzpatrick1988TheVI}), the Monk scale provides a perceptually balanced distribution covering the broad spectrum of human skin tones. Our primary goal is to leverage this scale to deliver a lightweight differentiable inverse rendering method that is physically grounded, equitable across diverse populations, and directly compatible with real-time production pipelines.

\noindent\textbf{Our key contributions are:}
\begin{enumerate} 
    \item A lightweight, low-cost, single-light, single-view capture setup explicitly designed for the practical acquisition of facial skin patches across diverse skin tones. 
    \item A differentiable rendering framework introducing a novel optimization stage to estimate state-of-the-art real-time scattering parameters directly from image data, bypassing complex path-tracing intermediaries.
    \item A masked loss formulation that isolates gradients within shadowed regions, where multiple scattering is physically dominant, ensuring accurate and stable optimization of subsurface parameters. 
    \item A large dataset of calibrated facial skin patches and their estimated optical properties, sampled across the Monk Skin Tone Scale to drive equitable and representative skin modeling.
\end{enumerate}

\section{Related Work} \label{sec:related_work}
In this section, we review prior work on skin rendering, categorizing approaches into subsurface scattering approximations, biophysical and material models, physical acquisition methods, and inverse rendering strategies.

\begin{figure}
    \centering
    \subfloat[]{%
        \includegraphics[width=.2\columnwidth]{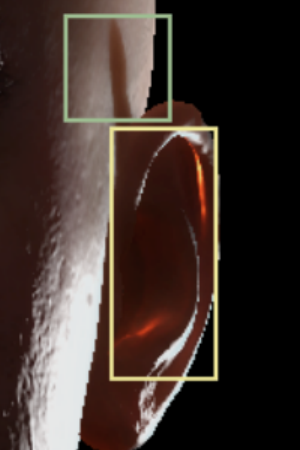}}
    \subfloat[]{%
        \includegraphics[width=.8\columnwidth]{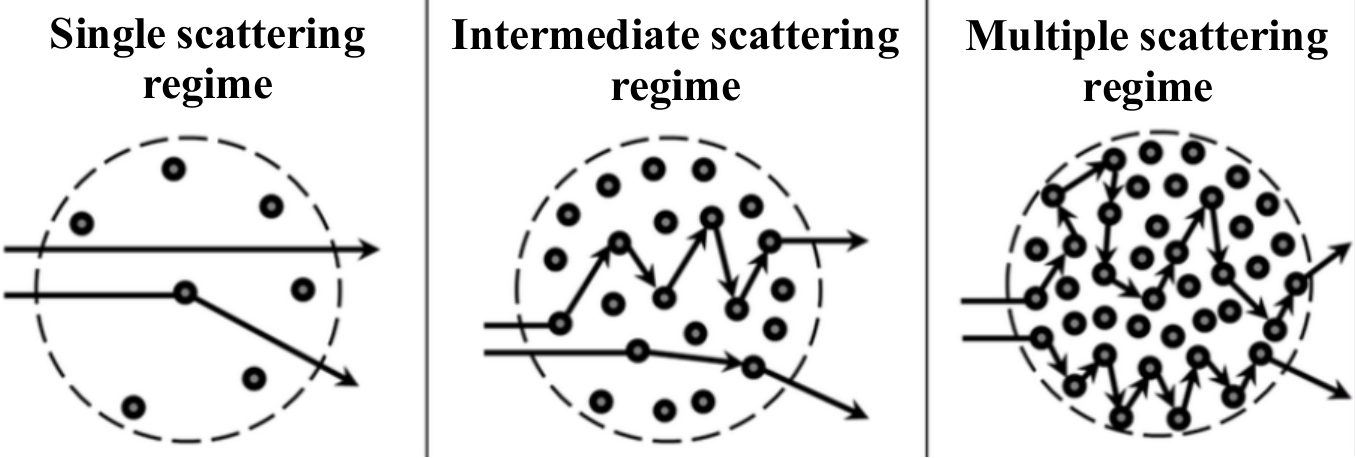}}
    \caption{Illustration of subsurface scattering regimes. Left: a rendered ear where the yellow box highlights a single scattering dominant area (thin cartilage), while the green box shows a region dominated by multiple scattering, emphasizing the variability of light transport in human skin. Right: representation of single, intermediate, and multiple scattering. Adapted from \cite{Mishra2018DropletTechniques}.}
    \label{fig:sss_comparison}
    \vspace{-2em}
\end{figure}

\textbf{Subsurface Scattering Approximations.} Realistic skin rendering requires simulating the complex transport of light through layers of varying opacity. As described by the BSSRDF \cite{Burley2015ExtendingScattering}, light transport in translucent media falls into three regimes: single scattering (thin materials), multiple scattering (thick materials), and intermediate scattering (turbid media) (Fig.~\ref{fig:sss_comparison}). While full simulation is computationally expensive, most methods approximate the multiple scattering component to achieve visual plausibility \cite{Che2020TowardsScattering}. The dipole model \cite{Jensen2001ATransport} remains a foundational approximation for offline rendering, modeling subsurface transport via virtual light source pairs. Subsequent improvements include the multipole method \cite{Donner2005LightMaterials} for multi-layered materials and the directional dipole \cite{dEon2012ADipole, Frisvad2014DirectionalScattering}, which introduces anisotropy to better capture detailed surface features. For higher precision, the quantized diffusion model \cite{DEon2011AMaterials} discretizes the diffusion process, offering finer control over scattering behavior.

For real-time pipelines, scattering is typically separated into surface and subsurface components. Translucency effects (single scattering) are often approximated using thickness maps generated via inverted-normal ambient occlusion \cite{Barre-Brisebois2011ApproximatingLook}. For multiple scattering, screen-space techniques are the standard. \cite{Jimenez2010Screen-SpaceScattering} and \cite{Jimenez2015SeparableScattering} approximate diffusion by convolving irradiance maps with weighted Gaussian kernels, while \cite{Christensen2015AnScattering} proposed an exponential diffusion profile matching Monte Carlo simulations. Recently, neural approaches such as \cite{TG2024NeuralRepresentation, TG2024NeuPreSS:Objects} have proposed encoding scattering profiles into lightweight networks, though screen-space convolution remains the industry standard for its predictability and direct integration with standard shading models.

\textbf{Biophysical and Skin Material Models.} Human skin exhibits significant spatial heterogeneity. Early measurement-based approaches \cite{Weyrich2005AEditing} demonstrated that fixed scattering parameters fail to capture the variation between regions like the forehead, cheeks, and ears. \cite{Ghosh2008PracticalReflectance} introduced layered models to account for epidermal and dermal variations, while \cite{Tariq2006EfficientRelighting} utilized region-based lookup tables, and \cite{Lanza2024PracticalCosmetics} extended layered models to simulate cosmetics. Beyond empirical layers, recent biophysical models explicitly simulate physiological processes, such as sun-induced erythema and tanning \cite{10.2312:pg.20241286}, or multi-layer heterogeneous volume transport \cite{wang2025biophysical}.

While volumetric biophysical models offer high biological fidelity, they require computationally demanding Monte Carlo solvers. In contrast, we utilize the Disney 2015 BSSRDF \cite{Burley2015ExtendingScattering} as our parametric foundation. While the earlier Disney BRDF \cite{Burley2012Physically-basedDisney}—widely referred to in the industry as Burley 12—provided a robust model for surface reflection, it lacked volumetric scattering controls. The 2015 extension, commonly known as Burley 15, introduces a physically derived scattering distance parameter that ensures energy conservation while supporting anisotropic diffusion. This balance of physical plausibility and artist-friendly control makes Burley 15 ideal for our optimization pipeline. Inspired by \cite{Yatagawa2020DataBlending}, we optimize a scalar scattering distance per facial region under this model and blend them at runtime to achieve anatomical fidelity without sacrificing real-time efficiency.

\textbf{Skin Appearance Acquisition.} High-fidelity skin capture traditionally relies on complex hardware to separate reflectance components. Systems such as Light Stages \cite{Debevec2000AcquiringFace}, time-multiplexed lighting rigs \cite{Ghosh2011MultiviewIllumination}, and polarization-based setups \cite{Ma2007RapidIllumination., Fyffe2017Multi-ViewTopology} deliver ground-truth quality. Recent advances include spectral reconstruction methods \cite{Gitlina2020PracticalReflectance} and single-shot polarized capture systems \cite{Riviere2020Single-shotCapture} that recover high-quality geometry and appearance. However, these methods require specialized equipment and controlled environments. In contrast, we focus on a lightweight acquisition regime to democratize skin reconstruction. Our differentiable framework optimizes scattering parameters from a single-view, single-light setup, eliminating the need for costly optical components or synchronized lighting arrays. Furthermore, this simplified pipeline remains robust across the full spectrum of human skin tones. This combination of hardware accessibility and tonal generalization provides a scalable solution for realistic digital human creation.

\textbf{Inverse Rendering for Material Estimation.} Material estimation generally falls into learning-based or optimization-based categories. Learning-based methods \cite{Deschaintre2018Single-imageNetwork, Zhou2023APrior, Rodriguez-Pardo2025Single-imageScanner} train neural networks to infer parameters from images, offering fast inference but often lacking generalization to out-of-distribution examples. Optimization-based methods minimize the reconstruction error between observed and rendered images. For example, \cite{Garces2023TowardsSystem} optimized spatially varying parameters for fabrics using a differentiable rasterizer. In the domain of skin, \cite{Aliaga2023AProperties} and \cite{Li2024PracticalReflectance} have successfully estimated parameters from hyperspectral data. Most closely related to our work is \cite{Weier2025PracticalAppearance}, which performs inverse rendering for textured translucent appearance. However, while \cite{Weier2025PracticalAppearance} focuses on general translucent objects, our pipeline is specialized for human skin, incorporating domain-specific constraints (such as the Monk scale and facial region masks) and directly optimizing the screen-space parameters used in game engines. Furthermore, while \cite{wang2025biophysical} estimates biological parameters from a pre-extracted albedo map for offline volume rendering, our framework optimizes directly from raw photographs under directional lighting, decoupling surface reflection (albedo, roughness, specular) and subsurface scattering for real-time screen-space execution.

\section{Overview and Rendering Model}
\label{sec:overview_rendering}
This section outlines our proposed pipeline for skin appearance estimation and describes the real-time rendering techniques integrated into our engine to simulate subsurface scattering.

\subsection{Overview}
Our approach builds upon the diffusion-based skin model introduced by \cite{Christensen2015AnScattering}, which extends Burley 12 with a subsurface scattering parameter $s$. We adopt Burley 15 model for its physical grounding and proven photorealism \cite{Garces2023TowardsSystem}. Unlike prior works that fix subsurface parameters to fit surface properties, we explicitly estimate the subsurface scattering terms that dominate skin appearance. We define the scattering distance as $d = \frac{1}{s}$ (assuming a unit mean free path). We replace the standard uniform parameterization with an optimized homogeneous scattering parameter per facial region, which are then blended to create a spatially varying map for local anatomical variations.

We focus on the human face due to its perceptual importance in recognition \cite{Bruce1986UnderstandingRecognition} and the significant variation in skin thickness that influences light diffusion \cite{Jacques2013OpticalReview}. Consequently, our acquisition targets facial skin patches. 

\begin{figure*}
    \centering
    \includegraphics[width=\linewidth]{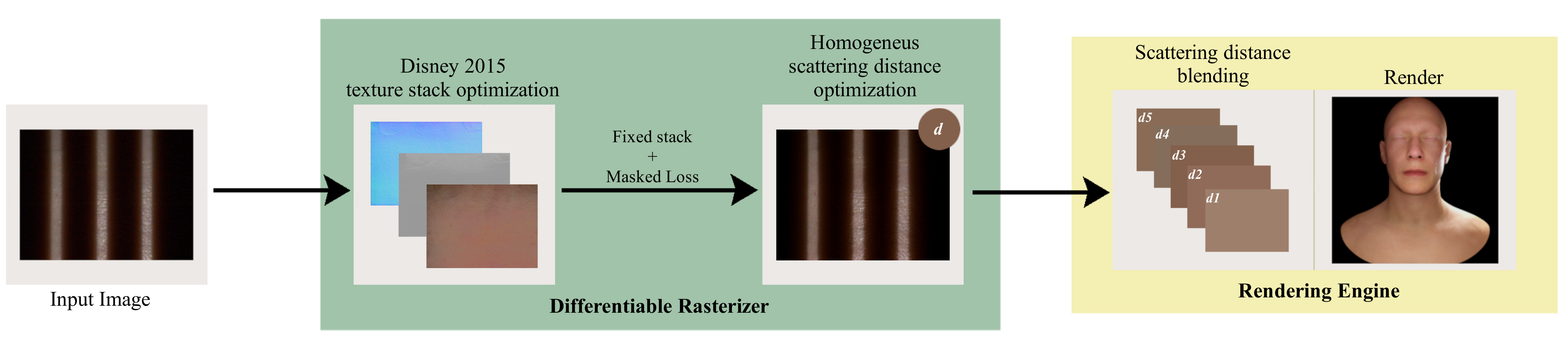}
    \caption{Overview of our inverse skin rendering pipeline: we optimize Burley 15 texture parameters and a homogeneous scattering distance using a differentiable rasterizer, then generate spatially varying subsurface profiles via profile blending for final rendering.}
    \label{fig:pipeline}
    \vspace{-2em}
\end{figure*}

We utilize a single-view, single-light capture setup to extract skin data. To ensure representational diversity, we select subjects spanning the Monk Skin Tone Scale \cite{Monk2023TheScale}, capturing the optical variability caused by differing melanin and hemoglobin levels. Our proposed architecture proceeds in two sequential phases: first, we optimize a homogeneous scattering distance $d$ for each individual skin patch; second, we generate a full facial scattering map by blending these values based on anatomical regions, similar to the profile blending in \cite{Yatagawa2020DataBlending}. To execute this initial patch-fitting phase, our inverse rendering pipeline directly optimizes a spatially varying Burley 15 texture stack by backpropagating reconstruction errors through a differentiable rasterizer, extending the core optimization framework from \cite{Garces2023TowardsSystem}. The complete pipeline is illustrated in Fig.~\ref{fig:pipeline}. 

This automated approach eliminates manual parameter tuning, driving skin rendering directly from photographic observations. Accordingly, the remainder of the paper details our acquisition protocol (Section~\ref{subsec:acquisition}), the tensorized scattering architecture (Section~\ref{subsec:ssss_vectorized}), implementation details (Section~\ref{sec:implementation_details}), and our experimental results and validation (Sections~\ref{subsec:fitting_results} and~\ref{subsec:validation}).

\subsection{Rendering Model}

We implement a custom real-time rendering engine using WebGL and Babylon.js \cite{Moreau-Mathis2016Babylon.Essentials}. This choice provides a broadly deployable, cross-platform verification environment that runs across a wide range of browsers and devices, including resource-constrained and low-end platforms, allowing us to assess whether the fitted scattering distances produce realistic and physically consistent appearances under practical cross-device rendering constraints. In addition to visualizing the final assets, the engine acts as a synthetic data generator, producing reference images with known ground-truth parameters to validate our optimization procedure.

The shading model relies on two primary inputs: a thickness texture and a spatially varying scattering distance texture. In the absence of accurate data textures (measured or artist-painted), local thickness is approximated automatically via inverted-normal SSAO \cite{Barre-Brisebois2011ApproximatingLook}. The idea is to flip the surface normal, so that the AO hemisphere points into the object instead. Nearby surfaces encountered by those inward-facing samples then provide a rough indication of how close other parts of the object are. When inward-facing polygons are close to one another, they generate more occlusion; this is interpreted as the region being locally thinner.

\begin{figure}
    \centering
    \subfloat[\label{subfig:thickness_tex}]{%
        \includegraphics[width=.3\columnwidth]{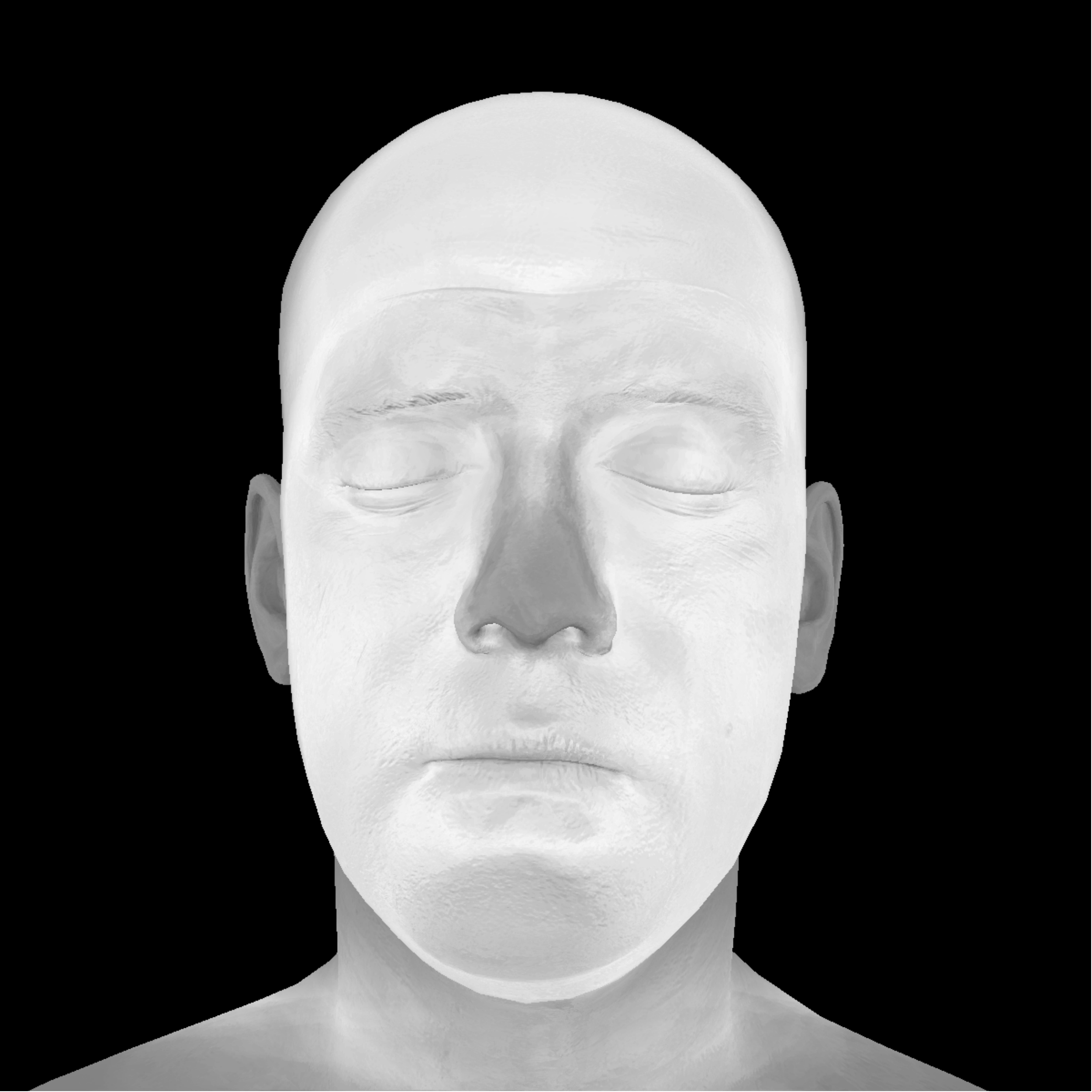}}
        \hspace{.02cm}
    \subfloat[\label{subfig:region_tex}]{%
        \includegraphics[width=.3\columnwidth]{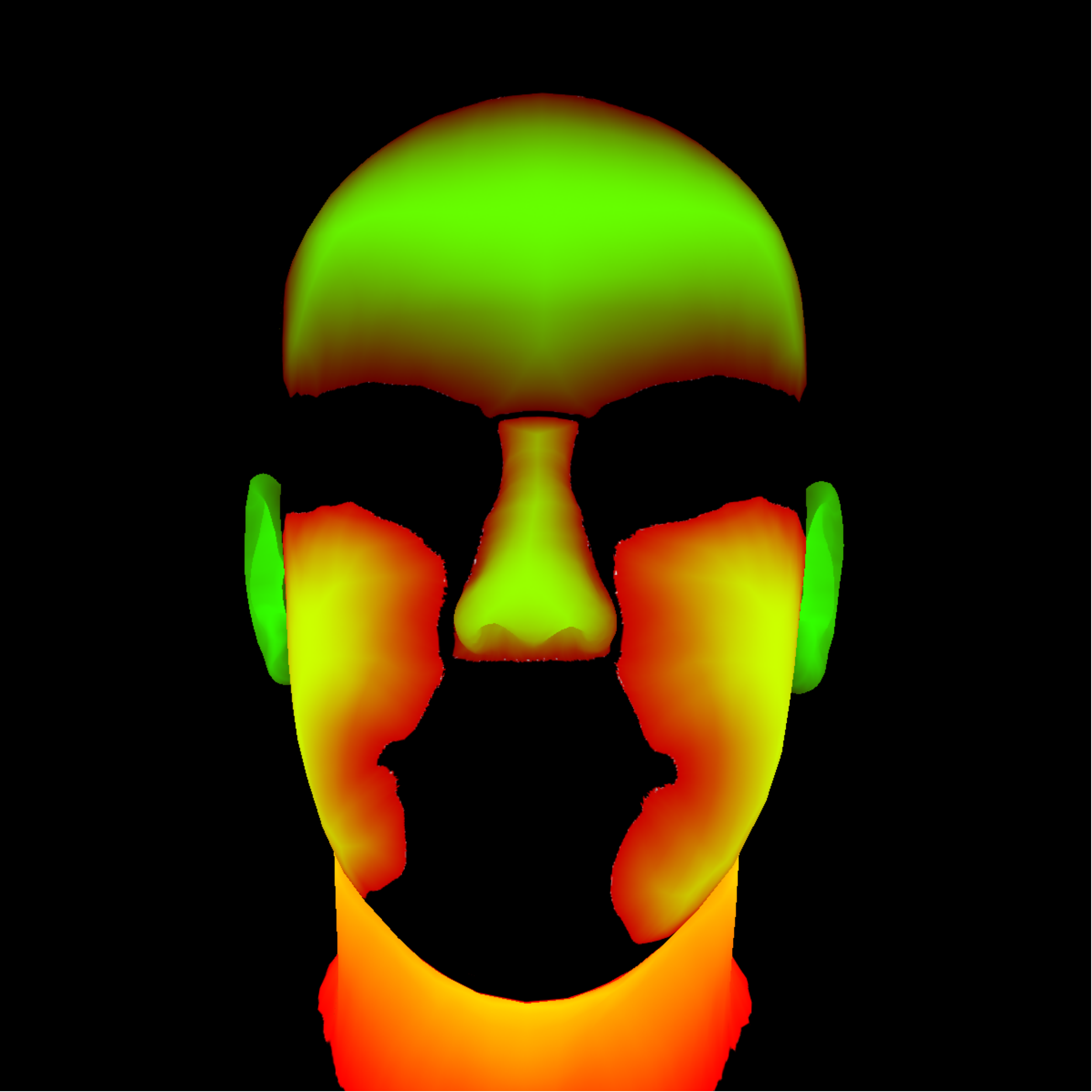}}
        \hspace{.02cm}
    \subfloat[\label{subfig:scattering_tex}]{%
        \includegraphics[width=.3\columnwidth]{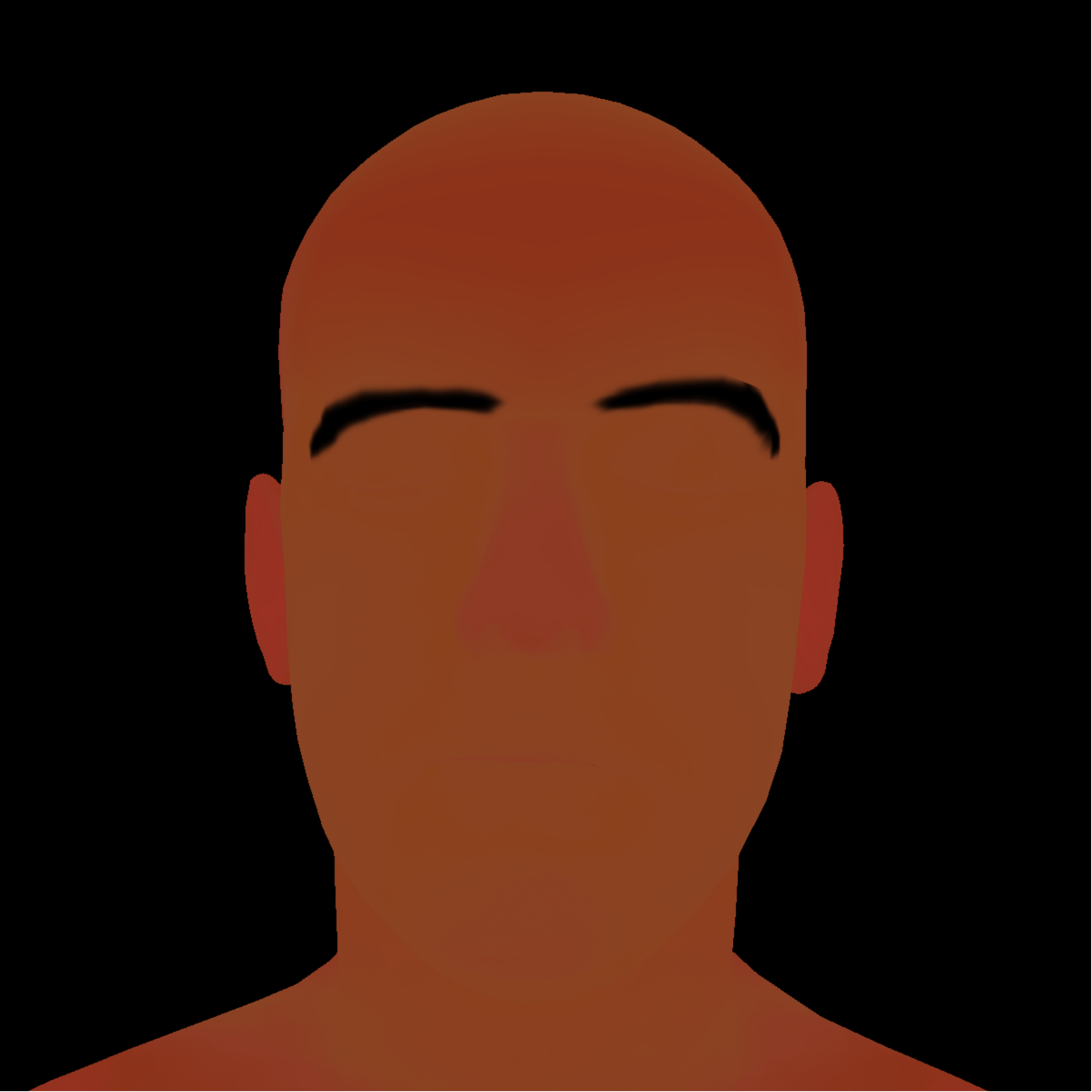}}
    \caption{Texture attachments of our rendering engine: (a) thickness, (b) region, and (c) blended scattering distance. For the region texture (b), the red channel encodes discrete region tags, while the green channel stores the interpolation factor for scattering distance blending .}
    \label{fig:tex_attach}
    \vspace{-2em}
\end{figure}

To encode local scattering behavior, we generate the scattering distance texture via profile blending. Homogeneous scattering values optimized from patches are interpolated using the thickness texture (Fig.~\ref{subfig:thickness_tex}) as a blending weight. This produces a smooth, continuous map (Fig.~\ref{subfig:scattering_tex}) that accounts for thinner skin areas (e.g., ears, nose) without requiring unique profiles per pixel. Optionally, an anatomical region map (Fig.~\ref{subfig:region_tex}) can be supplied to assign distinct scattering baselines to specific zones.

\textbf{Scattering Distance Scaling.}
The scattering distance texture is parameterized in metric units rather than arbitrary scalar weights. Following the convention of real-time engines such as Unreal Engine 5, the normalized texture domain $[0, 1]$ maps linearly to a scattering distance range of $[0, 1.0]\,\text{cm}$. To align this with our engine's world-space coordinate system (measured in meters), we apply a fixed dimensional conversion factor of $0.01$ ($1.0\,\text{cm} = 0.01\,\text{m}$). Unlike screen-space approximations that rely on projected pixel distances, we compute the true 3D Euclidean distance between the surface point and the sample point using the depth buffer. This ensures that the diffusion profile remains physically consistent regardless of camera distance or field of view.

\textbf{Multiple Scattering.}
We simulate multiple scattering using the screen-space approach proposed by \cite{Golubev2018EfficientReal-time}, which integrates the diffusion profile by distributing samples within a 2D screen-space disk and reconstructing the 3D surface geometry under perspective projection. We employ the exponential approximation of the Burley profile:
\begin{equation}
    R(r, d) = A \ \frac{e^{-r/d} + e^{-r/3d}}{8 \pi d r}
    \label{eq:burley_equation}
\end{equation}
where $A$ is the diffuse volume albedo, $r$ represents the 3D Euclidean distance, and $d$ is the scaled scattering distance inherent to the material properties. This formulation simplifies light transport through layered skin into a dual-exponential profile over a single homogeneous volume. Because it does not model individual epidermal and dermal layers, the recovered parameter $d$ represents an effective parameter for the real-time shader rather than an intrinsic scattering coefficient ($\sigma_s$).

To minimize Monte Carlo variance, these samples are generated in polar coordinates $(\theta_i, r_i)$ via an analytical Importance Sampling scheme. The resulting Probability Density Function (PDF) $p(r)$ and Complementary Cumulative Distribution Function (CCDF) $\bar{P}(r)$ derived from the diffusion profile are defined as:
\begin{equation} 
    p(r) = \frac{rR(r, d)}{A}, \quad \bar{P}(r) = \frac{1}{4}e^{-r/d}+\frac{3}{4}e^{-r/3d}
    \label{eq:pdf_ccdf_equation}
\end{equation}
According to \cite{Golubev2018EfficientReal-time}, $p(r)$ models the linear profile shape along a radial ray; the full $2\pi$ circle integration factor is explicitly accounted for during numerical CCDF inversion and applied within the final diffuse irradiance estimator ($2\pi/N$).

\begin{figure}
    \centering
    \subfloat[\label{subfig:sampling}]{%
        \includegraphics[width=.3\columnwidth]{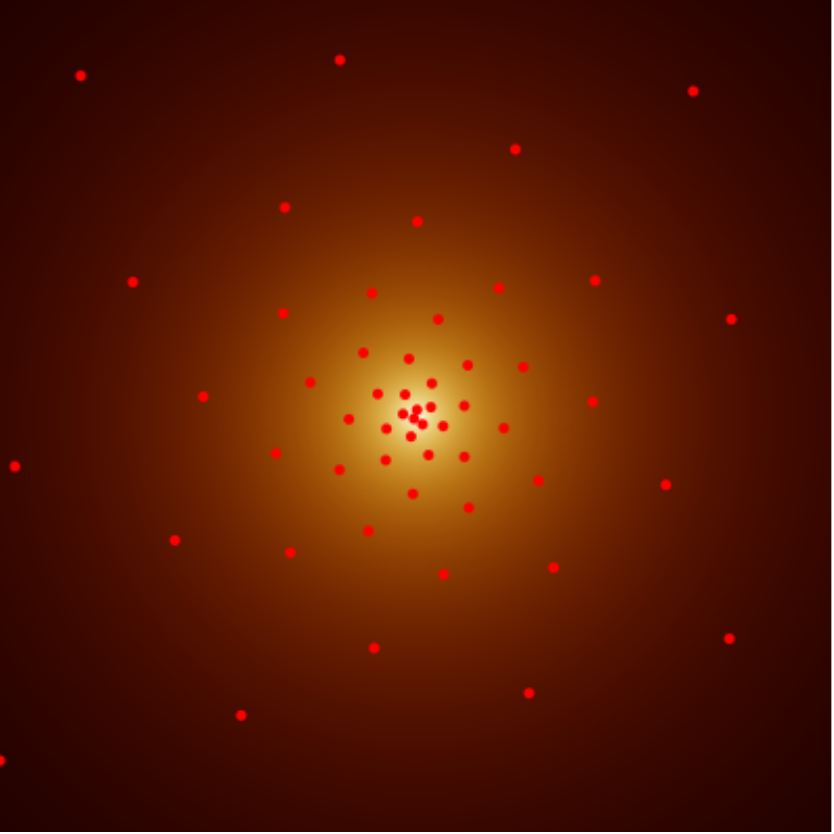}} 
        \hspace{.3cm}
    \subfloat[\label{subfig:vertical_fringing}]{%
        \includegraphics[width=.3\columnwidth]{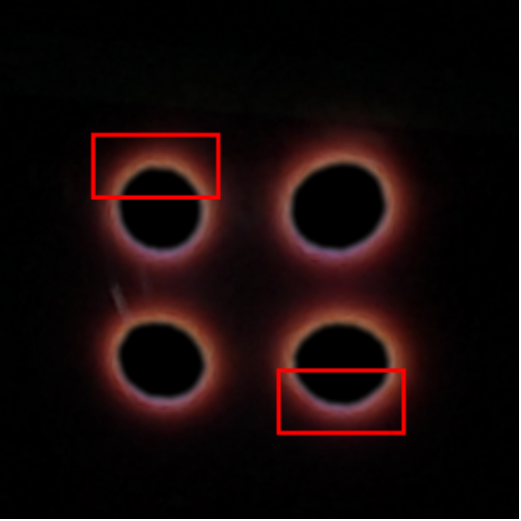}}
    \caption{Left: Fibonacci sampling pattern used for importance sampling the diffusion profile. Right: Chromatic fringing observed in captures due to spectral dependencies in the projector light.}
    \vspace{-2em}
\end{figure}

The sampling radius $r_i \in [0, r_{\text{max}}]$ is generated by inverting the CCDF in Eq.~\ref{eq:pdf_ccdf_equation} via inverse transform sampling. The angular coordinate $\theta_i$ follows a Fibonacci pattern~\cite{Hannay2004FibonacciSphere} (Fig.~\ref{subfig:sampling}) combined with a stochastic rotational jitter to convert structured aliasing into high-frequency temporal noise.

For a target fragment with view-space position $\mathbf{x} \in \mathbb{R}^3$, the local 2D polar offset is projected to determine its corresponding screen-space coordinate $\mathbf{u}_i = [u_i, v_i]^T$. The true 3D view-space position of this surface sample, $\mathbf{x}_i$, is reconstructed directly by applying the standard inverse perspective projection matrix to the normalized screen coordinates and the depth map $D(\mathbf{u}_i)$.

The true 3D Euclidean distance through the scattering medium is subsequently evaluated as $s_i = \|\mathbf{x} - \mathbf{x}_i\|_2$. Following the bilateral formulation in~\cite{Golubev2018EfficientReal-time}, we evaluate a spectral diffusion weight $k_{i, c}$ for each color channel $c \in \{R, G, B\}$ that balances the 3D radial profile response against the 2D sampling density:
\begin{equation}
    k_{i, c} = \frac{s_i \cdot R(s_i, d_c)}{r_i \cdot R(r_i, r_{\text{max}})}
    \label{eq:diffusion_weight}
\end{equation}
where $d_c$ denotes the material scattering distance for the respective spectral channel, and $r_{\text{max}}$ represents the global boundary domain limit.

The final scattered diffuse radiance $I_c(\mathbf{x})$ is obtained via a normalized weighted average of the incident diffuse irradiance field $L_c(\mathbf{u}_i)$ across all $N$ screen-space samples:
\begin{equation}
    I_c(\mathbf{x}) = A_c \left( \frac{\sum_{i=1}^{N} k_{i, c} L_c(\mathbf{u}_i)}{\sum_{i=1}^{N} k_{i, c}} \right)
    \label{eq:diffusion_equation}
\end{equation}
where $A_c$ is the channel albedo, and $L_c(\mathbf{u}_i)$ represents the standard Lambertian incident diffuse irradiance sampled at the projected screen-space coordinate $\mathbf{u}_i$. To ensure full reproducibility, the implementation details of the shader are provided in Section~5.3 of the Supplementary Material.

\begin{figure}
    \centering
    \subfloat[]{%
    \includegraphics[width=.3\columnwidth]{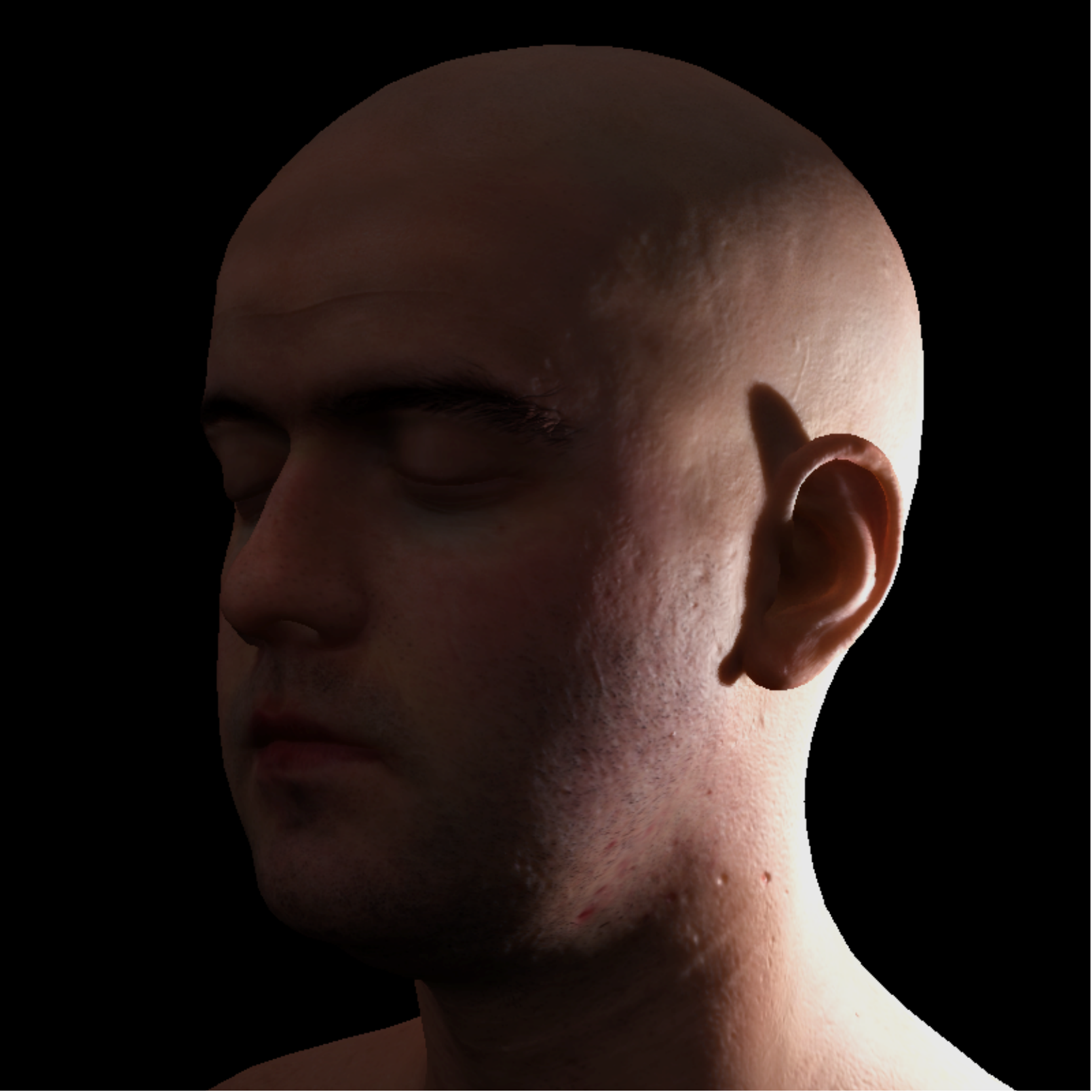}}
    \hspace{.02cm}
    \subfloat[]{%
    \includegraphics[width=.3\columnwidth]{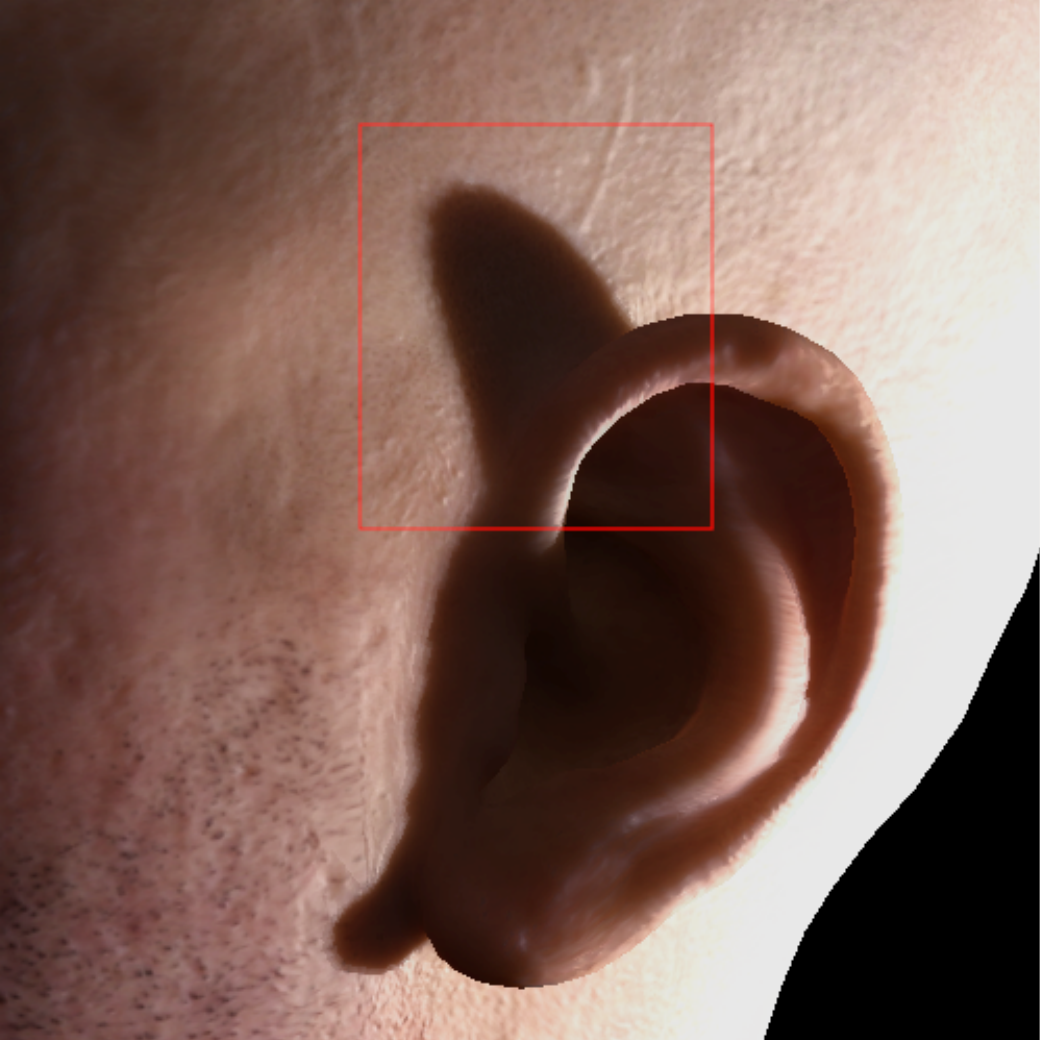}}
    \hspace{.02cm}
    \subfloat[]{%
    \includegraphics[width=.3\columnwidth]{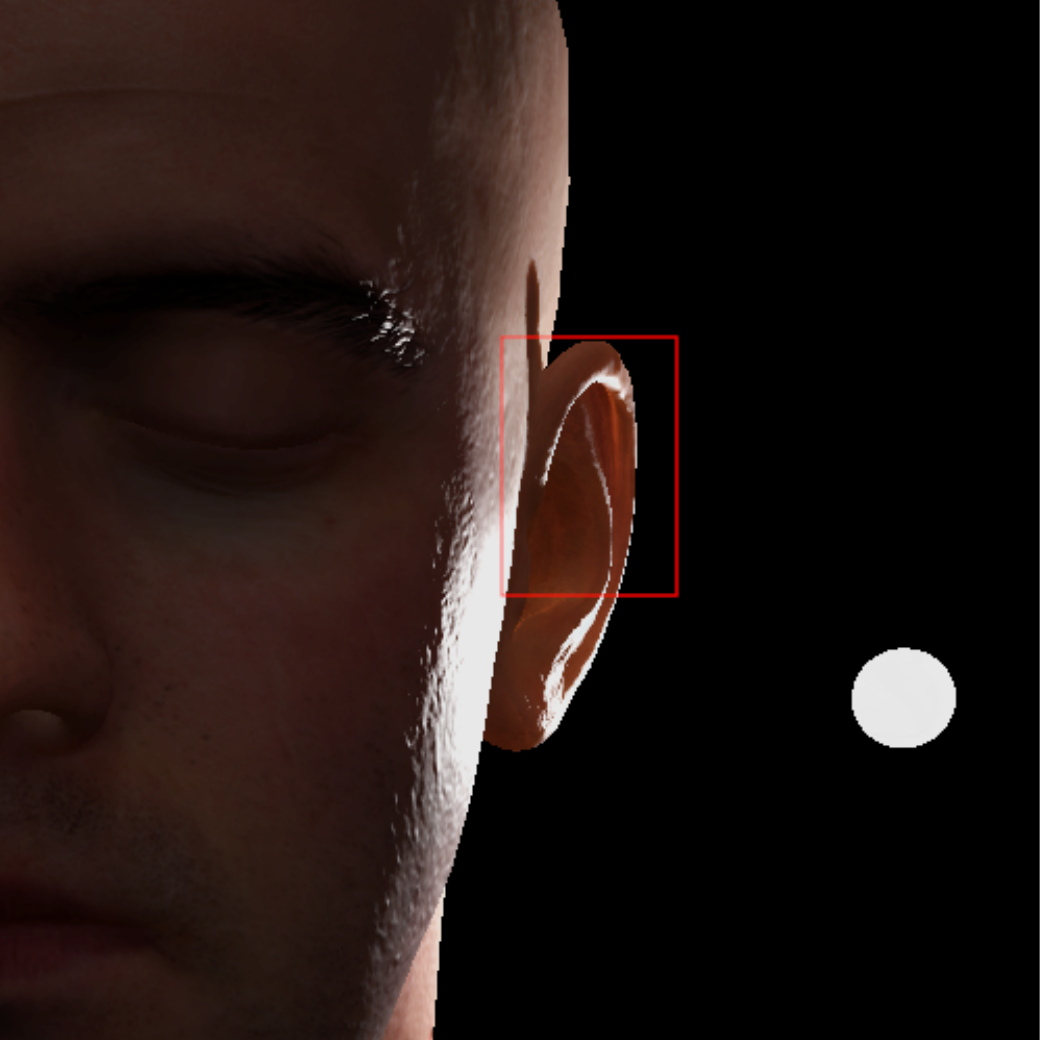}}
    \caption{Final render produced with our engine (a). Insets highlight regions dominated by (b) multiple scattering and (c) single scattering.}
    \label{fig:final_render}
    \vspace{-2em}
\end{figure}

\textbf{Single Scattering.}
For a complete skin shader, we also account for single scattering (translucency) through thin geometry (e.g., ears). We implement the approximation proposed by \cite{Barre-Brisebois2011ApproximatingLook}, which estimates light transmission using the Beer-Lambert law and the thickness texture. As our optimization framework focuses exclusively on the dominant multiple scattering component, we employ this term solely for visual completeness in the final renders (Fig.~\ref{fig:final_render}).

\section{Skin Fitting Pipeline}
\subsection{Acquisition}
\label{subsec:acquisition}

Estimating physically meaningful texture maps—such as normals, roughness, albedo, and scattering distances—from skin images typically demands complex acquisition systems. Common approaches rely on multi-view setups, polarized lighting, or hyperspectral imaging \cite{Aliaga2023AProperties, Li2024PracticalReflectance, Ha2024PolarimetricFaces}. While accurate, these systems suffer from limited scalability and high deployment costs. Furthermore, spatially varying estimation is highly sensitive to lighting directionality, local geometry, and microstructural detail \cite{Ghosh2008PracticalReflectance, Gotardo2018PracticalAcquisition, Ma2007RapidIllumination.}, making replication outside controlled laboratories difficult.

\begin{figure}
    \centering
    \subfloat[]{%
        \begin{minipage}{0.5\columnwidth}
            \centering
            \includegraphics[width=\columnwidth]{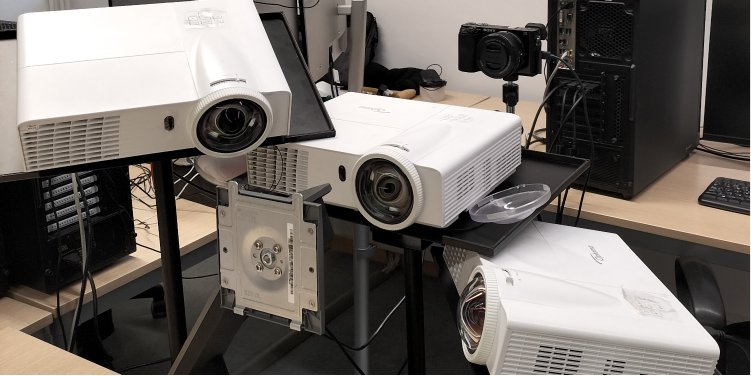}
        \end{minipage}
        \begin{minipage}{0.5\columnwidth}
            \centering
            \includegraphics[width=\columnwidth]{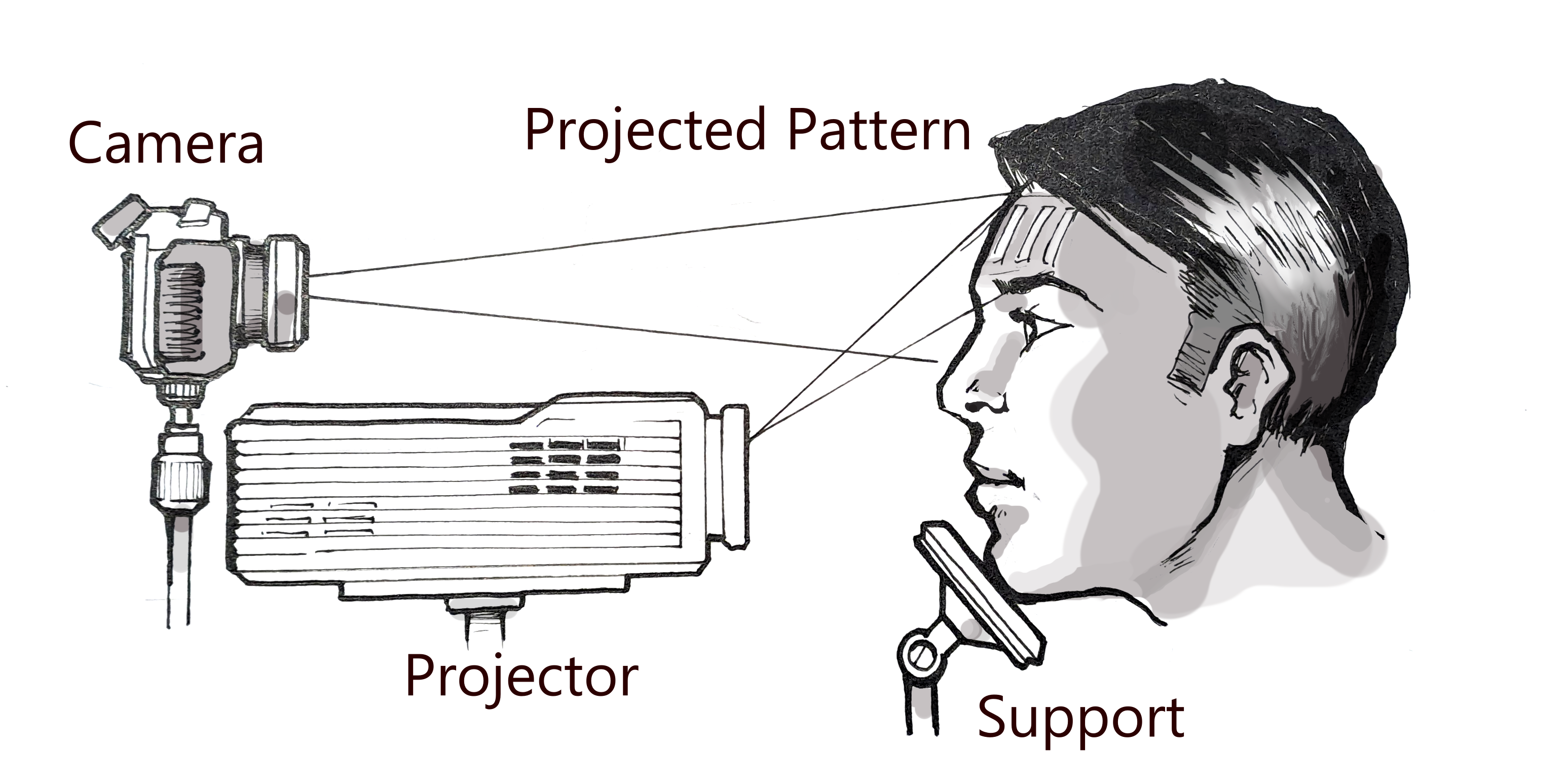}
        \end{minipage}
        \label{subfig:photos_setup}
    }
    
    \subfloat[]{%
        \begin{minipage}{\columnwidth}
            \centering
            \fcolorbox{gray}{lightgray}{\includegraphics[width=.2\columnwidth]{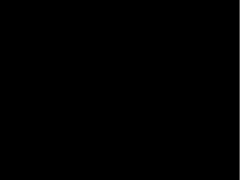}}%
            \fcolorbox{gray}{lightgray}{\includegraphics[width=.2\columnwidth]{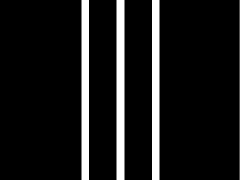}}%
            \fcolorbox{gray}{lightgray}{\includegraphics[width=.2\columnwidth]{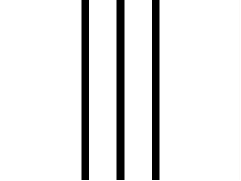}}%
            \fcolorbox{gray}{lightgray}{\includegraphics[width=.2\columnwidth]{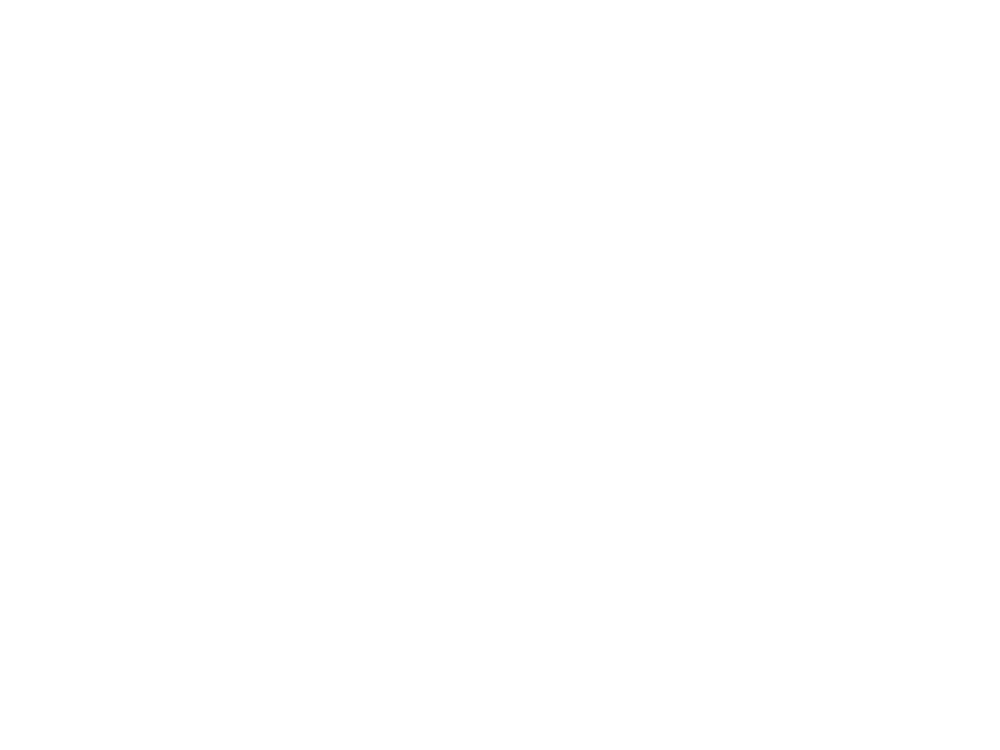}}
        \end{minipage}
        \label{subfig:light_patterns}
    }
    \caption{%
        Hardware acquisition setup and illumination patterns. (a) Physical configuration (left) and corresponding schematic ray-diagram (right) illustrating the scanning system geometry. (b) Temporal sequence of projected light patterns used during data acquisition: ambient baseline, structured stripes, inverted stripes, and uniform flat-field illumination.
    }
    \label{fig:acquisition_overview}
    \vspace{-1em}
\end{figure}

To address these limitations, we propose an acquisition setup that balances simplicity with fidelity and is lightweight in both hardware and operational complexity compared to such laboratory rigs. Rather than requiring custom optical components or electronic synchronization, our system relies entirely on consumer-grade devices, eliminates complex geometric and radiometric calibration protocols, and is easily deployed in standard indoor environments, specifically in a darkened room to eliminate ambient light interference. As shown in Fig.~\ref{subfig:photos_setup}, the setup consists of:

\begin{enumerate}[leftmargin=*]
    \item \textbf{Subject support.} A rigid horizontal platform stabilizes the target surface. Participants place a specific facial region (e.g., cheek or forehead) on a fixed marker to ensure positional repeatability.
    \item \textbf{Illumination source.} A projector is used to emit both high-contrast structured line patterns and uniform white light. It is positioned approximately 20\,cm from the subject, tilted upwards at 30° relative to the surface plane. This low-angle configuration creates strong directional lighting, maximizing the visibility of subsurface scattering effects while allowing for controlled binary illumination.
    \item \textbf{Imaging system.} A high-resolution RGB camera (DSLR) is mounted behind the projector at a distance of approximately 50\,cm from the subject. This near-coaxial alignment minimizes geometric distortion and allows the direct capture of reflectance and shading cues under controlled illumination.
    \end{enumerate}

Although we target a near-coaxial alignment, our pipeline ensures robustness against angular deviations. To maintain a lightweight workflow, we avoid complex projector-camera calibration and structured-light triangulation, relying instead on monocular depth estimation \cite{Yang2024DepthV2}. We map the network’s normalized output (0 for the near plane, 1 for the far plane) to a physical depth range of 2\,cm that bounds the local facial relief, anchored at a measured camera-to-subject distance of 47\,cm. All geometric and exposure settings are fixed to maintain consistency across subjects (Section~\ref{sec:implementation_details}).

Despite its simplicity, this setup successfully captures the subtle shading gradients and scattering cues required for material estimation, avoiding the cost and complexity of multi-camera or hyperspectral systems \cite{Ghosh2008PracticalReflectance, Aliaga2023AProperties}. While specialized hardware offers higher raw fidelity, we demonstrate that optimized differentiable rendering can extract production-quality assets from this minimal input.

\begin{figure}
    \centering
    \subfloat[]{%
        \includegraphics[width=.23\columnwidth]{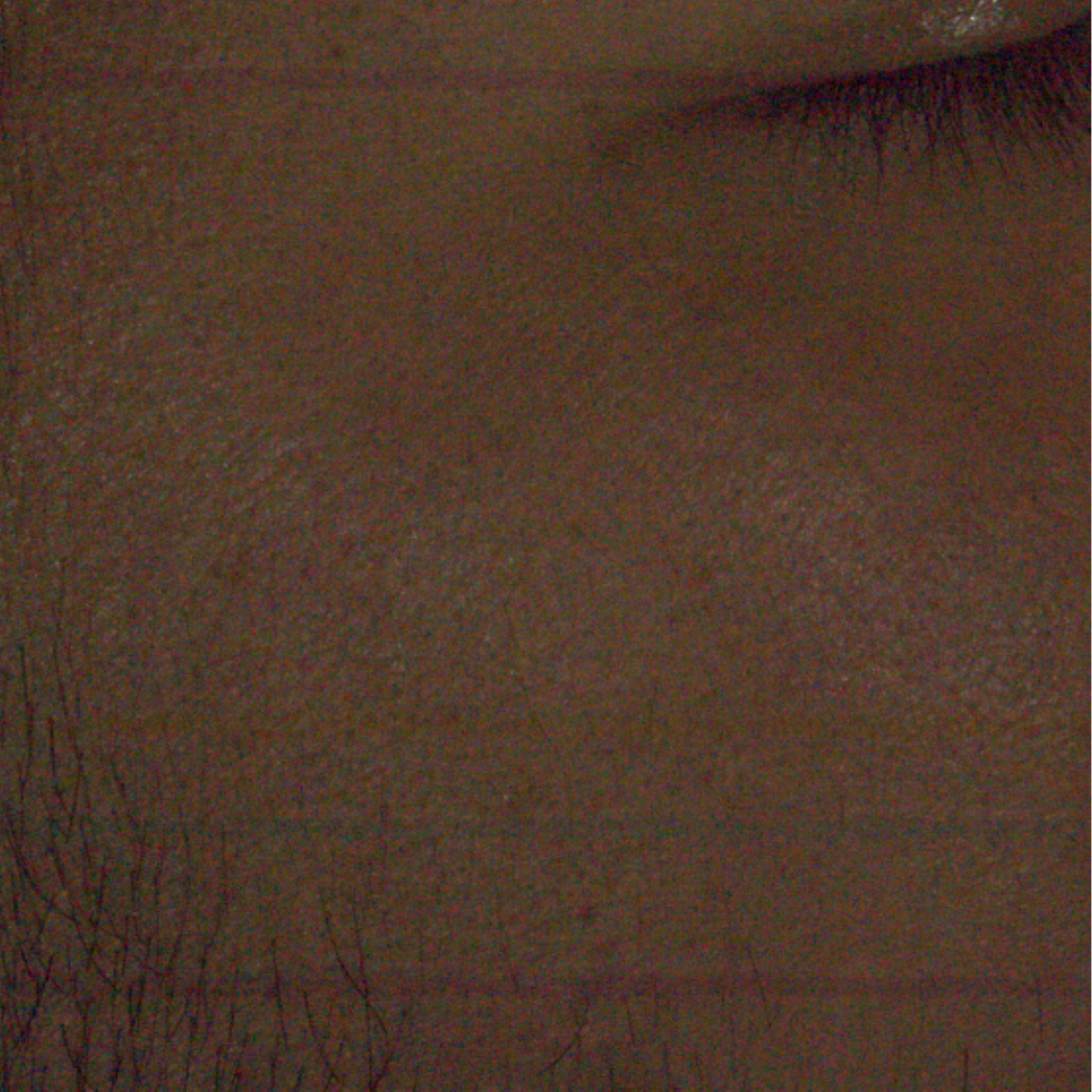}%
        \label{subfig:skin_photo_1}%
    }
    \hspace{.02cm}
    \subfloat[]{%
        \includegraphics[width=.23\columnwidth]{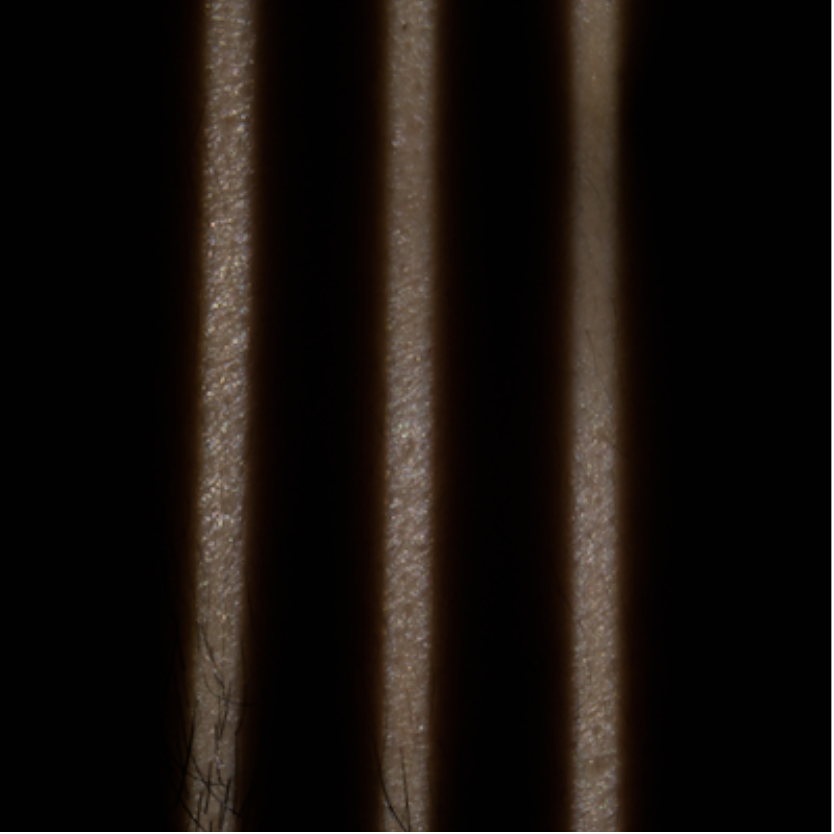}%
        \label{subfig:skin_photo_2}%
    }
    \hspace{.02cm}
    \subfloat[]{%
        \includegraphics[width=.23\columnwidth]{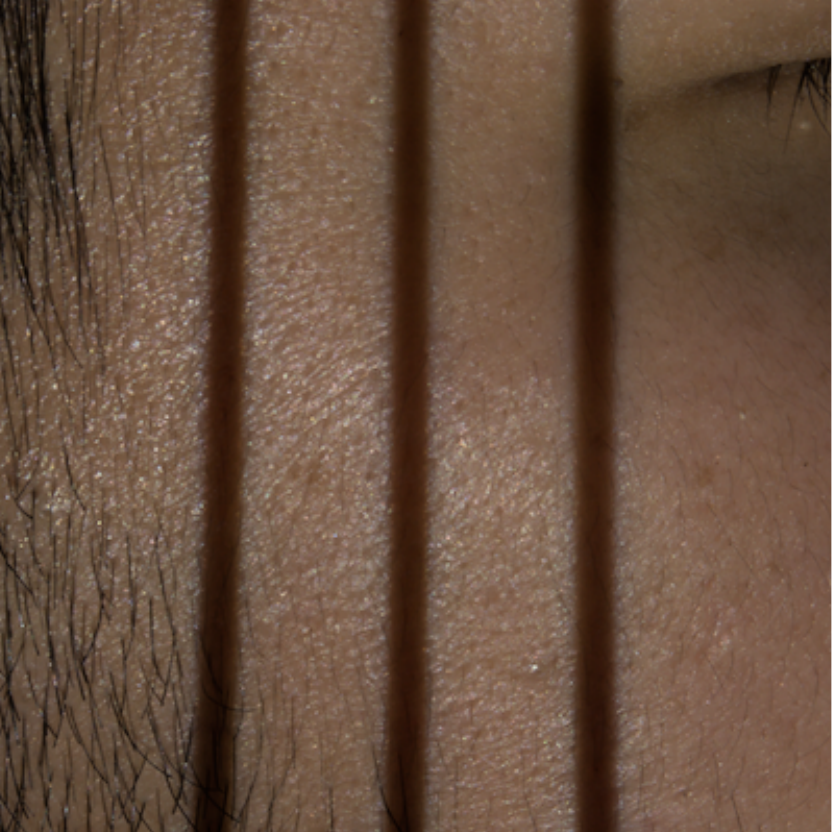}%
        \label{subfig:skin_photo_3}%
    }
    \hspace{.02cm}
    \subfloat[]{%
        \includegraphics[width=.23\columnwidth]{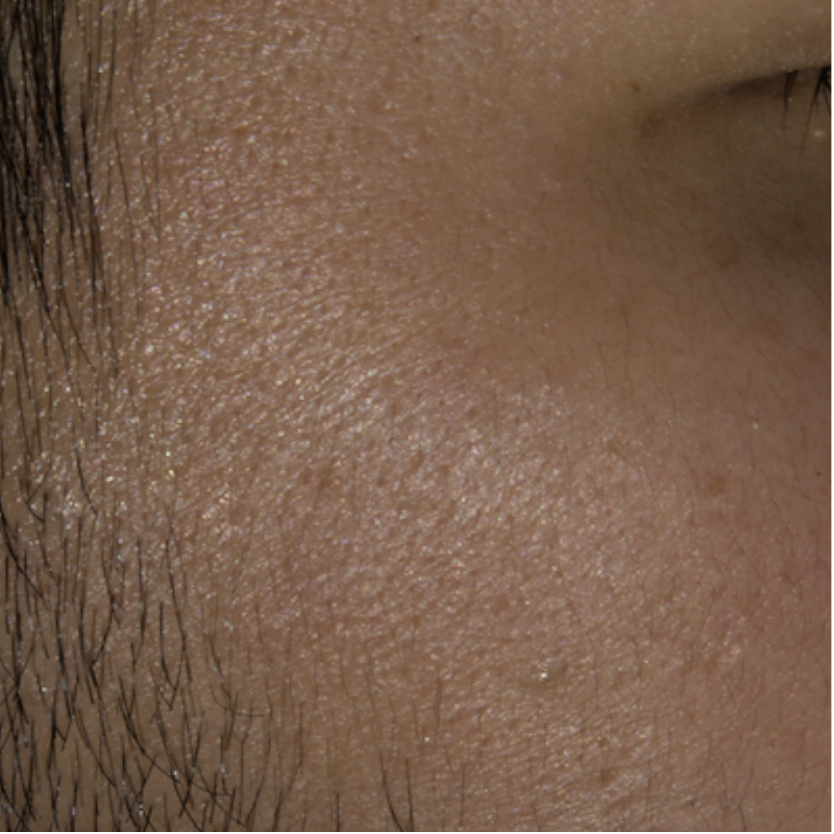}%
        \label{subfig:skin_photo_4}%
    }
    \caption{Localized view of the captured raw frame sequence during acquisition. (a) Ambient baseline frame (brightness-enhanced for visibility) showcasing residual environmental light. (b) Active surface capture under the high-frequency structured stripe pattern. (c) Corresponding active capture under the inverted stripe pattern. (d) Fully illuminated frame under uniform flat-field projection.}
    \label{fig:skin_photos}
    \vspace{-2em}
\end{figure}

We acquire four photographs per skin patch under distinct binary illumination patterns (Fig.~\ref{subfig:light_patterns}). We utilize vertically aligned line patterns to minimize projector-induced chromatic aberration and color fringing (Fig.~\ref{subfig:vertical_fringing}). Unlike the complex patterns conventional in structured-light reconstruction \cite{Woodham1980PhotometricImages, Salvi2004PatternSystems, Wu2011High-qualityIllumination}, our simplified linear patterns reduce spectral interference—a critical consideration given the wavelength-dependent nature of subsurface scattering. Fig.~\ref{fig:skin_photos} details the sequence:

\begin{itemize}[leftmargin=*]
    \item Photo 1 (\ref{subfig:skin_photo_1}): Captured with a black pattern to record ambient residuals. This image is subtracted from subsequent captures to isolate direct illumination.
    \item Photos 2 \& 3 (\ref{subfig:skin_photo_2}, \ref{subfig:skin_photo_3}): Captured using complementary vertical line patterns. These images highlight the skin’s directional response.
    \item Photo 4 (\ref{subfig:skin_photo_4}): Captured under full white illumination. This provides a uniformly lit reference for estimating global properties like albedo, decoupling shading from geometric variation.
\end{itemize}

\begin{figure}
    \centering
    \subfloat[]{%
        \begin{minipage}{.24\columnwidth}
            \centering
            \includegraphics[width=\linewidth]{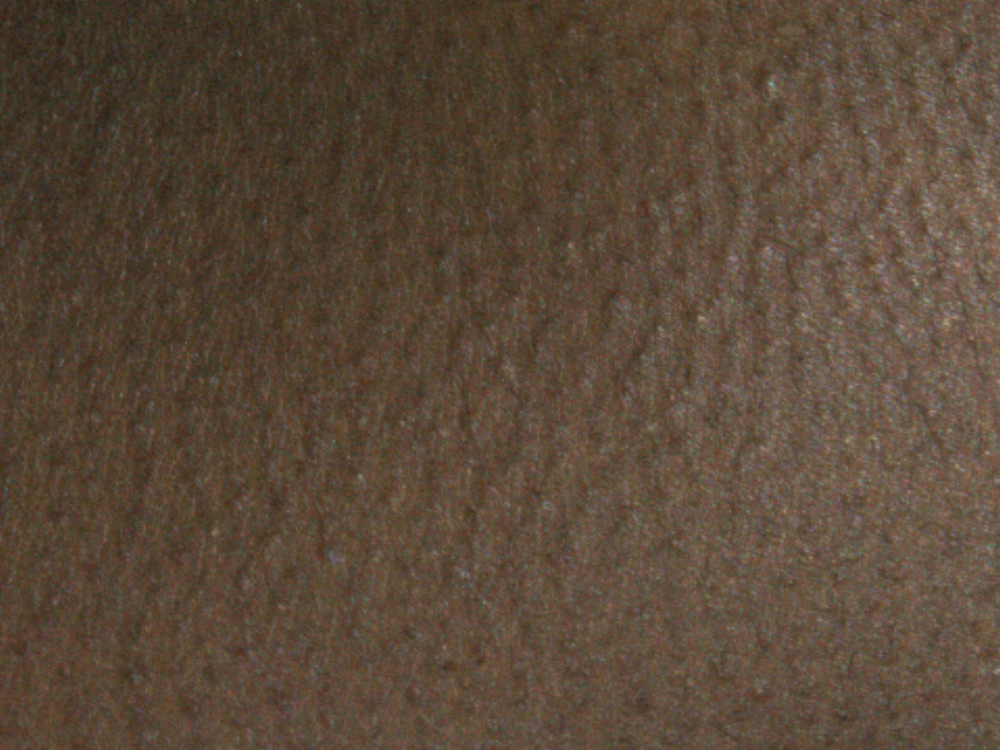}
            \includegraphics[width=\linewidth]{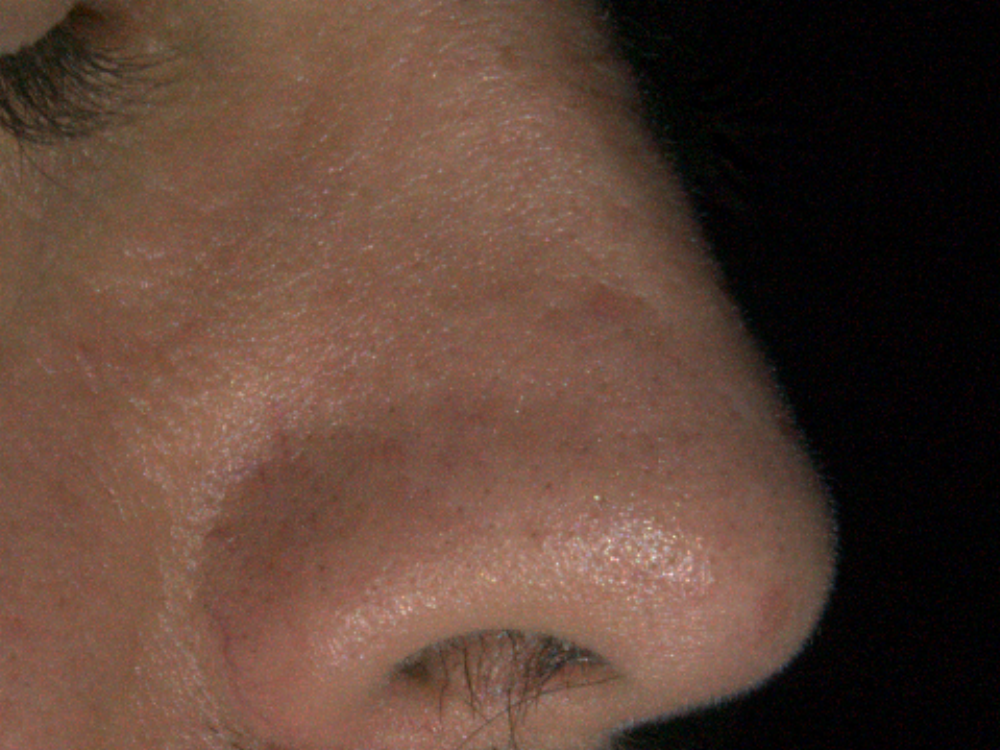}
            \includegraphics[width=\linewidth]{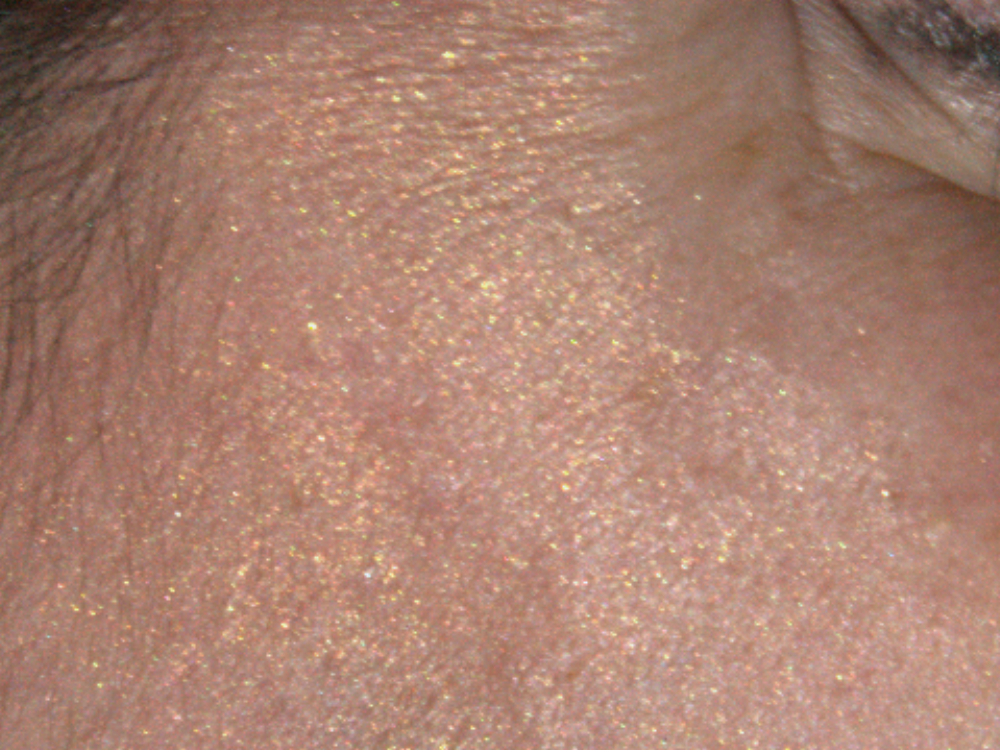}%
        \end{minipage}%
        \label{subfig:col_reference}%
    }\hfil
    \subfloat[]{%
        \begin{minipage}{.24\columnwidth}
            \centering
            \includegraphics[width=\linewidth]{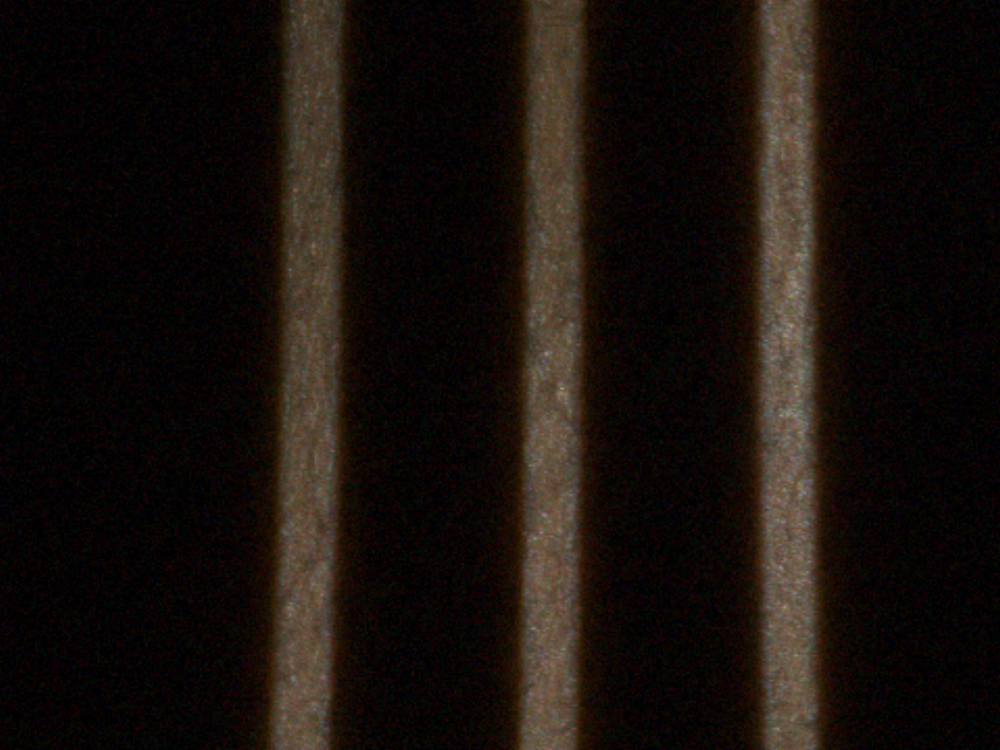}
            \includegraphics[width=\linewidth]{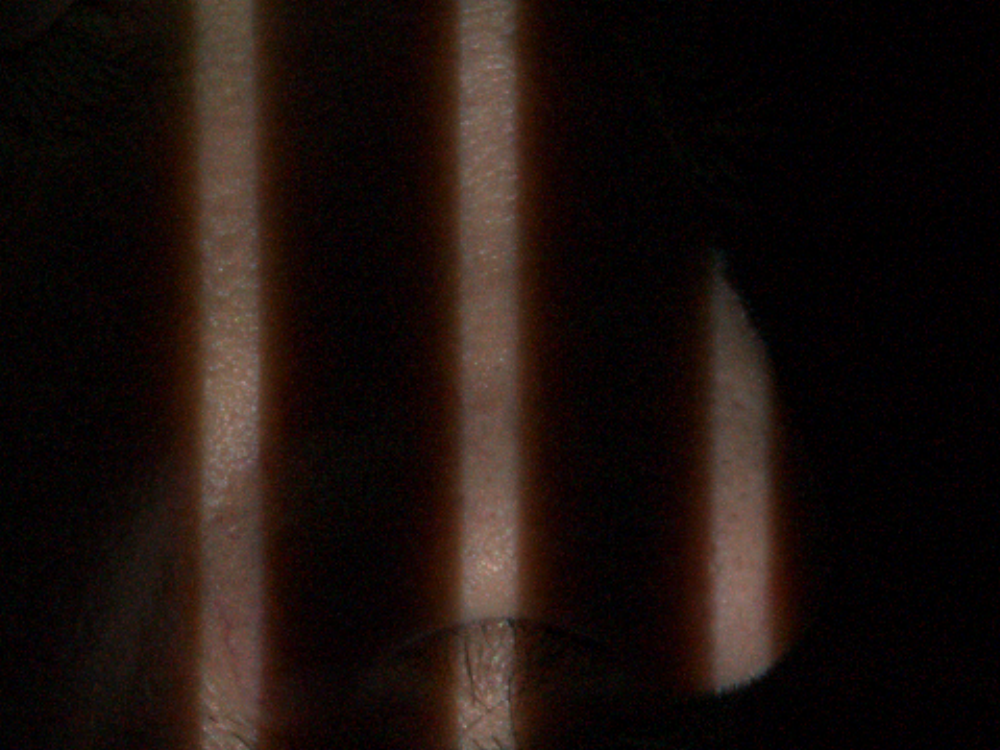}
            \includegraphics[width=\linewidth]{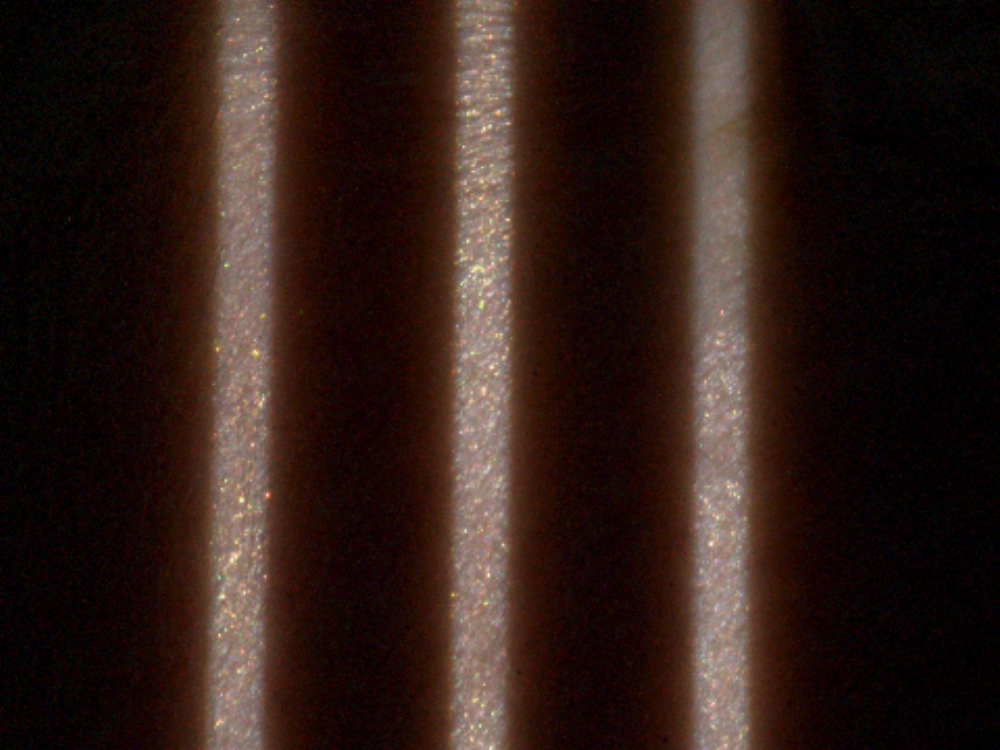}%
        \end{minipage}%
        \label{subfig:col_ssscapture}%
    }\hfil
    \subfloat[]{%
        \begin{minipage}{.24\columnwidth}
            \centering
            \includegraphics[width=\linewidth]{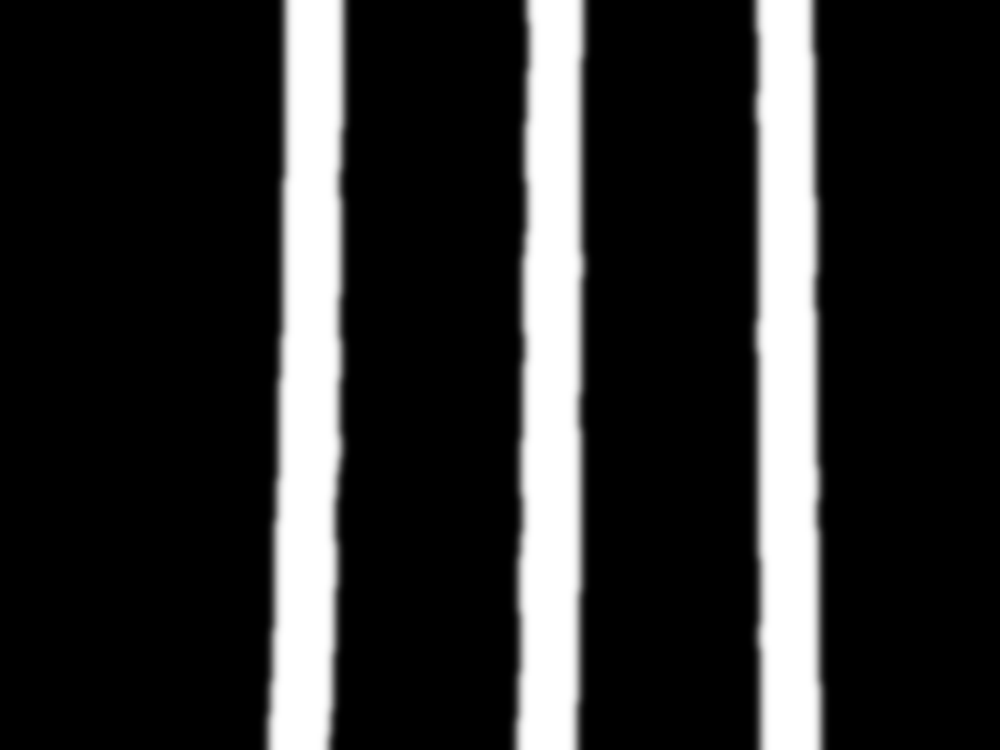}
            \includegraphics[width=\linewidth]{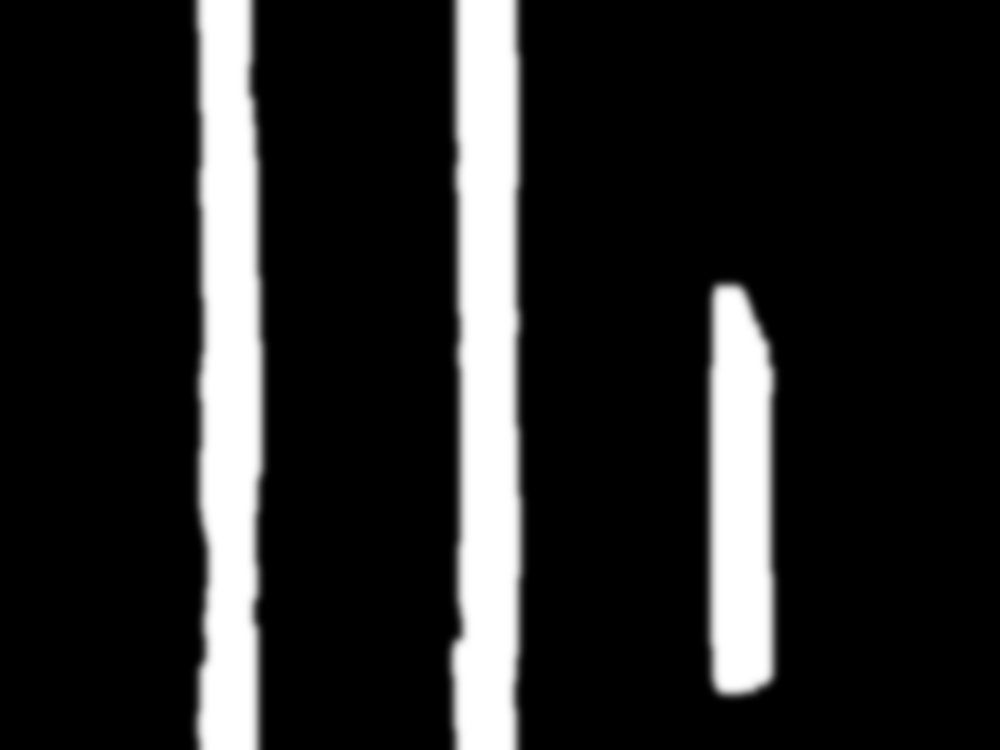}
            \includegraphics[width=\linewidth]{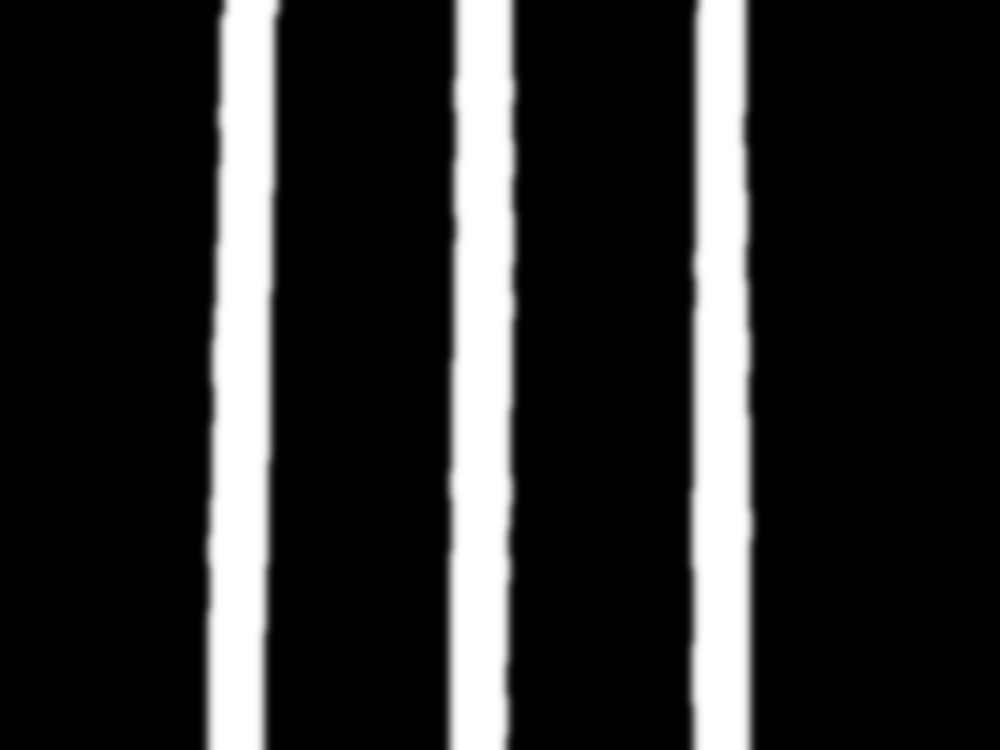}%
        \end{minipage}%
        \label{subfig:col_lightmasks}%
    }\hfil
    \subfloat[]{%
        \begin{minipage}{.24\columnwidth}
            \centering
            \includegraphics[width=\linewidth]{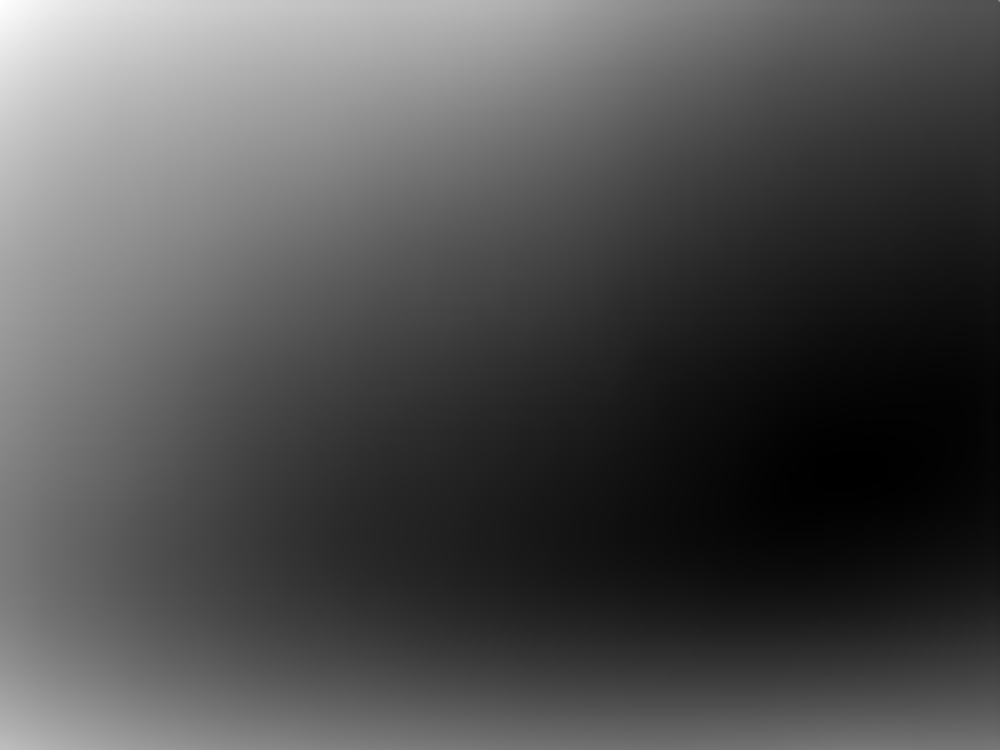}
            \includegraphics[width=\linewidth]{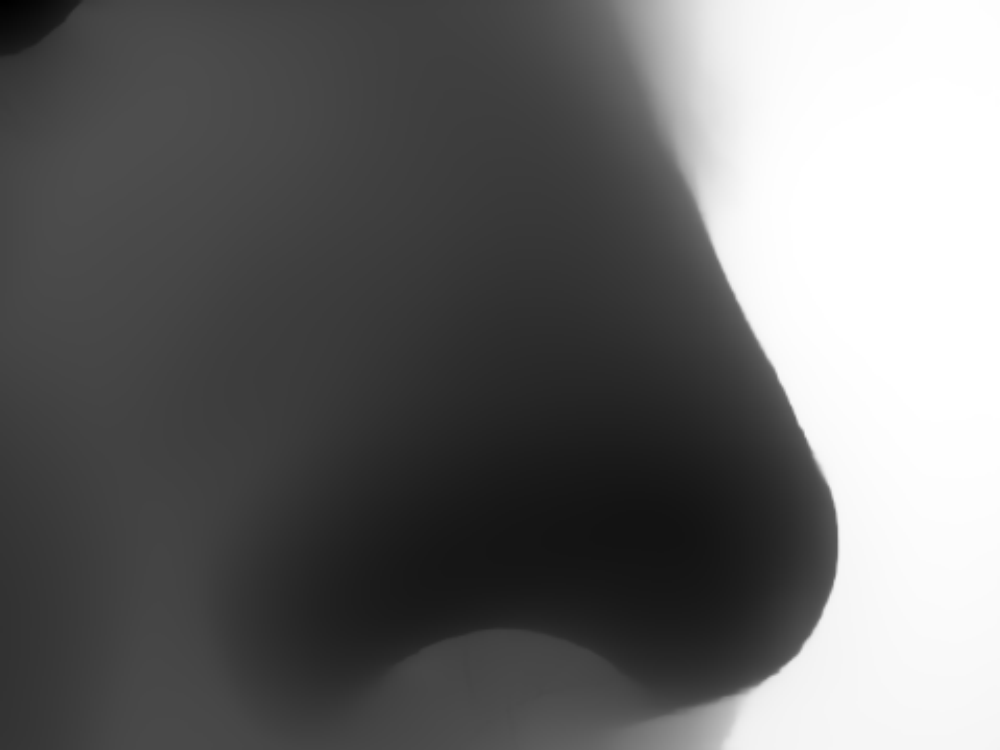}
            \includegraphics[width=\linewidth]{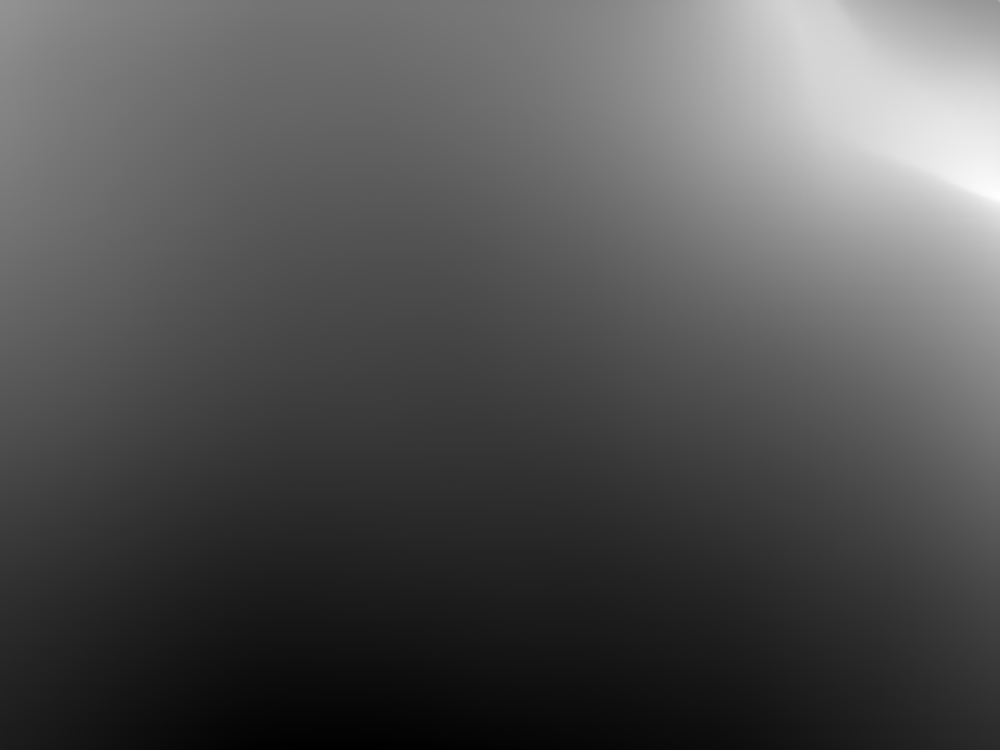}%
        \end{minipage}%
        \label{subfig:col_depthmaps}%
    }
    \caption{Optimization input data evaluated across different anatomical skin regions (rows). From left to right: (a) Uniformly illuminated reference frames, (b) active strip-lit captures displaying subsurface scattering effects, (c) light masks, and (d) estimated monocular depth maps.}
    \label{fig:skin_patches}
    \vspace{-2em}
\end{figure}

To isolate subsurface scattering effects, we generate a light mask by computing the difference between the patterned image (\ref{subfig:skin_photo_2}) and its inverse (\ref{subfig:skin_photo_3}). We binarize this difference map using Otsu’s thresholding \cite{Otsu1979AHistograms}, which automatically determines an optimal threshold by maximizing the between-class variance of the intensity distribution. To ensure spatial coherence, we apply morphological dilation followed by opening and closing operations. This process produces two distinct versions of the mask to balance optimization accuracy with visual reconstruction:
\begin{enumerate}
    \item \textit{Loss Mask}: An inverted, strictly binary shadow mask $\mathcal{M}$ (where $1$ corresponds to shadowed regions) used to focus the optimization loss. Maintaining a hard binary edge here is critical; any blurring of the mask during optimization would alter the edges of the illumination pattern and corrupt the scattering halo estimation.
    \item \textit{Rendering Mask}: A non-inverted light mask $\mathcal{M}_{\text{smooth}}$ (where $1$ corresponds to lit regions) smoothed via an $11 \times 11$ Gaussian filter. This blur emulates projector defocus and prevents hard-edge artifacts in the final render, ensuring a smooth transition where the specular component is added to the masked diffuse component.
\end{enumerate}
The resulting preprocessed images and the depth map (Fig.~\ref{fig:skin_patches}) serve as the input for our differentiable rasterizer optimization.

\subsection{Tensorized SSSS}
\label{subsec:ssss_vectorized}
Porting the screen-space multiple scattering algorithm to a differentiable framework presents a challenge: we must convert per-pixel shader operations into batched tensor computations to ensure efficiency and parallelization. This requires careful handling of texture sampling, coordinate projection, and memory management to maintain differentiability.

Key aspects of this transformation include:
\begin{enumerate}[leftmargin=*]
    \item \textit{Batch Processing} – Instead of iterating per pixel, we process the entire image and all scattering samples simultaneously as high-dimensional tensors.
    \item \textit{View-Space Reconstruction} – To ensure 1:1 consistency with the rendering engine, we reconstruct the 3D position of every pixel using depth and camera intrinsics, rather than relying on 2D screen-space approximations.
    \item \textit{Differentiable Sampling} – We replace discrete texture lookups with differentiable bilinear interpolation, allowing gradients to propagate through the scattering inputs.
    \item \textit{Memory Optimization} – We broadcast operations across dimensions to minimize loop overhead, a critical requirement for performance in tensorized frameworks.
\end{enumerate}

The pipeline begins by computing the diffuse irradiance tensor $\mathbf{L}_T$ modulated by the smoothed rendering mask $\mathcal{M}_{\text{smooth}}$ to filter the incoming illumination before diffusion: $\mathbf{L}_T = \mathcal{M}_{\text{smooth}} \odot (\mathbf{n} \cdot \mathbf{l})_+$, where $\mathbf{n}$ and $\mathbf{l}$ denote the surface normal and light direction tensors.

\begin{figure}
    \centering
    \includegraphics[width=.9\columnwidth]{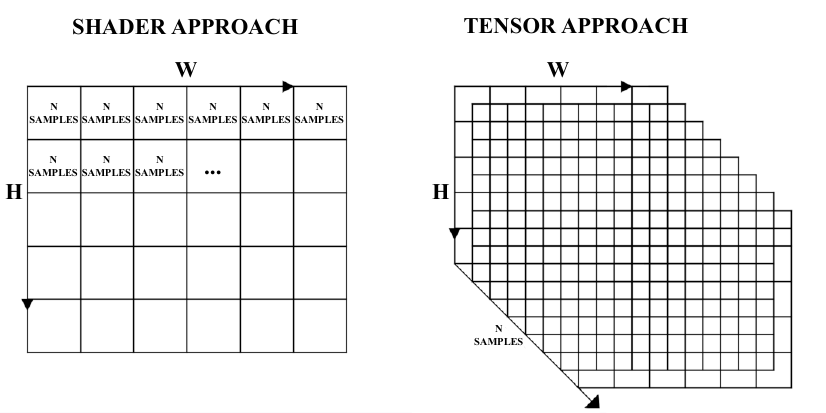}
    \caption{Comparison of the shader-based approach (left) vs. our tensorized scheme (right). By broadcasting the image across the sample dimension $N$, we perform view-space reconstruction and integration in a single parallelized pass.}
    \label{fig:tensorized_scheme}
    \vspace{-2em}
\end{figure}

In a standard shader, diffusion is computed via a loop over $N$ samples for each pixel. Imitating this loop in a tensor framework is inefficient. Instead, we adopt a fully vectorized approach where the entire image is replicated $N$ times, creating tensors of shape $(N, H, W, C)$. This allows us to apply the scattering kernel across all pixels and samples simultaneously (Fig.~\ref{fig:tensorized_scheme}). The process is as follows:

We define a virtual camera matching the real-world acquisition system (focal length $50\,\text{mm}$, sensor width $23.5\,\text{mm}$). Using the estimated depth map (Fig.~\ref{subfig:col_depthmaps}), we unproject every pixel into 3D view-space coordinates $\mathbf{P}_v$. To avoid high-dimensional matrix inversions during backpropagation, we linearize depth and unproject coordinates component-wise:
\begin{equation}
    \mathbf{z} = \frac{P_{23}}{(2 \cdot \mathbf{D}_T - 1) - P_{22}}
    \label{eq:depth_linearization}
\end{equation}
\begin{equation}
    \mathbf{x}_v = \frac{\mathbf{U} \odot \mathbf{z}}{P_{00}}, \quad \mathbf{y}_v = \frac{\mathbf{V} \odot \mathbf{z}}{P_{11}}
    \label{eq:lateral_unprojection}
\end{equation}
where $\mathbf{P}_v = [\mathbf{x}_v, \mathbf{y}_v, -\mathbf{z}]^T$, the tensors $\mathbf{U}, \mathbf{V}$ represent the normalized grid coordinates, and $P_{00}, P_{11}, P_{22}, P_{23}$ are the perspective projection coefficients derived from the camera geometry and clipping planes. This explicit tensorized reconstruction eliminates the need for 2D pixel-distance approximations, ensuring pixel-level geometric alignment with the WebGL forward rasterizer while decoupling the simulation from the fixed image plane for robust angular consistency, as validated in Section~\ref{subsec:validation}.

We utilize the precomputed sampling tensors $\mathbf{samples}_{\theta}$ and $\mathbf{samples}_r$ representing the Fibonacci angles and stratified Burley Inverse CCDF normalized radii, respectively. The angles are shifted by the stochastic jitter pattern, while the normalized radii are scaled by the maximum scattering distance $r_{\text{max}}$. These updated polar coordinates are then projected to obtain the spatial offsets $(\mathbf{x}_o, \mathbf{y}_o)$.

We apply these offsets to $\mathbf{P}_v$ in view space and project the resulting positions back into Normalized Device Coordinates (NDC) to form a sampling grid. This perspective projection is defined component-wise as:
\begin{equation}
    \mathbf{U}_s = \frac{(\mathbf{x}_v + \mathbf{x}_o) \cdot P_{00}}{\mathbf{z}}, \quad \mathbf{V}_s = \frac{(\mathbf{y}_v + \mathbf{y}_o) \cdot P_{11}}{\mathbf{z}}
    \label{eq:perspective_projection}
\end{equation}
where $\mathbf{U}_s$ and $\mathbf{V}_s$ are stacked to form the final 2D sampling coordinate grid $\mathbf{G}$. To accurately replicate the hardware-accelerated bilinear filtering used in standard shaders, we retrieve the incident irradiance ($\mathbf{L}_T$) and depth ($\mathbf{D}_T$) at these coordinates using grid sampling with bilinear interpolation. This ensures our optimization kernel mathematically matches the runtime renderer while maintaining full differentiability.

We calculate the true 3D Euclidean distance between the surface point $\mathbf{P}_v$ and the sampled point $\mathbf{P}_s$. This distance is fed into the Burley diffusion profile to compute the scattering weight (Eq.~\ref{eq:diffusion_weight}). The weighted samples are summed along the sample dimension ($N$) to produce the integrated diffuse irradiance tensor.

Finally, this integrated diffuse irradiance is multiplied element-wise by the albedo tensor $\mathbf{A}_T$ to obtain the final scattered diffuse radiance component. For visualization during optimization, we add the specular lobe, modulated by the rendering mask $\mathcal{M}_{\text{smooth}}$, to produce the complete rendered image. To ensure full reproducibility, the tensorized implementation is provided in Section~5.4 of the Supplementary Material.

\section{Implementation details}
\label{sec:implementation_details}
\paragraph*{Capture Hardware}
To capture the input data, we utilize a Sony $\alpha$6100 (ILCE-6100L), a consumer-grade camera that balances affordability with high image quality. Its high-resolution sensor and reliable low-light performance make it an effective instrument for acquiring precise material data, while its accessible price point ensures framework reproducibility. The camera is connected via USB to a PC and remotely controlled using the \href{https://creatorscloud.sony.net/catalog/es-es/ie-desktop/index.html}{Sony Imaging Edge} desktop application, which provides a live view, full manual parameter control, and automated interval shooting. All images were captured using fixed camera parameters to ensure strict consistency across the dataset. Specifically, we established a resolution of $6020 \times 4024$ pixels, an ISO of $100$, an exposure time of $1/20\,\text{s}$, an aperture of $f/20$, a white balance of $6000\,\text{K}$, and a $50\,\text{mm}$ focal length.

Illumination is provided by an Optoma projector rated at $3500\,\text{lumens}$, which projects the sequence of structured light patterns. We synchronize the camera and projector using a combination of Sony Imaging Edge and an \href{https://www.autohotkey.com/}{AutoHotkey} script. While Imaging Edge manages the timed shooting in continuous mode, the AutoHotkey script coordinates each projection frame change with the camera's shutter trigger, ensuring precise temporal synchronization between pattern display and image acquisition.

\paragraph*{Input processing}
The RAW captures (.ARW) are decoded using \textit{rawpy} and subsequently center-cropped to $480 \times 360\,\text{px}$ to satisfy GPU memory constraints. To ensure chromatic accuracy, we calibrate the raw sensor space to linear sRGB using a $3 \times 3$ Color Correction Matrix (CCM) computed via Ridge Regression on an X-Rite ColorChecker, normalizing pixel intensities to the $[0, 1]$ range. For complete processing and calibration details, see Section~5.1 of the Supplementary Material.

\paragraph*{Optimization details}
Our pipeline optimizes a subset of material property tensors from a single view under a single light source, specifically, the spatially varying albedo $\mathbf{A}_T$ and roughness $\boldsymbol{\rho}_T$, alongside the homogeneous scattering distance $\mathbf{S}_T$. Due to the limited geometric cues of this underconstrained setup, we do not optimize surface normals and tangents, relying instead on a fixed Sobel-filtered map derived from a pre-estimated depth map~\cite{Yang2024DepthV2}. To comply with commercial production standards, the parameter $\mathbf{S}_T \in [0, 1]$ (encoding up to $1.0$\,cm) maps to world-space distance in meters via the metric conversion $d = \mathbf{S}_T \times 0.01$, where $d_{\text{opt}}$ represents the resulting homogeneous scattering color texture map. Achieving a clean separation of the diffuse albedo $\mathbf{A}_T$ is critical to avoid bias during the scattering optimization; since the Burley diffusion profile (Eq.~\ref{eq:burley_equation}) scales directly with $\mathbf{A}$, any specular energy incorrectly baked into the albedo would artificially shift its chromatic tone. To capture these high-frequency highlights without corrupting the core material parameters, we introduce an auxiliary residual specular tensor $\boldsymbol{\Delta}_{\text{spec}, T}$. This tensor functions as a temporary residual buffer that absorbs specular discrepancies during optimization, mitigating their bleeding into the base material maps, and is discarded after convergence.

Textures are initialized using physically informed priors: the unwarped IOR map is initialized at a uniform value of $0.145$ (linearly mapping to a physical IOR $\approx 1.44$), and the homogeneous scattering distance map $\mathbf{S}_T$ begins at a uniform $0.5$. To isolate a clean skin albedo independent of lighting, hair, or deep shadows, we perform $k$-means clustering ($k=2$) on the $a^*$ and $b^*$ chromatic channels in CIE LAB space. This adaptive mask is refined by excluding the brightest $10\%$ and darkest $5\%$ of pixels. A robust mean color is then computed via $2\sigma$ Sigma Clipping on the remaining pixels to provide a stable initialization baseline (see Section~5.2 of the Supplementary Material for complete algorithmic details).

We employ a multi-stage optimization strategy using the Adam optimizer~\cite{Kingma2017Adam:Optimization}. The objective function balances a masked $\mathcal{L}_1$ reconstruction loss with an $\mathcal{L}_2$ roughness regularization prior:
\begin{equation}
\mathcal{L} = \frac{1}{|\Omega|} \left\| \hat{\mathbf{I}} \odot \mathcal{M}_{\text{opt}} - \mathbf{I} \odot\mathcal{M}_{\text{opt}} \right\|_1 + \lambda_{\text{prior}} \frac{1}{|\Omega|} \left\| \boldsymbol{\rho}_T - 0.6 \right\|_2^2
\end{equation}
where $\mathbf{I}$ is the captured reference tensor, $\mathcal{M}_{\text{opt}}$ denotes the stage-dependent optimization mask and $|\Omega|$ is the total number of elements within the active spatial domain. The $\mathcal{L}_2$ value prior minimizes the transfer of high-frequency albedo details into the roughness map under single-view constraints, stabilizing the reflection envelope without suppressing regional variability. The anchor value of 0.6 corresponds to the default skin roughness baseline adopted by industry-standard real-time engines. Within this framework, the complete rendered image is formulated as a differentiable function $\hat{\mathbf{I}} = f(\mathbf{A}_T, \boldsymbol{\rho}_T, \boldsymbol{\Delta}_{\text{spec}, T}, \mathbf{S}_T)$, whose active parameter dependencies, hyperparameter weights, and spatial domains adapt dynamically according to the multi-stage schedule outlined in Table~\ref{tab:optimization_schedule} and detailed below:

\begin{table}[t]
\centering
\caption{Optimization Schedule and Hyperparameters.}
\label{tab:optimization_schedule}
\resizebox{\columnwidth}{!}{%
\begin{tabular}{@{}lccc@{}}
\toprule
\textbf{Configuration} & \textbf{Stage 1} & \textbf{Stage 2} & \textbf{Stage 3} \\ \midrule
\textbf{Optimized Parameters} & $\boldsymbol{\Delta}_{\text{spec}, T}, \boldsymbol{\rho}_T$ & $\mathbf{A}_T, \boldsymbol{\rho}_T, \boldsymbol{\Delta}_{\text{spec}, T}$ & $\mathbf{S}_T$ \\
\textbf{Frozen Parameters} & $\mathbf{A}_T, \mathbf{S}_T$ & $\mathbf{S}_T$ & $\mathbf{A}_T, \boldsymbol{\rho}_T, \boldsymbol{\Delta}_{\text{spec}, T}$ \\
\textbf{Iterations} & 700 & 500 & 600 \\
\textbf{Learning Rate} & $5 \times 10^{-3}$ & $5 \times 10^{-3}$ & $1 \times 10^{-3}$ \\
\textbf{Active Mask ($\mathcal{M}_{\text{opt}}$)} & Full Image ($\mathbf{1}$) & Full Image ($\mathbf{1}$) & Shadow Mask ($\mathcal{M}$) \\
\textbf{Roughness Prior ($\lambda_{\text{prior}}$)} & $1 \times 10^{-2}$ & $1 \times 10^{-2}$ & None \\
\bottomrule
\end{tabular}%
}
\vspace{-1em}
\end{table}

\begin{itemize}[leftmargin=*]
    \item \textbf{Stage 1 (Specular \& Roughness Pre-fit):} 
    Parameters $\boldsymbol{\Delta}_{\text{spec}, T}$ and $\boldsymbol{\rho}_T$ are optimized; $\mathbf{A}_T$ and $\mathbf{S}_T$ are frozen. This stage fits global reflection, capturing unmodeled specular highlights into $\boldsymbol{\Delta}_{\text{spec}, T}$. Freezing albedo preserves its clustering initialization, shielding base color from early lighting errors and cross-parameter interference.

    \item \textbf{Stage 2 (Surface Maps Refinement):} 
    Parameters $\mathbf{A}_T$, $\boldsymbol{\rho}_T$, and $\boldsymbol{\Delta}_{\text{spec}, T}$ are optimized; $\mathbf{S}_T$ is frozen. Unfreezing $\mathbf{A}_T$ absorbs diffuse details, allowing the $\mathcal{L}_2$ prior to smooth $\boldsymbol{\rho}_T$. Retaining active $\boldsymbol{\Delta}_{\text{spec}, T}$ prevents residual specular discrepancies from being incorrectly baked into albedo.

    \item \textbf{Stage 3 (Subsurface Scattering Fit):} 
    Parameter $\mathbf{S}_T$ is optimized; $\mathbf{A}_T$, $\boldsymbol{\rho}_T$, and $\boldsymbol{\Delta}_{\text{spec}, T}$ are frozen. This isolates the low-frequency subsurface scattering pass from high-frequency surface variations, preventing the coupling inherent to Eq.~\eqref{eq:burley_equation} between linear albedo scaling and non-linear decay $d$ (verified synthetically in Figures~\ref{fig:synthetic_validation} and~\ref{fig:monk_synthetic_validation}). Restricting the loss domain $\mathcal{M}_{\text{opt}}$ to the shadow mask $\mathcal{M}$ forces the optimizer to fit the multiple scattering halos exclusively, removing interference from surface reflections and preventing direct lighting discrepancies from skewing $d$ (Figure~\ref{fig:masked_loss_validation}).
\end{itemize}

This schedule is held constant across all captured patches and skin tones, requiring no per-subject hyperparameter tuning. The learning rates, iteration counts, and prior weight for surface estimation (Stages~1 and~2) build upon established baseline configurations~\cite{Garces2023TowardsSystem}, with the residual specular map $\boldsymbol{\Delta}_{\text{spec}, T}$ sharing the same learning rate ($5 \times 10^{-3}$). For Stage~3, we reduce the learning rate to $1 \times 10^{-3}$ over 600 iterations to ensure stable convergence against the lower gradient magnitudes of subsurface scattering, particularly across darker skin tones where the signal-to-noise ratio is lower. The progressive evolution of material maps across these stages is illustrated in Section~2 of the Supplementary Material.

\section{Results}
\label{sec:results}
We evaluate our method qualitatively and quantitatively on both real and synthetic data to demonstrate its reconstruction fidelity, computational efficiency, and robustness across diverse skin tones. Our evaluation is structured into two main parts: (1) an analysis of fitting results on real facial patches across the Monk Skin Tone Scale, evaluating perceptual metrics, runtime benchmarks, and material plausibility (Section~\ref{subsec:fitting_results}); and (2) a comprehensive validation study verifying parameter isolation via synthetic ground truth, assessing our shadow-masked loss through ablation, and benchmarking our single-light setup against a multi-light configuration (Section~\ref{subsec:validation}). Validation of angular consistency via tilted-view experiments is provided in Section~3 of the Supplementary Material.

We built a dataset of $225$ real skin patches from $45$ participants across five facial regions using a single-light setup. A subset of 17 subjects (85 samples) was additionally captured using a multi-light setup for validation.

\paragraph*{Ethics, consent, and data handling}. 
The acquisition protocol was reviewed and approved by the Research Ethics Committee of our institution prior to data collection. All participants received written information describing the purpose and procedures of the study and provided written informed consent prior to participation, including consent for the processing of their personal data. Participation was voluntary, and participants could withdraw their consent at any time without adverse consequences. Skin images were associated with generic participant identifiers, while the correspondence between these identifiers and the contact information of the participants was stored separately under restricted access available only to the research team. Contact information was used solely when a participant needed to be contacted for a repeated acquisition. Only non-identifying skin-patch images were used for scientific dissemination, and any additional compatible research use requires prior anonymization. Personal data were handled according to applicable international regulations.

\subsection{Fitting Results}
\label{subsec:fitting_results}
\begin{figure}
    \centering
    \includegraphics[width=\columnwidth]{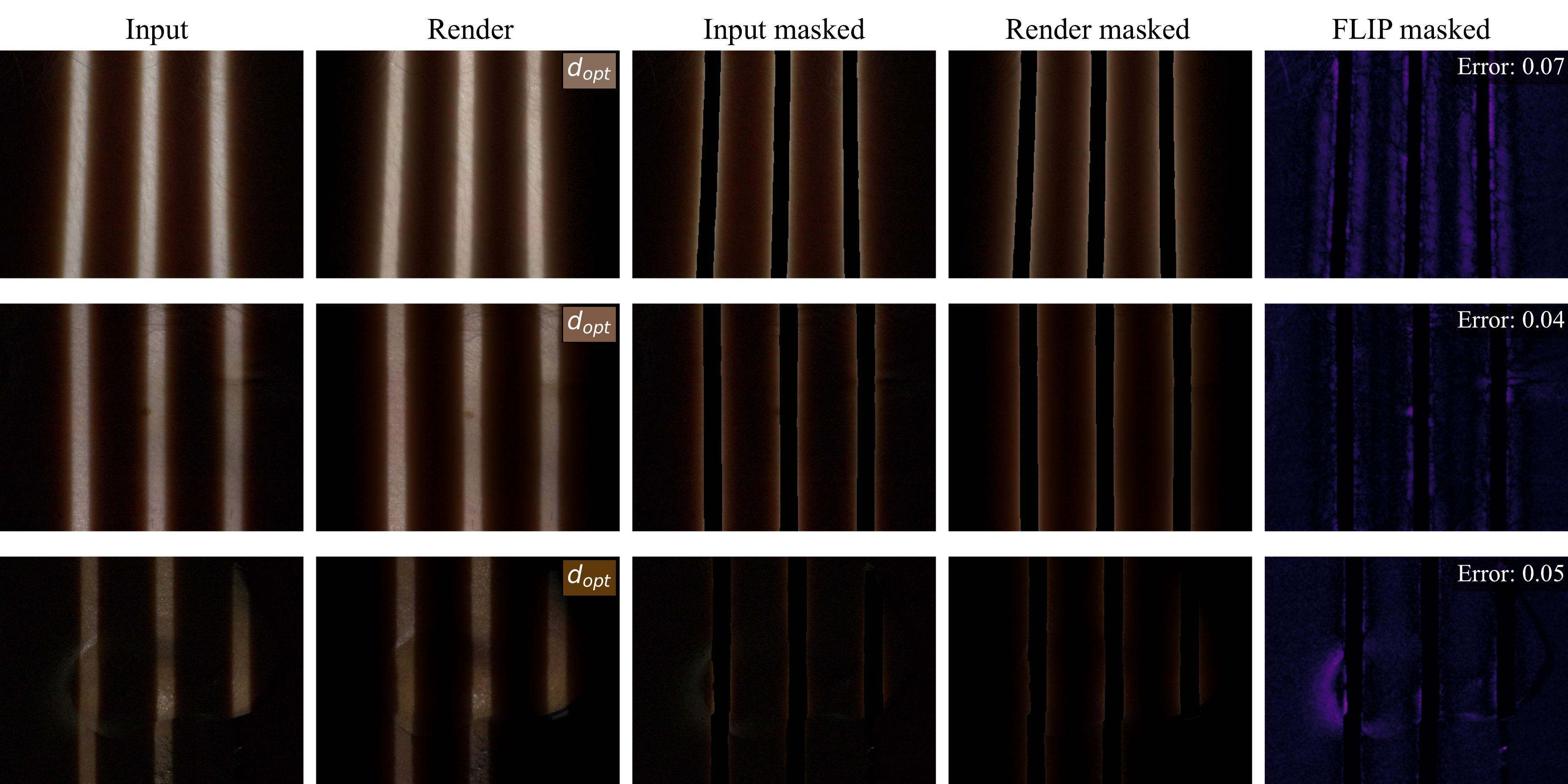}
    \caption{Reconstruction results on real facial skin patches across the Monk Skin Tone Scale. Rows display different facial regions: neck (top), cheek (middle), and nose (bottom). Columns show (left to right): the captured input image with strip lighting, the rendered reconstruction, the masked input image, the masked render, and the masked perceptual error map (FLIP). The optimized scattering distance for each patch is indicated as $d_{opt}$.}
    \label{fig:single_image_fitting}
    \vspace{-1em}
\end{figure}

We apply our inverse rendering pipeline to real facial skin patches sampled across the Monk Skin Tone Scale. We evaluate reconstruction quality using the perceptual metric FLIP~\cite{Andersson2020FLIP}, computed over masked regions to isolate the subsurface scattering halo. Fig.~\ref{fig:single_image_fitting} shows representative examples from light, medium, and dark skin tones. The reconstructions closely match the input photographs, particularly in the shadowed transition regions where scattering dominates.

We report complementary metrics—LPIPS \cite{Zhang2018TheMetric}, SSIM \cite{ZhouWang2004ImageSimilarity}, and PSNR—in Table~\ref{tab:metrics}. While SSIM is sensitive to luminance variations in shadowed regions (resulting in lower scores), perceptual metrics like FLIP and LPIPS yield consistently low error values. This supports the perceptual plausibility of our reconstructions across the full skin tone spectrum.
\begin{table}
  \centering
  \scriptsize
  \caption{Quantitative reconstruction errors across the Monk Scale, computed on masked image regions.}
  \label{tab:metrics}
  \resizebox{\columnwidth}{!}{%
    \begin{tabu} to \linewidth {r *{4}{c}}
      \toprule
      Monk Range & FLIP $\downarrow$ & LPIPS $\downarrow$ & SSIM $\uparrow$ & PSNR $\uparrow$ \\
      \midrule
      Monk 1-3  & $0.093 \pm 0.026$ & $0.086 \pm 0.020$ & $0.615 \pm 0.066$ & $28.61 \pm 1.68$ \\
      \midrule
      Monk 4-6  & $0.093 \pm 0.031$ & $0.088 \pm 0.018$ & $0.614 \pm 0.066$ & $28.67 \pm 1.89$ \\
      \midrule
      Monk 7-10 & $0.077 \pm 0.021$ & $0.115 \pm 0.015$ & $0.503 \pm 0.072$ & $29.65 \pm 1.35$ \\
      \bottomrule
    \end{tabu}%
    }
    \vspace{-1em}
\end{table}

In addition to perceptual quality, we benchmark the computational efficiency of our pipeline. Because all skin patches are standardized to $480\times360$~px and evaluated under our fixed multi-stage schedule, runtime is deterministic across subjects. Table~\ref{tab:computational_cost} details the stage-wise computational breakdown on an NVIDIA GeForce RTX 4070~Ti (12~GB). Inverse optimization converges in under 3.5 minutes ($198.4$~s) per patch with a peak footprint of $9.4$~GB during Stage~3 integration, confirming that our pipeline enables fast and accessible digitization on consumer desktop hardware.
\begin{table}
  \centering
  \scriptsize
  \caption{Computational cost per skin patch ($480\times360$~px, $N=128$ scattering samples) across optimization stages on an AMD Ryzen 9 7900 CPU (12 cores, 24 threads) and an NVIDIA GeForce RTX 4070~Ti GPU. The \textit{Overall} column reports cumulative values for time and iterations, mean iteration latency, and maximum peak VRAM.}
  \label{tab:computational_cost}
  \resizebox{\columnwidth}{!}{%
    \begin{tabu} to \linewidth {l *{4}{c}}
      \toprule
      Metric & Stage 1 & Stage 2 & Stage 3 & Overall \\
      \midrule
      Execution Time (s)       & 51.1  & 38.3  & 109.0 & 198.4 \\
      \midrule
      Iterations               & 700   & 500   & 600   & 1800  \\
      \midrule
      Time per Iteration (ms)  & 72.9  & 76.6  & 181.7 & 110.2 \\
      \midrule
      Peak VRAM (GB)           & 2.5   & 2.6   & 9.4   & 9.4   \\
      \bottomrule
    \end{tabu}%
    }
    \vspace{-1em}
\end{table}

\begin{figure}
    \centering
    \includegraphics[width=\columnwidth]{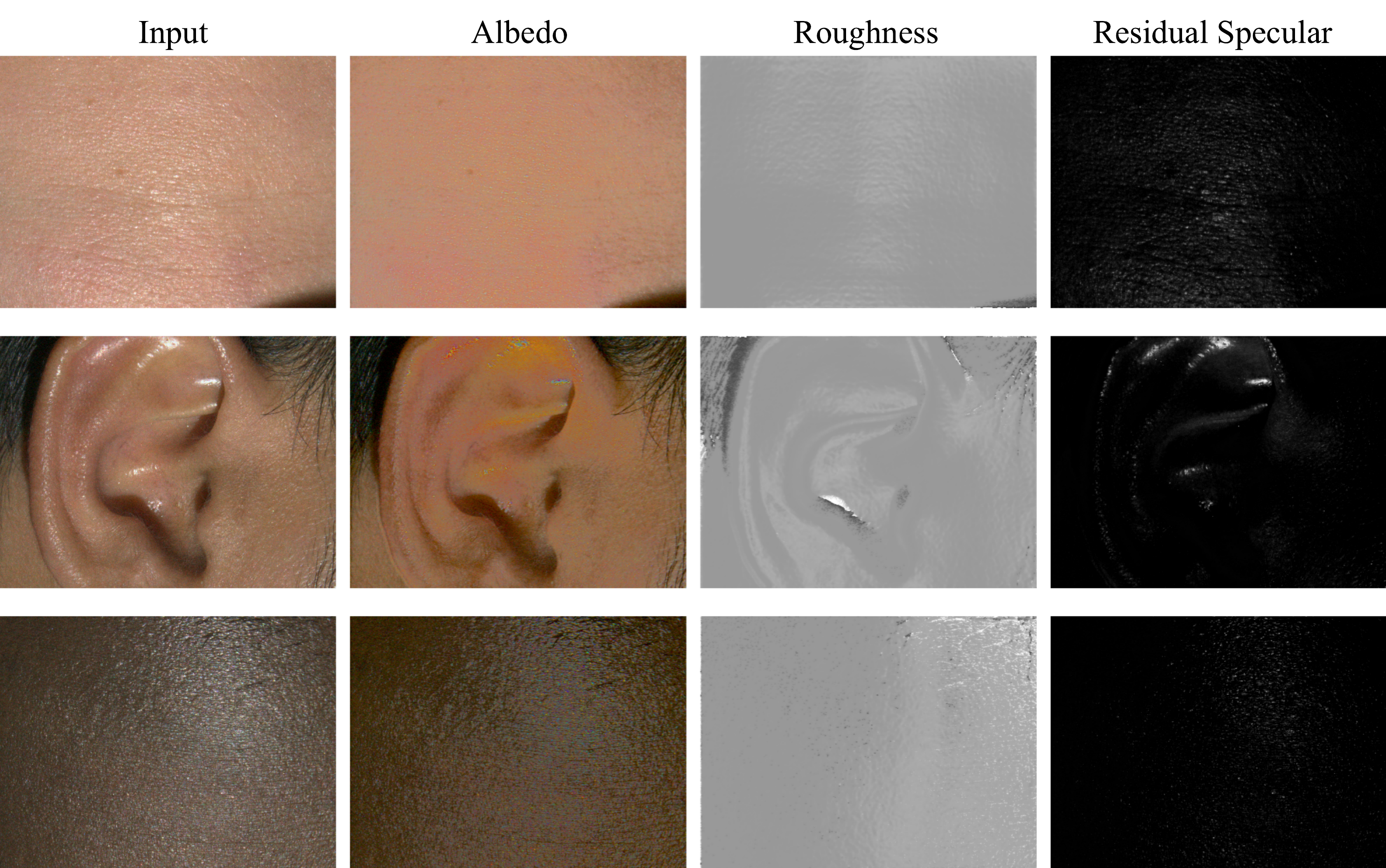}
    \caption{Optimized Burley 15 material stack across different skin tones from single-view, single-light capture. Columns show (left to right): full-light input image, estimated albedo, roughness and residual specular.}
    \label{fig:texture_stack_single}
    \vspace{-1em}
\end{figure}

We visualize the optimized Burley 15 material stack alongside the auxiliary residual specular map in Fig.~\ref{fig:texture_stack_single}. The estimated textures are spatially smooth and consistent across the dataset. Utilizing a monocular depth-based normal prior, rather than relying on photometric cues, regularizes the single-view optimization, mitigating parameter instability across complex structures like the ear. While this depth prior acts as a low-frequency approximation that lacks sharp pore-level details—yielding smoother roughness maps—and screen-space shading partially bakes self-shadows into albedo across steep relief (e.g., ear folds), our multi-light validation confirms the scattering parameter $d$ converges consistently. This demonstrates that subsurface optimization remains effectively decoupled from localized shadowing and micro-geometric inaccuracies.

\begin{figure}
  \centering
  \includegraphics[width=\columnwidth]{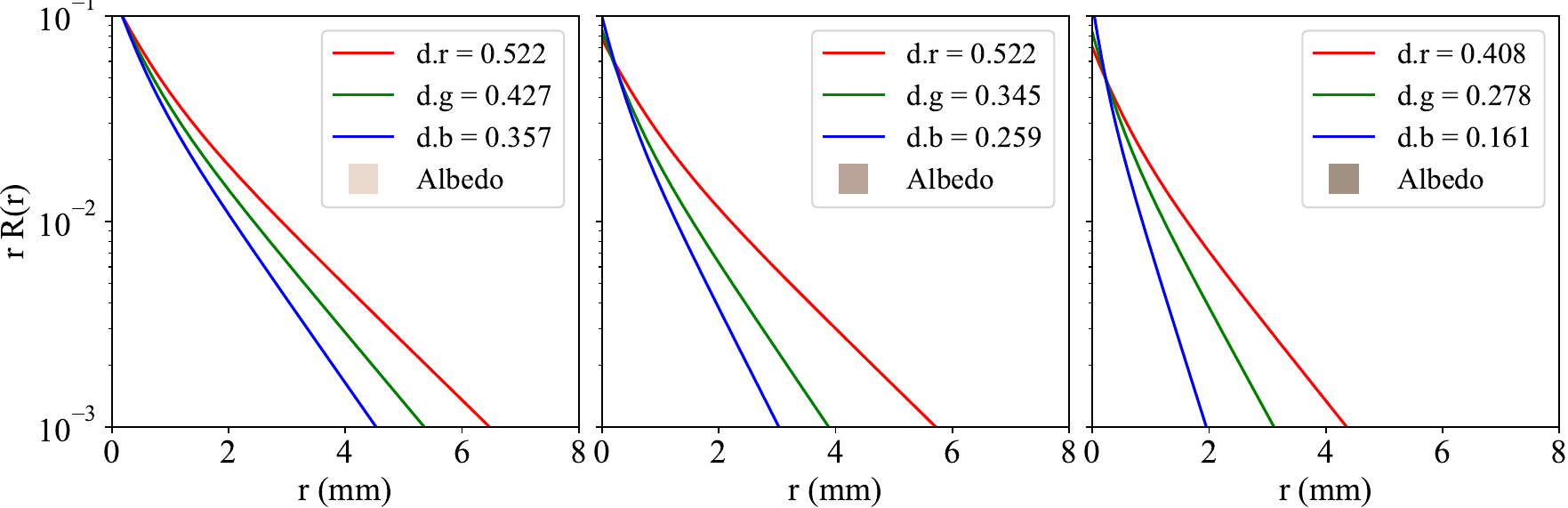}
  \caption{Radial scattering profiles plotted as $rR(r)$ in logarithmic space to visualize falloff and effective range. From left to right: light, medium, and dark skin tones. Curves show channel-wise reflectance decay (RGB) alongside estimated albedo. The trend of decreasing scattering distance with darker skin tones confirms the physical plausibility of our optimization.}
  \label{fig:scatter_plots}
  \vspace{-2em}
\end{figure}

Crucially, the estimated scattering distance maps exhibit intuitive trends across the skin tone spectrum. As illustrated in Fig.~\ref{fig:scatter_plots}, darker skin tones correspond to shorter scattering distances, consistent with the reduced subsurface light transport in melanin-rich tissue. These findings demonstrate the physical plausibility of our RGB-based optimizations without the need for hyperspectral data. Furthermore, our masking strategy effectively focuses supervision on shadowed regions, improving multiple scattering estimates in darker skin where the effect is subtle but directionally dependent.

\subsection{Validation}
\label{subsec:validation}

\begin{figure}
    \centering
    \includegraphics[width=\columnwidth]{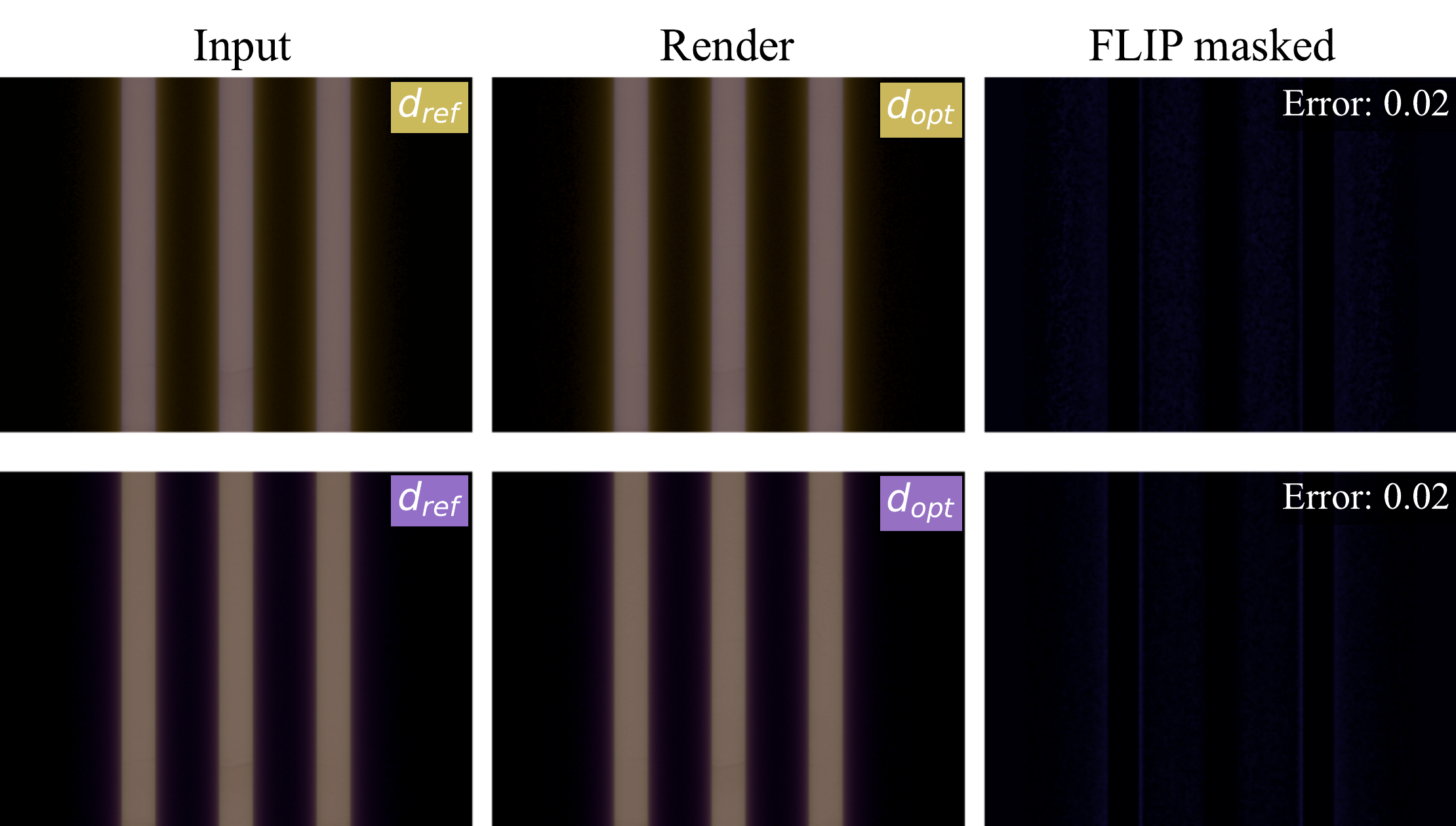}
    \caption{Synthetic validation using known ground truth. $d_{\text{ref}}$ denotes the ground-truth scattering distance, and $d_{\text{opt}}$ is the optimized value. The near-identical maps confirm the numerical self-consistency of $d$ under the inverse rendering formulation.}
    \label{fig:synthetic_validation}
    \vspace{-1em}
\end{figure}

\paragraph*{Validation on Synthetic Skin Patches}
We validate our pipeline using synthetic skin patches rendered under known Burley parameters~\cite{Burley2015ExtendingScattering} with an identical sampling budget ($N=128$) to ensure numerical consistency during inverse optimization. In the first experiment, we fix all surface attributes (albedo, normals, roughness, IOR) and optimize only the scattering distance. This isolates the scattering estimation stage. As shown in Fig.~\ref{fig:synthetic_validation}, the estimated scattering maps ($d_{opt}$) closely match the reference ground truth ($d_{ref}$), and the rendered reconstructions are visually indistinguishable from the inputs.

In the second experiment, we fix the scattering distance and vary the albedo across the ten synthetic tones of the Monk Scale to test robustness to pigmentation changes. While Figure~\ref{fig:monk_synthetic_validation} shows a representative subset, Section~4 of the Supplementary Material details the full 10-tone evaluation alongside the real-world dataset distribution ($N=45$) cross-referenced against participant self-assessments.

\begin{figure}
    \centering
    \includegraphics[width=\columnwidth]{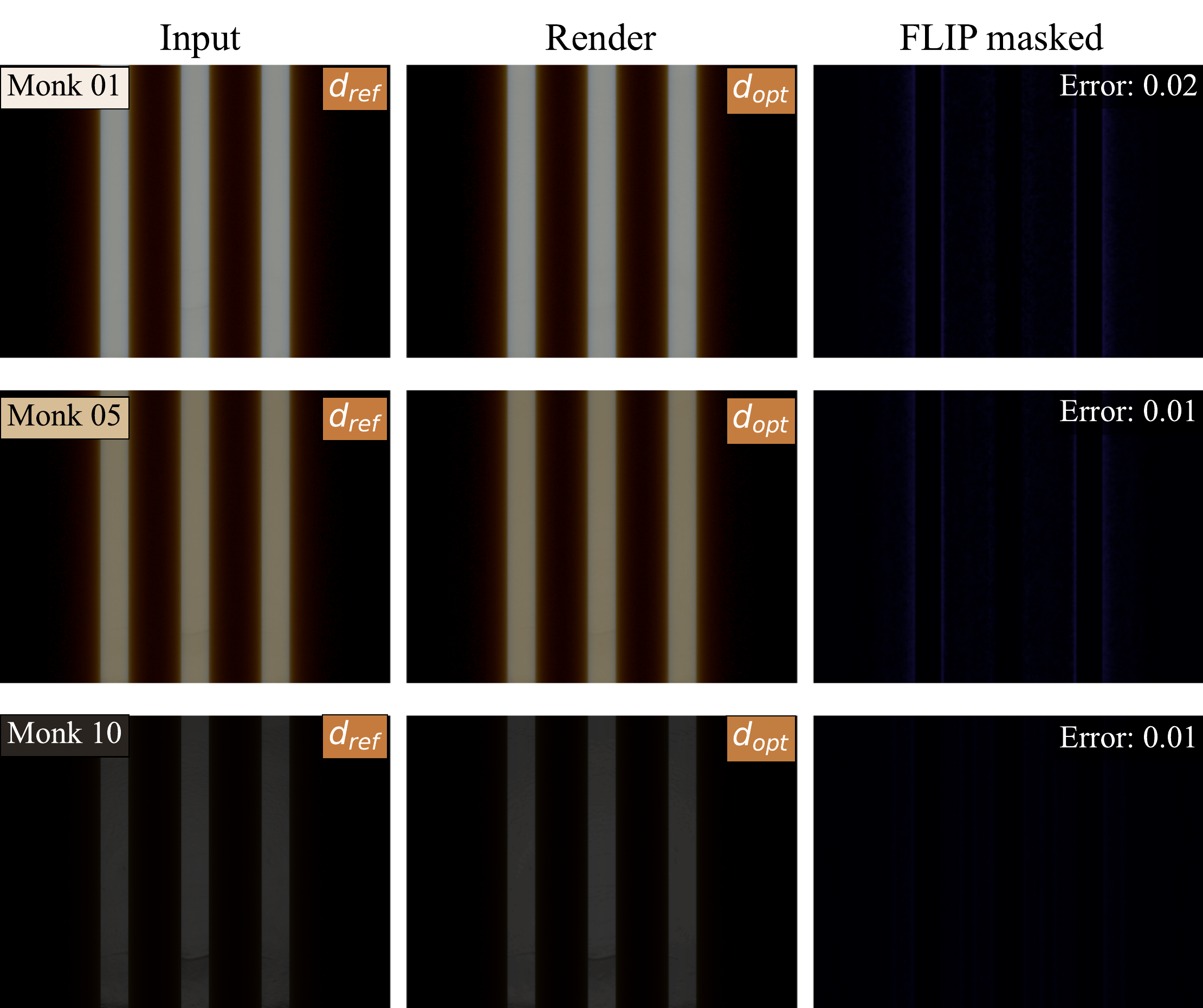}
    \caption{Synthetic validation across the Monk Scale. We fix scattering distance and vary albedo (Monk 01, 05, 10). The low perceptual error (FLIP) and consistent parameter recovery confirm robustness to albedo variation.}
    \label{fig:monk_synthetic_validation}
    \vspace{-1em}
\end{figure}

\begin{figure}
    \centering
    \includegraphics[width=\columnwidth]{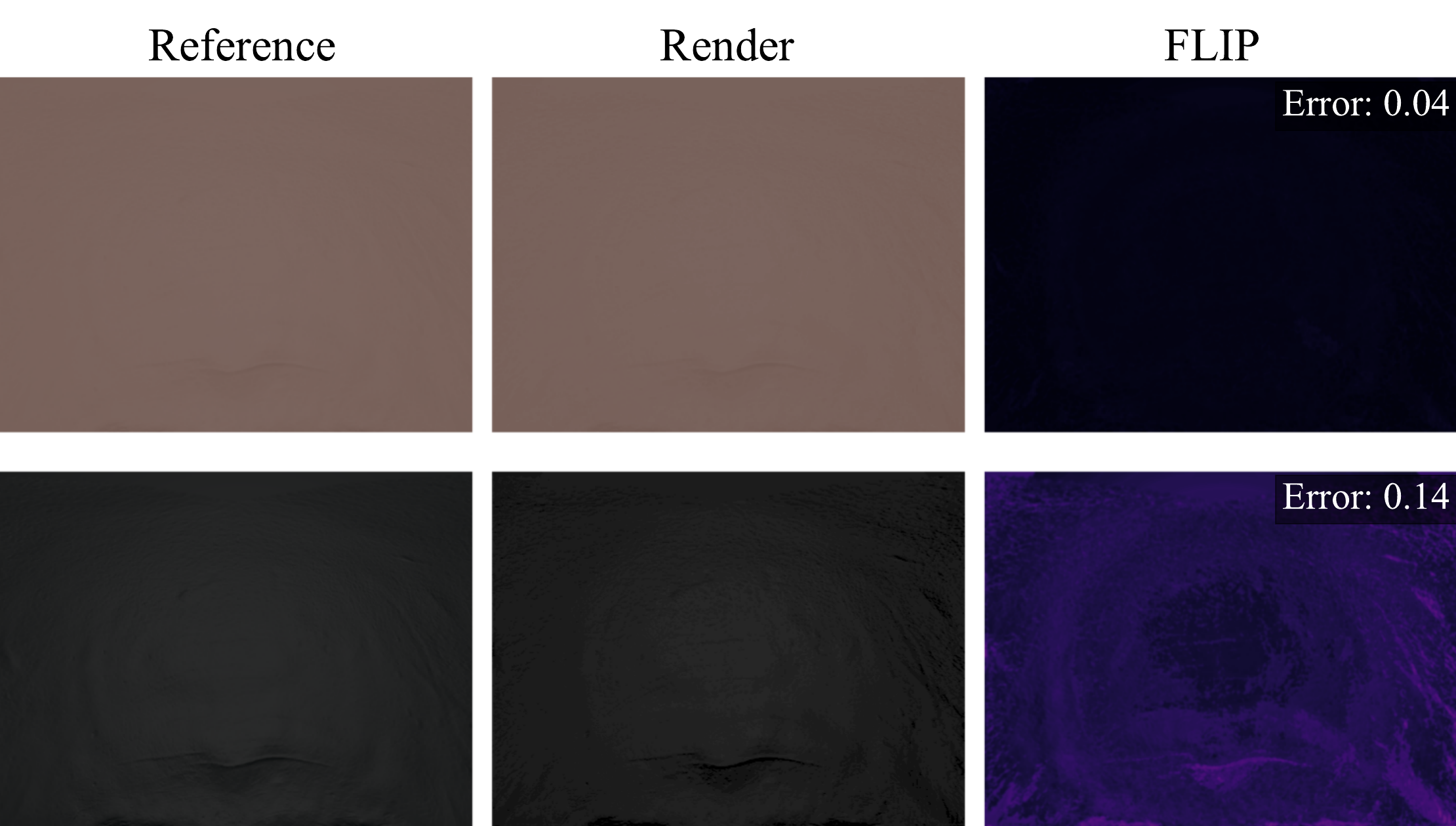}
    \caption{Validation of specular isolation. We compare the ground-truth specular contribution (Reference) against the optimized residual specular map (Render). The low error demonstrates that our albedo initialization enables the optimizer to effectively capture and absorb specular reflections within the residual map, mitigating their contamination of the base material parameters.}
    \label{fig:specular_validation}
    \vspace{-2em}
\end{figure}

\paragraph*{Specular Isolation Validation}
Our pipeline employs a residual specular map to absorb high-frequency highlights, mitigating their leakage into the albedo map. To validate this strategy, we optimize a synthetic scene where the ground-truth decomposition between albedo and specular reflection is known. Fig.~\ref{fig:specular_validation} compares the optimized residual map against the ground-truth specular contribution. The close resemblance confirms that our robust mean albedo computation effectively shields the diffuse component, allowing the residual map to accurately capture the specular response. Furthermore, this separation minimizes the risk of unmodeled specular energy corrupting other surface attributes, directly validating why our framework achieves highly stable and smooth roughness maps with suppressed artifact bleeding.

\begin{figure}
    \centering
    \includegraphics[width=\columnwidth]{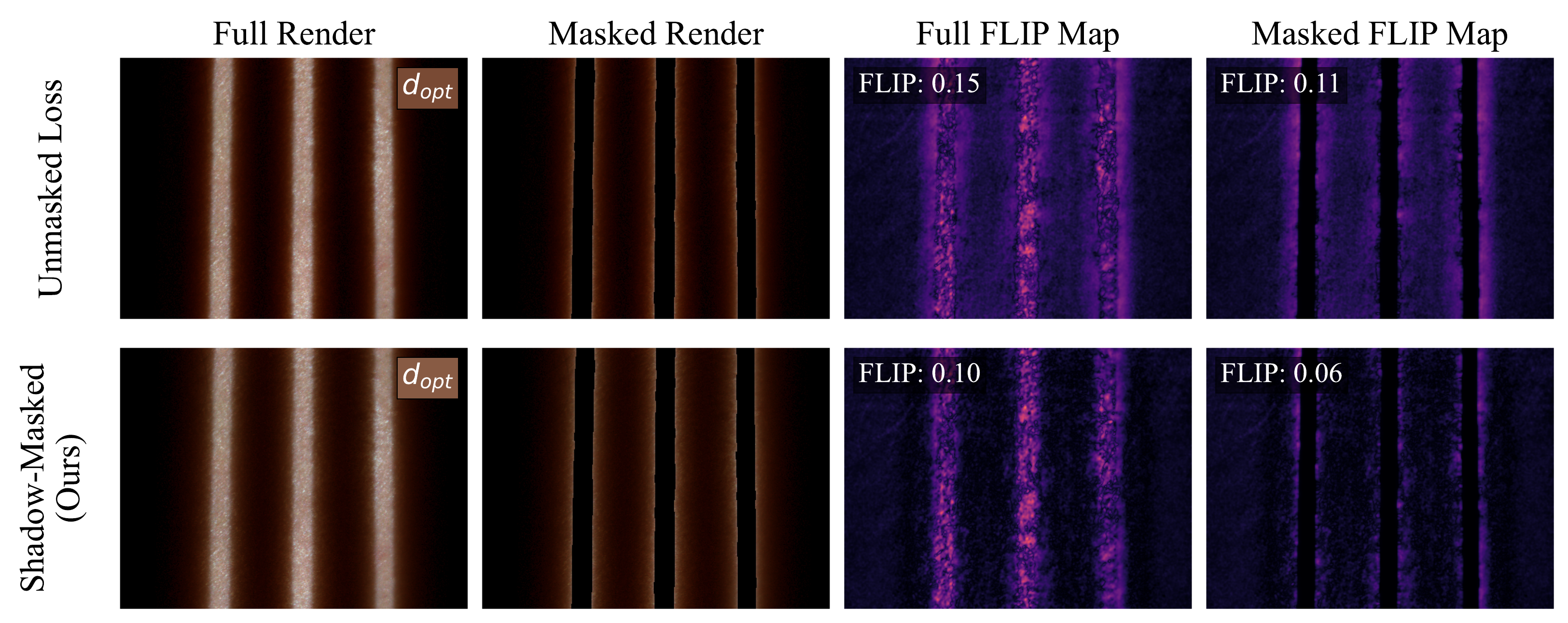}
    \caption{Ablation of Stage~3 loss supervision domain. Comparing unmasked optimization ($\mathcal{M}_{\text{opt}} = \mathbf{1}$, top) against our shadow-masked loss ($\mathcal{M}_{\text{opt}} = \mathcal{M}$, bottom). Supervising unlit regions isolates subsurface halos, preventing direct light from skewing the scattering distance $d_{\text{opt}}$ (insets) and reducing FLIP error across both the full patch and masked shadow regions.}
    \label{fig:masked_loss_validation}
    \vspace{-2em}
\end{figure}

\paragraph{Masked Loss Validation}
To isolate the contribution of our shadow-masked loss during Stage~3 ($\mathcal{M}_{\text{opt}} = \mathcal{M}$), we performed an ablation study comparing this formulation against unmasked full-image loss supervision ($\mathcal{M}_{\text{opt}} = \mathbf{1}$) under identical rendering conditions (Figure~\ref{fig:masked_loss_validation}). In the unmasked setting, incident irradiance across directly illuminated stripes is orders of magnitude higher than within shadowed regions. Consequently, backpropagation gradients are dominated by surface reflectance components (diffuse albedo and specular highlights), introducing severe gradient competition against the subtle, low-frequency subsurface diffusion halo. The optimizer attempts to compensate for subtle direct-light discrepancies by skewing the scattering distance parameter $d$.

Interestingly, while the relative spectral proportionality ($d_r > d_g > d_b$) is naturally preserved due to the lower absorption of melanin and hemoglobin at longer wavelengths, the unmasked optimization yields an artificially saturated and magnitude-shifted $d$. This magnitude shift truncates the diffusion tail along transition boundaries. In contrast, restricting the loss domain strictly to the binary shadow mask $\mathcal{M}$ decouples parameter estimation from direct surface illumination, forcing the objective function to evaluate the pure radial decay of multiple scattering. This targeted supervision eliminates parameter overcompensation, ensuring that the estimated parameter $d$ yields a reconstruction closer to the reference capture, decreasing the perceptual FLIP error from $0.11$ to $0.06$ ($45\%$ reduction) in shadowed transition regions, and from $0.15$ to $0.10$ across the full patch.

\paragraph*{Comparison with Multi-Light Acquisition}
To evaluate robustness under real-world conditions, we compare our single-light results against a multi-light setup (three synchronized projectors). This configuration allows us to introduce an additional preliminary optimization stage to the pipeline. In this new Stage 1, we leverage the multi-light observations to specifically optimize and refine the initial Sobel-filtered normal map, improving geometric consistency before proceeding with the standard material estimation steps (now shifted to Stages 2, 3, and 4).

\begin{figure}
    \centering
    \includegraphics[width=\columnwidth]{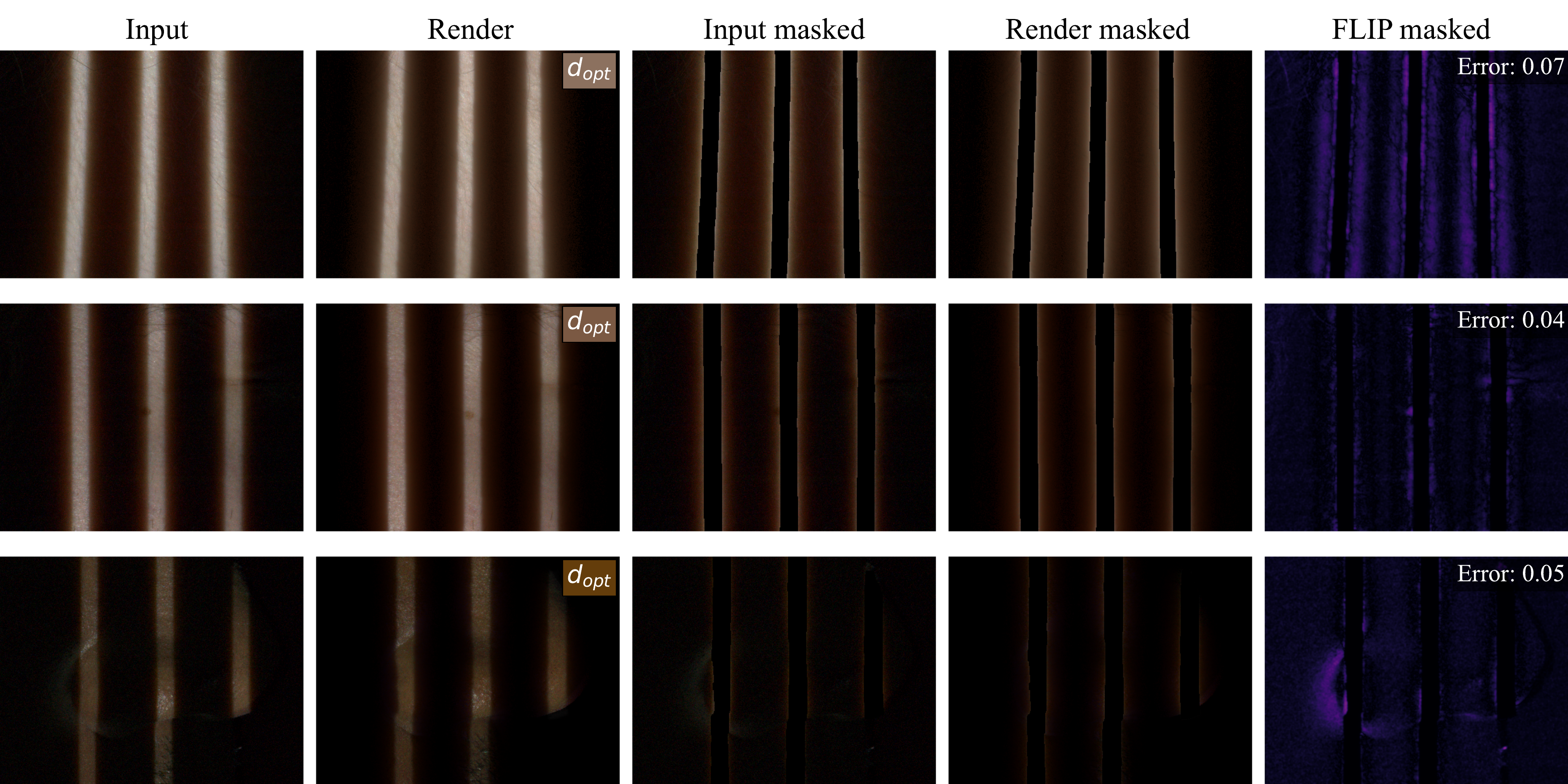}
    \caption{Multi-light results for the skin patches analyzed in Fig.~\ref{fig:single_image_fitting}. Each row shows (left to right): photo, render, masked pairs, and FLIP map. Consistency with single-view $d_{opt}$ validates our simplified setup's robustness despite additional lighting data.}
    \label{fig:multilight_comparison}
    \vspace{-1em}
\end{figure}

We compare the scattering distance maps estimated from our standard single-light pipeline against those produced using this augmented four-stage multi-light configuration. Despite the improved geometric detail provided by the refined normals, the single-light method yields scattering distances that are visually and numerically consistent (mean $\mathcal{L}_1$ difference $<2\%$). As shown in Fig.~\ref{fig:multilight_comparison}, discrepancies are minimal and confined to occluded regions.

\begin{figure}
    \centering
    \includegraphics[width=\columnwidth]{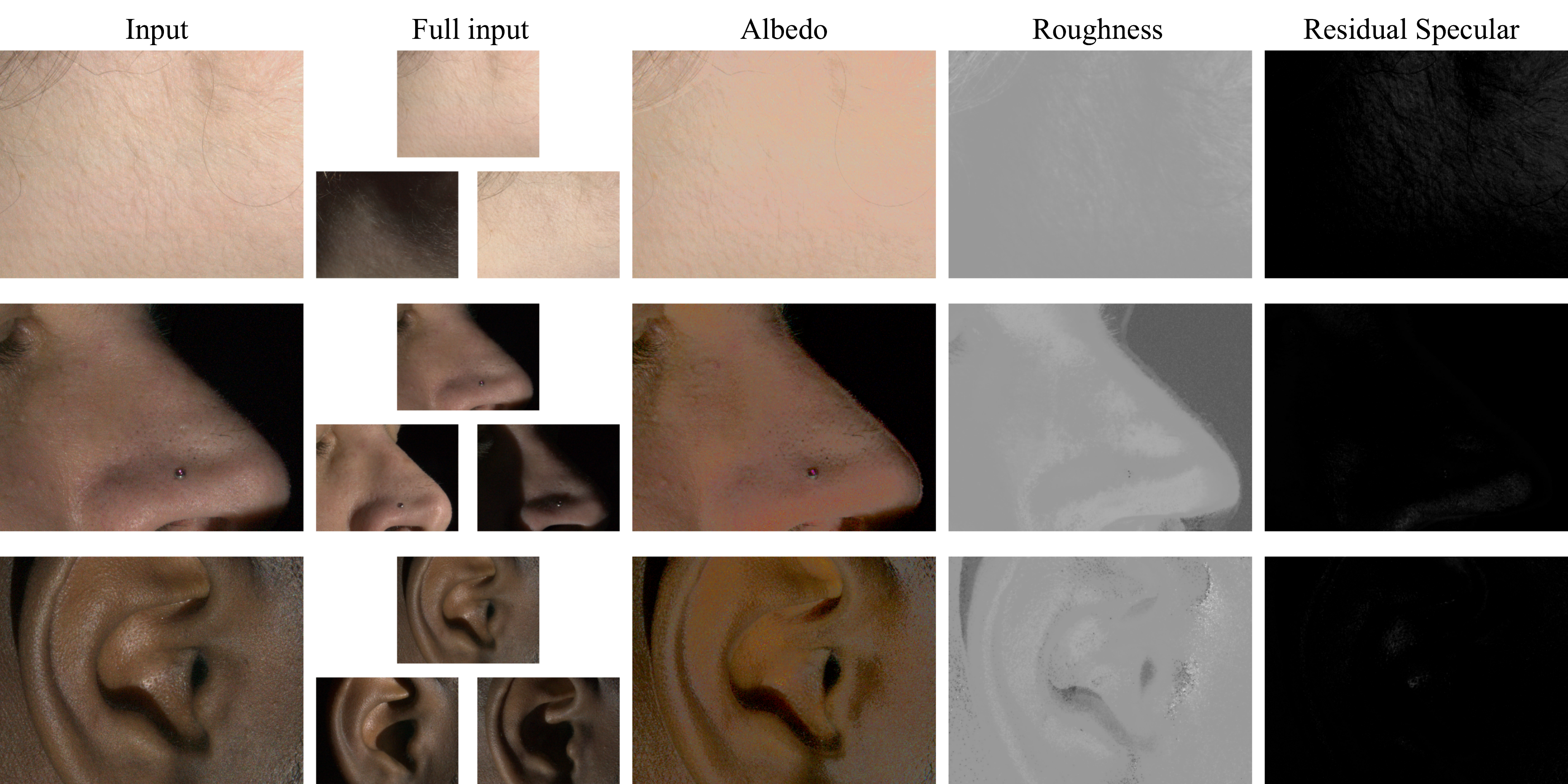}
    \caption{Optimized material stack from the multi-light setup. Extra lighting data enables a more detailed roughness map and cleaner parameter separation, reducing the information baked into the residual specular map.}
    \label{fig:multilight_stack}
    \vspace{-2em}
\end{figure}

Fig.~\ref{fig:multilight_stack} illustrates the material maps derived from the multi-light setup. By leveraging additional lighting data, this configuration improves the estimation of surface properties—particularly roughness—and better handles challenging facial regions such as the nose or ears. Although the setup is still constrained by the limited angular coverage of the three projectors, the multi-light approach provides more detailed material stacks. However, the recovered scattering distance parameters and the resulting rendering reconstructions are nearly identical to those produced by the single-view baseline. These results confirm that for the specific task of estimating subsurface scattering, our single-light, single-view approach remains both sufficient and robust.

\section{Conclusions} \label{sec:conclusions} 

We presented a lightweight differentiable inverse rendering pipeline for estimating subsurface scattering parameters of human skin directly from single-view, single-light captures. By optimizing the same parameters used in standard real-time shading models, our method establishes a direct bridge between physical acquisition and production-ready rendering pipelines, eliminating the need for complex optical setups or intermediate physical volumetric parameter conversion.

Our experiments demonstrate that physically plausible scattering distances can be estimated from minimal input data. The recovered parameters produce a realistic skin appearance across a wide range of skin tones, exhibiting consistent numerical convergence in synthetic benchmarks and high perceptual reconstruction fidelity on real-world captures. The proposed shadow-aware loss formulation improves robustness in regions dominated by multiple scattering, enabling reliable estimation even under challenging pigmentation conditions. Furthermore, our dataset includes a diverse collection of calibrated skin samples aligned with the Monk Skin Tone Scale, supporting inclusive and scalable digital human modeling.

A key result of this work is that production-ready subsurface scattering parameters can be estimated from an accessible capture setup without relying on laboratory-grade acquisition systems. Rather than requiring specialized hardware or complex measurement procedures, our approach makes the problem tractable through differentiable optimization, physically motivated parameterization, and carefully designed supervision. This substantially lowers the barrier to high-quality skin digitization while preserving compatibility with real-time rendering workflows.

The method relies on a locally homogeneous scattering assumption, and monocular depth priors. These assumptions allow the inverse problem to remain practical while still producing parameters that are physically plausible and visually effective. In the single-light setting, the ambiguity between albedo and scattering is addressed through a multi-stage optimization strategy that stabilizes parameter recovery. In thin or highly translucent structures, such as nose cartilage or the ear helix, omitting explicit single-scattering optimization may lead to slightly larger estimated scattering distances, as the model explains missing translucency through the available parameters. Our regional blending scheme constrains these estimates within realistic ranges and preserves a consistent appearance across facial regions. Similarly, omitting non-local visibility handling causes residual self-shadows to be partially absorbed into the albedo in areas of steep relief.

Future work can build on this foundation by jointly recovering additional scattering components, incorporating stronger learned priors, integrating differentiable visibility testing to resolve self-shadowing, and extending the capture protocol to less constrained environments. More flexible acquisition strategies could further improve the fidelity of reconstruction while maintaining the accessibility of the proposed setup.

By combining accessible acquisition, differentiable optimization, physically grounded constraints, and validation across diverse skin tones, our method provides a scalable path toward realistic and inclusive digital human creation for real-time applications.

\section*{Acknowledgments}
\begingroup\raggedright
This work has been supported by the Spanish Agencia Estatal de Investigación (AEI) under Project PID2024-159930OB-I00 funded by MICIU/AEI/10.13039/501100011033 and by ERDF, EU\par
\endgroup


\bibliographystyle{elsarticle-num}
\bibliography{references}

\end{document}


\let\WriteBookmarks\relax

\shorttitle{A Lightweight Differentiable Method for Real-Time Skin Rendering and Digitization}

\shortauthors{Marquez-Balduque et al.}

\twocolumn[{%
  \raggedright
  \vspace*{0.5em}
  {\fontsize{18pt}{22pt}\selectfont A Lightweight Differentiable Method for Real-Time Skin Rendering and Digitization\par}
  \vspace*{0.4em}
  {\large Supplementary Material\par}
  \vspace*{0.6em}
  {\normalsize Mario Marquez-Balduque\textsuperscript{1}, Carlos Rodr\'iguez-Pardo\textsuperscript{2,3,4}, Jorge Lopez-Moreno\textsuperscript{1}\par}
  \vspace*{0.4em}
  {\small \textsuperscript{1}Universidad Rey Juan Carlos, Spain\par
  \textsuperscript{2}Politecnico di Milano, Milan, Italy\par
  \textsuperscript{3}Euro-Mediterranean Center on Climate Change (CMCC), Lecce, Italy\par
  \textsuperscript{4}RFF-CMCC European Institute on Economics and the Environment (EIEE), Milan, Italy\par}
  \vspace*{1.8em}
}]

\section{Introduction}
This supplementary document provides extended validation, dataset analysis, and core implementation details supporting the main manuscript, organized as follows:

\begin{itemize}
    \item \textbf{Map Evolution Across Optimization Stages (Section~\ref{sec:stage_evolution}):} Illustrates the step-by-step visual progression of active and frozen maps across optimization stages, justifying our staged schedule in decoupling surface reflections and subsurface diffusion.
    \item \textbf{Tilted View Validation (Section~\ref{sec:tilted_view_validation}):} Demonstrates the angular consistency and physical grounding of our 3D depth-aware optimization under varying camera orientations.
    \item \textbf{Dataset across the Monk Scale (Section~\ref{sec:monk_dataset}):} Presents comprehensive synthetic validation across the full Monk Skin Tone Scale by isolating albedo variations under fixed scattering parameters, and evaluates real-world participant distributions by cross-referencing subjective self-identification against our objective color-matching algorithm via a unified confusion matrix.
    \item \textbf{Replicability and Algorithmic Details (Section~\ref{sec:reproducibility}):} Provides concrete pipeline blueprints, including RAW color calibration via Ridge Regression (Section~\ref{subsec:color_calib}), robust albedo initialization (Section~\ref{subsec:albedo_init}), and our GPU shader (Section~\ref{subsec:shader_impl}) and differentiable tensorized implementations (Section~\ref{subsec:tensorized_impl}).
\end{itemize}

\section{Map Progression Across Optimization Stages}
\label{sec:stage_evolution}

\begin{figure}
    \centering
    \includegraphics[width=\columnwidth]{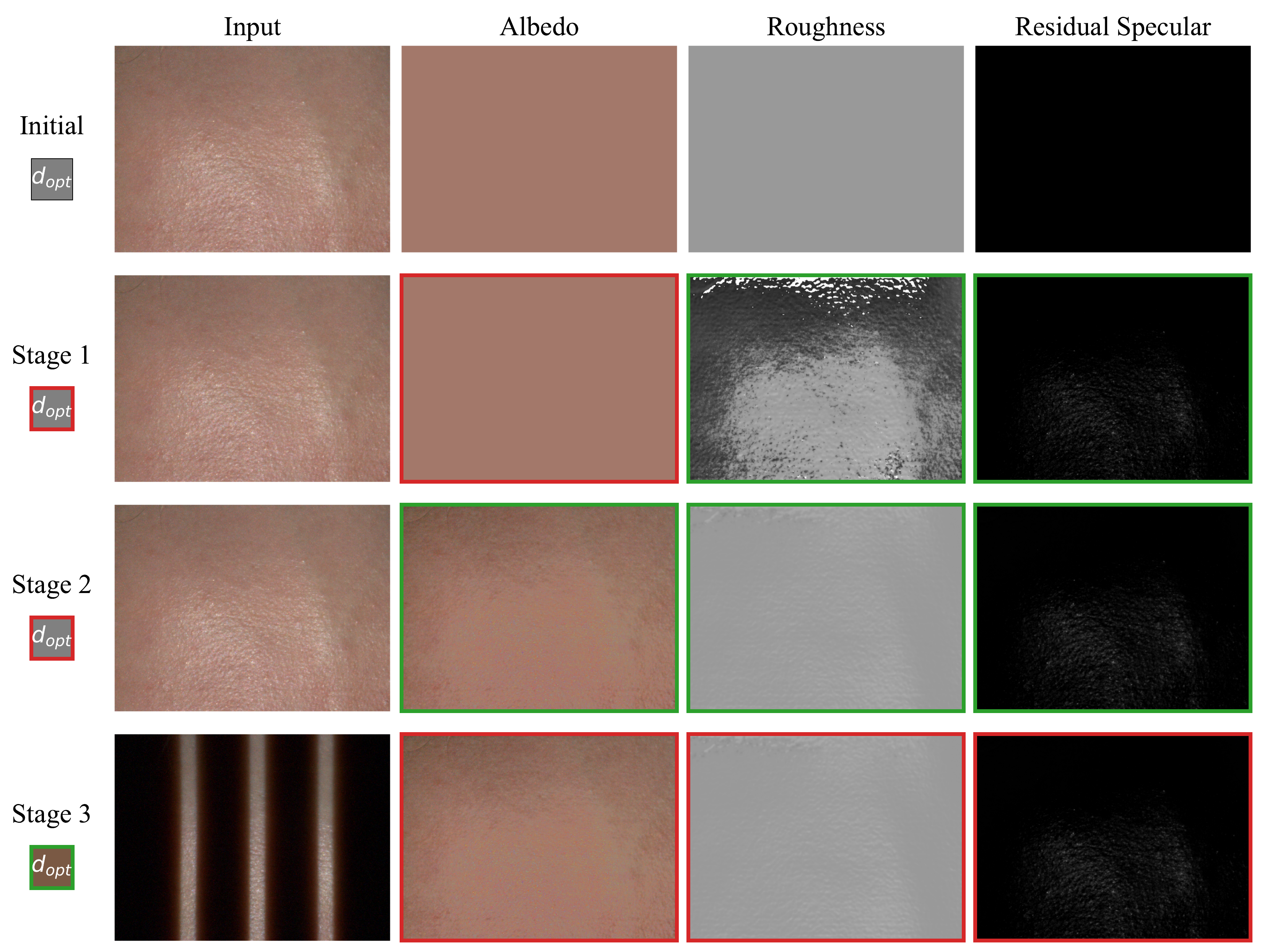}
    \vspace{-1em}
    \caption{Map progression across optimization stages. Green and red denote active and frozen maps, respectively. Stage 1 fits global reflection, capturing unmodeled specular highlights into the residual map while shielding initialized albedo. Stage 2 unfreezes albedo to absorb diffuse spatial details, enabling the $\mathcal{L}_2$ prior to smooth roughness while preserving the specular residual. Stage 3 optimizes scattering distance over shadow-masked regions (Input) with surface maps frozen.}
    \label{fig:stage_evolution}
    \vspace{-1em}
\end{figure}

To complement the optimization methodology described in the main manuscript, Figure~\ref{fig:stage_evolution} illustrates the step-by-step visual progression of active and frozen material maps across our three optimization stages.

During Stage~1, freezing the albedo $\mathbf{A}_T$ forces the reconstruction loss to temporarily drive the active roughness map ($\boldsymbol{\rho}_T$) to fit low-frequency diffuse shading gradients. In Stage~2, unfreezing $\mathbf{A}_T$ provides the direct path to resolve these diffuse reconstruction errors. As the albedo rapidly absorbs the spatial shading details, the reconstruction gradient on other maps vanishes. Consequently, the $\mathcal{L}_2$ roughness prior smoothly regularizes $\boldsymbol{\rho}_T$ while leaving the residual specular map virtually unperturbed, matching its role as a buffer designed for unmodeled specular highlights rather than diffuse light transport.

\section{Tilted View Validation}
\label{sec:tilted_view_validation}

\begin{figure}
    \centering
    \includegraphics[width=\columnwidth]{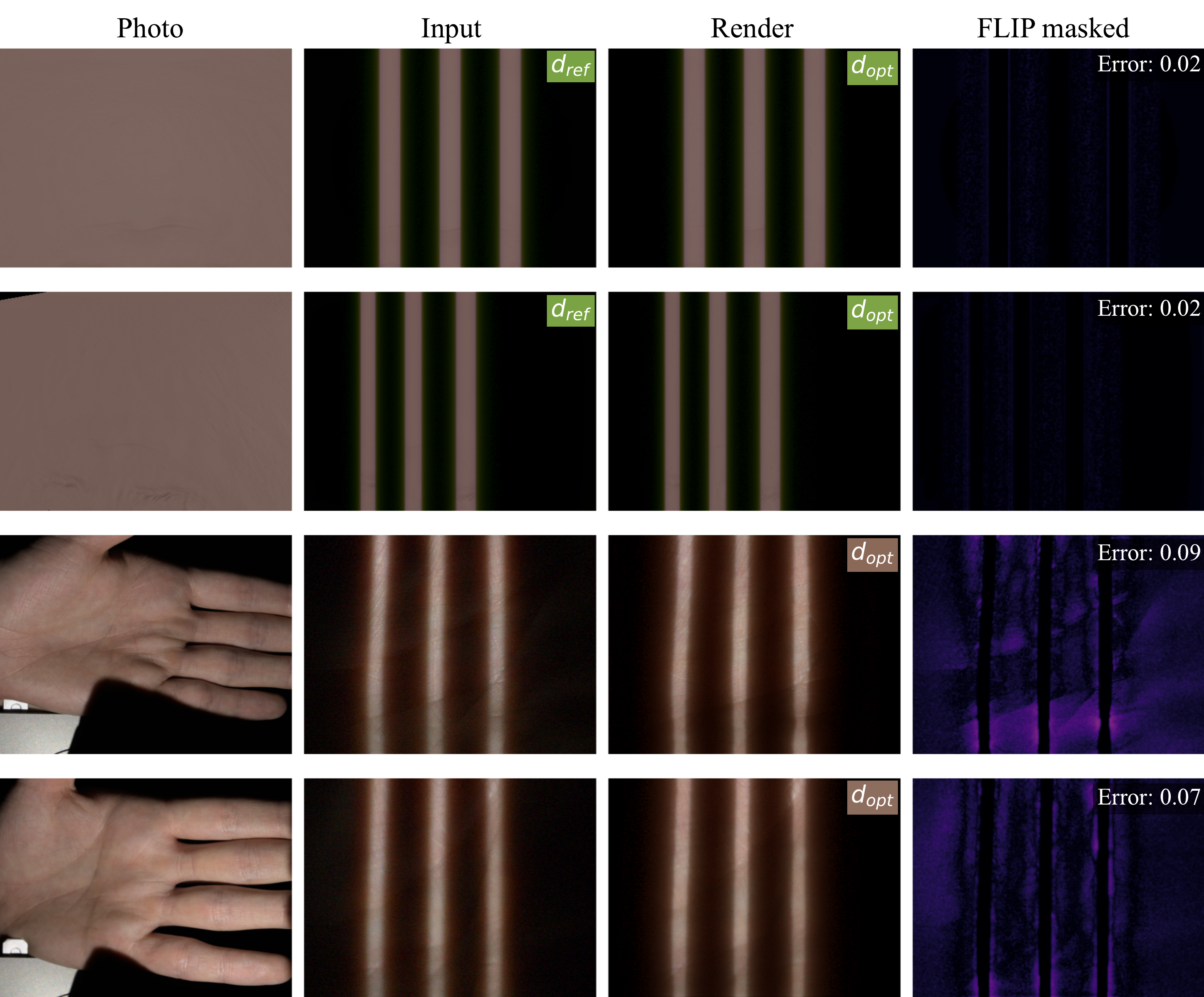}
    \vspace{-1em}
    \caption{Tilted view validation. In synthetic cases (top), $d_{\text{opt}}$ closely tracks $d_{\text{ref}}$. For real captures (bottom), scattering distances remain stable across views, demonstrating that 3D depth-aware computations prevent parameter shrinkage caused by perspective foreshortening.}
    \label{fig:tilted_validation}
    \vspace{-2em}
\end{figure}

To verify that our frontal-view optimization yields physically grounded parameters rather than view-dependent overfitted solutions, we perform tilted-view cross-validation on both synthetic and real data. For synthetic targets, we optimize parameters independently from frontal and novel tilted views, verifying that the estimated scattering distance ($d_{\text{opt}}$) consistently matches the ground truth ($d_{\text{ref}}$). For real targets, we evaluate acquisitions of identical skin regions under varying camera angles. As shown in Fig.~\ref{fig:tilted_validation}, estimated scattering distances remain invariant across orientations, yielding minimal reconstruction error.

Our pipeline avoids 2D projection approximations by executing scattering calculations directly in 3D view space using the estimated geometric depth. This integration ensures that the Screen-Space Subsurface Scattering (SSSS) algorithm naturally accounts for surface curvature and perspective foreshortening. By computing physical distances rather than simple 2D pixel offsets, the model recovers orientation-invariant scattering profiles.

\section{Dataset across the Monk Scale}
\label{sec:monk_dataset}

\begin{figure*}
    \centering
    \subfloat{%
        \includegraphics[width=0.45\linewidth]{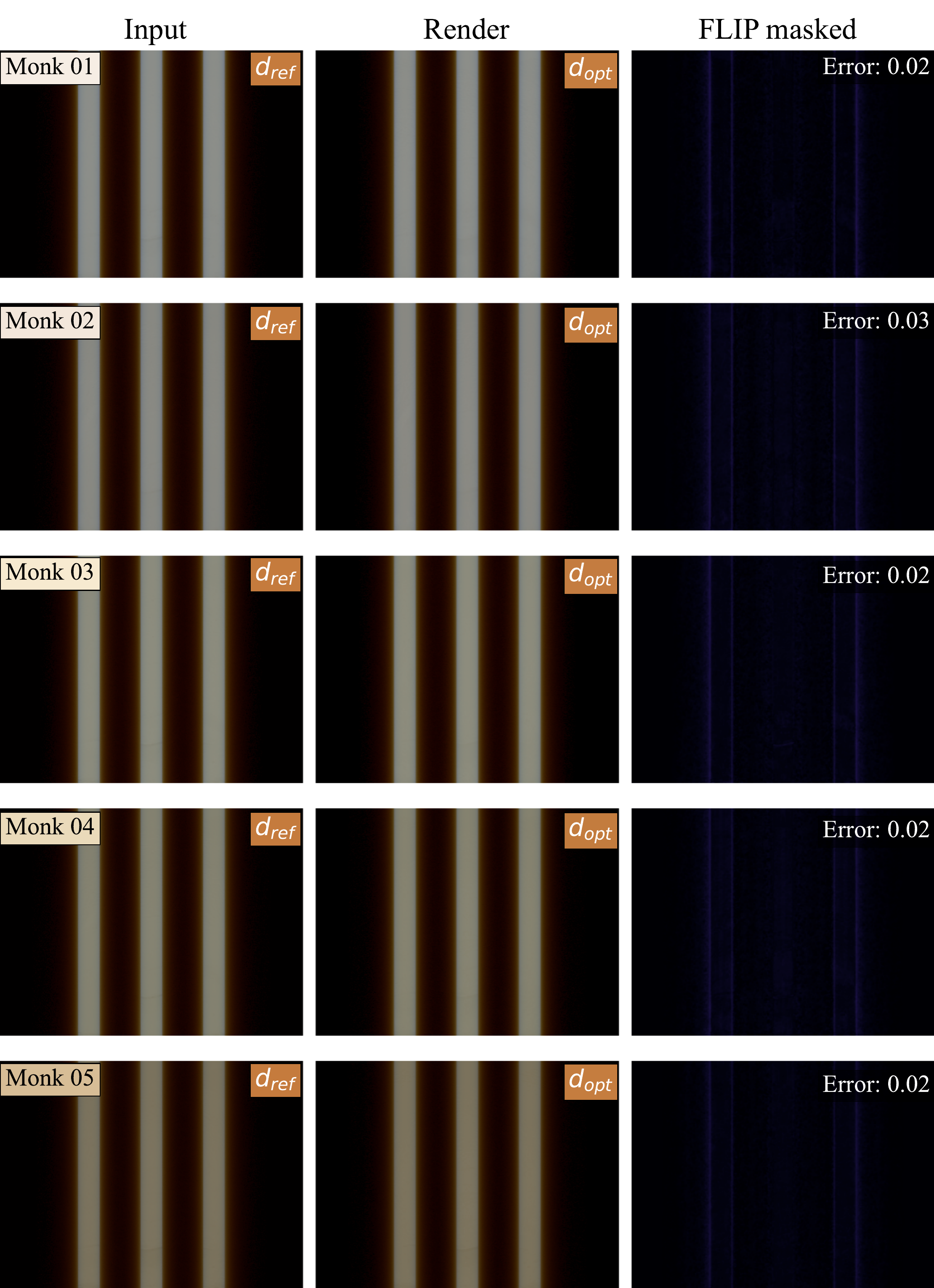}}
    \hspace{0.2mm}
    \subfloat{%
        \includegraphics[width=0.45\linewidth]{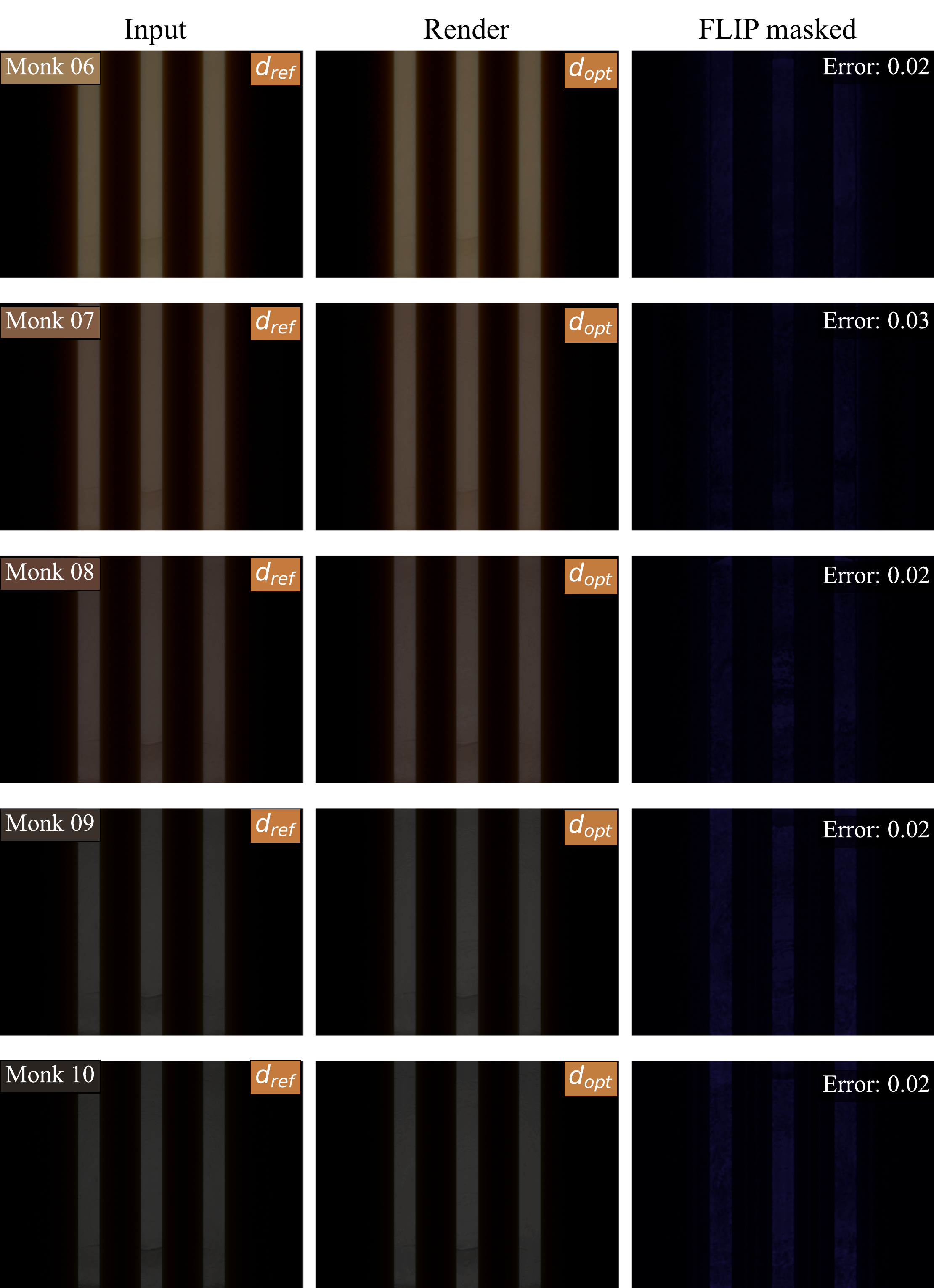}}
    \caption{%
    Full synthetic validation across the Monk Skin Tone Scale. Per tone (left to right): input, optimized reconstruction, and FLIP error map. Inset swatches denote $d_{\text{ref}}$ (known reference value) and $d_{\text{opt}}$ (optimized value). Low perceptual errors confirm estimation invariance across diverse skin tones.
    }
    \label{fig:synth_monk_full}
    \vspace{-2em}
\end{figure*}

\begin{figure}
    \centering
    \includegraphics[width=\columnwidth]{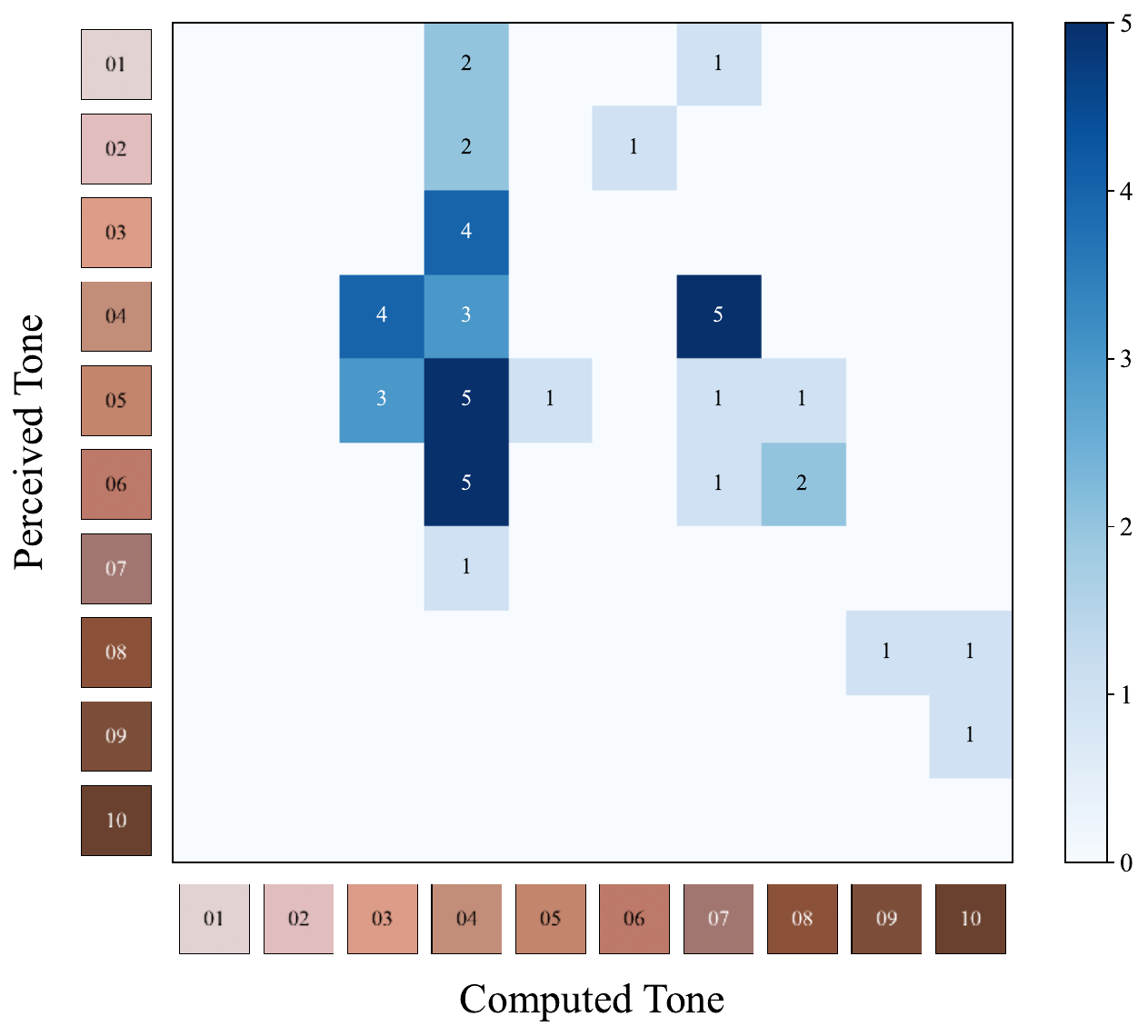}
    \vspace{-1em}
    \caption{Unified confusion matrix cross-referencing participant self-identified (Perceived) skin tones against our automatic appearance-based color-matching algorithm (Computed) across the Monk Skin Tone Scale ($N=45$). The clustering behavior highlights systematic discrepancies between subjective human self-assessment and objective physical albedo extraction, driven by visual perception biases and the uncalibrated nature of the original scale.}
    \label{fig:monk_confusion_matrix}
\end{figure}

Figure~\ref{fig:synth_monk_full} presents the full synthetic validation workflow, confirming optimization stability across all ten Monk classes. By fixing the ground-truth scattering distance and varying only the target albedo, we verify that our method decouples scattering parameters independently of surface reflectance.

Figure~\ref{fig:monk_confusion_matrix} cross-references participant skin tone self-assessments against our automatically computed baseline using a unified confusion matrix ($N=45$). As expected from a Mediterranean–Caucasian population, the sample distribution concentrates around intermediate Monk tones (3–6), highlighting the representative challenges of real-world skin digitization. For automated classification, the mean skin color is extracted from the neutral frontal-light capture to eliminate specular and shadow contamination, and subsequently mapped via CIE LAB distance to uniformly lit reference exemplars from the MST dataset.

Discrepancies between computed tones and self-assessments in the matrix reveal inherent human perception biases and visual ambiguities when isolating pure albedo from overall lightness under casual observation. Furthermore, the original Monk scale is inherently perceptual, with reference colors defined under uncalibrated lighting conditions. Consequently, standardized computational environments and objective physical optimization pipelines are essential for consistent, illumination-invariant tone classification.

\section{Reproducibility and Algorithmic Details}
\label{sec:reproducibility}

\subsection{RAW Processing and Color Calibration}
\label{subsec:color_calib}

\begin{figure}
    \centering
    \includegraphics[width=0.9\columnwidth]{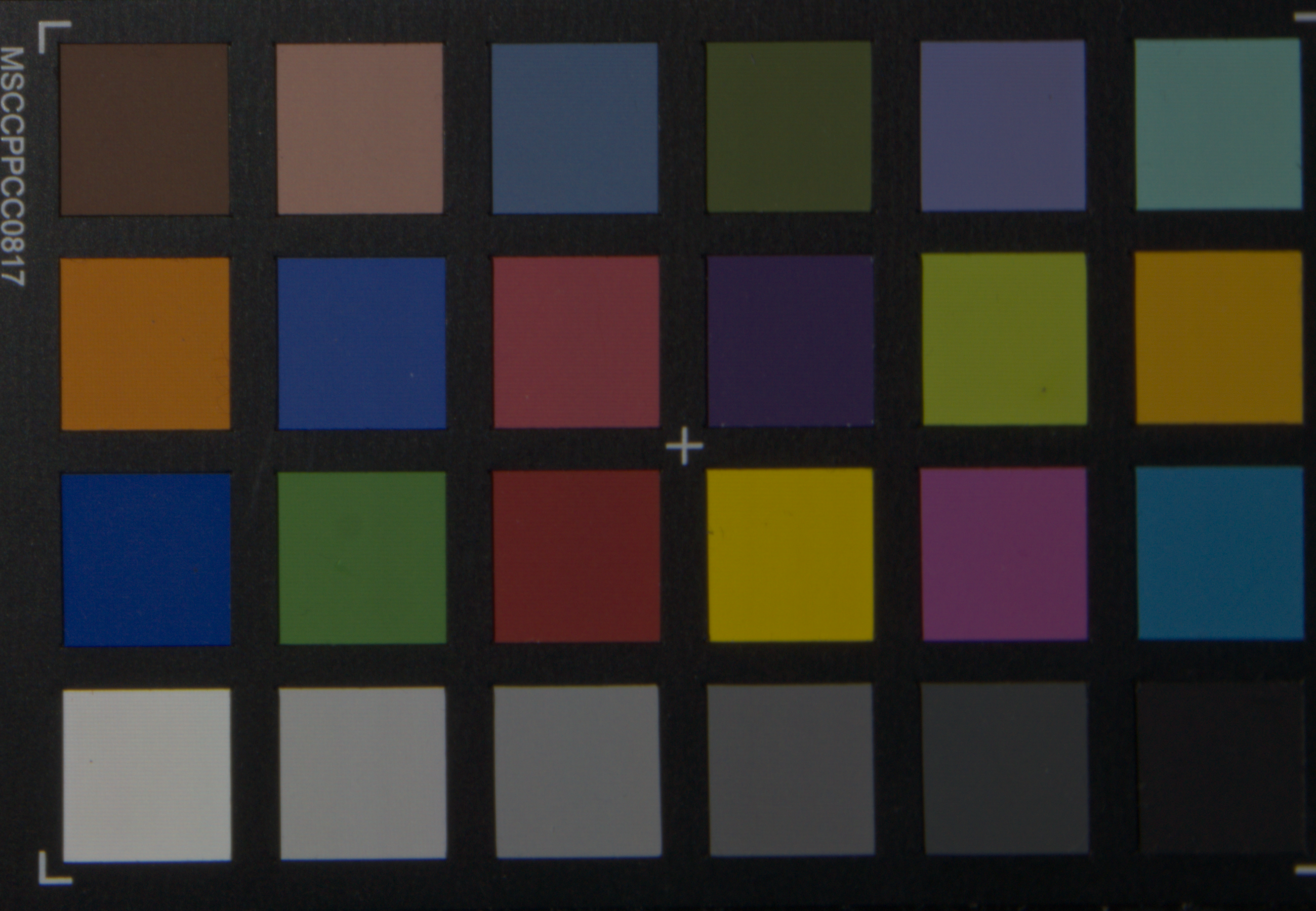}
    \caption{Physical color calibration setup under the projector's flat-field white illumination. The 24 patch coordinates are automatically extracted to fit the $3\times3$ CCM.}
    \label{fig:calibration_target}
    \vspace{-1em}
\end{figure}

RAW images ($6020 \times 4024$\,px) are decoded via \textit{rawpy}, bypassing all non-linear ISP operations to enforce a linear gamma ($\gamma = 1.0$) and native camera color space. Normalized 16-bit float intensities ($[0,1]$) are downsampled using bicubic interpolation to $1204 \times 804$\,px to preserve global spatial context, and center-cropped to a final operating resolution of $480 \times 360$\,px.

To map the linear RAW space to standard D65 linear sRGB, we extract the mean RAW intensity of the 24 patches from an X-Rite ColorChecker Classic target (Fig.~\ref{fig:calibration_target}) to build the measurement matrix $\mathbf{X}$. Target sRGB reference values are linearized via gamma decoding ($\gamma = 2.2$) to form matrix $\mathbf{Y}$. 

To suppress noise amplification, the $3 \times 3$ Color Correction Matrix (CCM) $\mathbf{C}$ is computed channel-wise using $\mathcal{L}_2$-regularized Ridge Regression. For each target channel column $\mathbf{y}_c$ ($c \in \{r, g, b\}$), the optimization yields:
\begin{equation}
\mathbf{c}_c = \left(\mathbf{X}^T\mathbf{X} + \alpha \mathbf{I}\right)^{-1}\mathbf{X}^T\mathbf{y}_c
\end{equation}
where $\alpha = 10^{-3}$ and $\mathbf{I}$ is the identity matrix. The final matrix $\mathbf{C} = [\mathbf{c}_r^T; \mathbf{c}_g^T; \mathbf{c}_b^T]$ maps the image via $\mathbf{I}_{\text{linear}} = \mathbf{I}_{\text{raw}} \mathbf{C}^T$. Inverse gamma encoding ($1/2.2$) is applied strictly for display previews.

\subsection{Albedo Initialization}
\label{subsec:albedo_init}

\begin{algorithm}
\caption{Adaptive Albedo Initialization}
\label{alg:albedo_initialization}
\footnotesize
\begin{algorithmic}[1]
\Require 
    \Statex $\mathbf{I}$ : Normalized RGB input image
\Ensure 
    \Statex $\mathbf{c}_{\text{init}}$ : Robust initialization RGB color vector

\smallskip
\Function{ExtractAdaptiveSkinMask}{$\mathbf{I}$}
    \LineComment{\textbf{Step 1: Color Space Transformation}}
    \State $\mathbf{I}_{\text{lab}} \gets \text{ConvertRGBToLab}(\mathbf{I})$
    \Statex
    \LineComment{\textbf{Step 2: Background Discarding \& Feature Extraction}}
    \State $\mathcal{M}_{\text{valid}} \gets \{ (x,y) \mid \mathbf{I}_{\text{lab}}[x,y,L] > 20 \}$
    \State $\mathbf{P} \gets \mathbf{I}_{\text{lab}}[\mathcal{M}_{\text{valid}}, (a, b)]$
    \Statex
    \LineComment{\textbf{Step 3: Chromatic Feature Clustering}}
    \State $\mathcal{C}_1, \mathcal{C}_2 \gets \text{ExecuteKMeans}(\mathbf{P}, K=2)$
    \Statex
    \LineComment{\textbf{Step 4: Skin Cluster Identification \& Mapping}}
    \State $idx_{\text{skin}} \gets \text{argmax}(\text{GetCentroid}(\mathcal{C}_k)[a])$
    \State $\mathcal{M}_{\text{raw}} \gets \text{MapToImageSpace}(\mathcal{C}_{idx_{\text{skin}}})$
    \Statex
    \LineComment{\textbf{Step 5: Luminance Filtering (Specularities \& Shadows)}}
    \State $L_{\text{skin}} \gets \mathbf{I}_{\text{lab}}[\mathcal{M}_{\text{raw}}, L]$
    \State $T_{\text{bright}} \gets \text{ComputePercentile}(L_{\text{skin}}, 90)$
    \State $T_{\text{dark}} \gets \text{ComputePercentile}(L_{\text{skin}}, 5)$
    \State $\mathcal{M}_{\text{final}} \gets \mathcal{M}_{\text{raw}} \land (T_{\text{dark}} < \mathbf{I}_{\text{lab}}[:,:,L] < T_{\text{bright}})$
    \Statex
    \State \Return $\mathcal{M}_{\text{final}}$
\EndFunction

\smallskip
\Function{InitializeAlbedo}{$\mathbf{I}$}
    \LineComment{\textbf{Step 1: Adaptive Mask Generation}}
    \State $\mathcal{M} \gets \text{\textsc{ExtractAdaptiveSkinMask}}(\mathbf{I})$
    \Statex
    \LineComment{\textbf{Step 2: Active Pure-Skin Pixel Gathering}}
    \State $\mathbf{P}_{\text{skin}} \gets \mathbf{I}[\mathcal{M}]$
    \State $\mathbf{c}_{\text{init}} \gets \begin{bmatrix} 0 & 0 & 0 \end{bmatrix}^T$
    \Statex
    \LineComment{\textbf{Step 3: Channel-wise Statistical Sigma Clipping}}
    \For{channel $c \in \{R, G, B\}$}
        \State $\mathbf{v} \gets \mathbf{P}_{\text{skin}}[:, c]$
        \State $\mu_c, \sigma_c \gets \text{ComputeMean}(\mathbf{v}), \text{ComputeStdDev}(\mathbf{v})$
        \Statex
        \LineComment{\textbf{Apply $2\sigma$ Outlier Filter}}
        \State $\mathbf{v}_{\text{clipped}} \gets \{ x \in \mathbf{v} \mid |x - \mu_c| < 2\sigma_c \}$
        \State $\mathbf{c}_{\text{init}}[c] \gets \text{ComputeMean}(\mathbf{v}_{\text{clipped}})$
    \EndFor
    \Statex
    \State \Return $\mathbf{c}_{\text{init}}$
\EndFunction
\end{algorithmic}
\end{algorithm}
\begin{algorithm}
\caption{Burley Multiple Scattering Shader}
\label{alg:burley_shader}
\footnotesize
\begin{algorithmic}[1]
\Require 
    \Statex $\mathbf{u}$ : 2D screen-space texture coordinates $[u, v]$
    \Statex $\mathbf{S}$ : Screen-space scattering map storing spatially blended $d_{\text{opt}}$ values
    \Statex $\mathbf{D}$ : Screen-space depth texture map
    \Statex $\mathbf{A}$ : Screen-space surface albedo texture map
    \Statex $\mathbf{L}$ : Screen-space diffuse irradiance texture map
    \Statex $\mathbf{N}$ : Sample count $\in [2, 128]$
    \Statex $\mathbf{samples}$ : Array of $\text{vec2}(\theta, r)$ (Fibonacci angles, Stratified Burley Inverse CCDF normalized radii (Eq.~2))
    \Statex $\mathbf{M_{\text{proj}}}, \mathbf{M_{\text{proj}}^{-1}}$ : Camera projection matrix and its inverse
\Ensure 
    \Statex $\mathbf{I}$ : Scattered diffuse radiance vector for the target fragment

\smallskip
\Procedure{EvaluateMultipleScattering}{}
    \LineComment{\textbf{Subsurface Scattering Setup}}
    \State $w_{\text{scale}} \gets 0.01, \quad \epsilon \gets 10^{-6}$
    \State $d \gets S(u).rgb \times w_{\text{scale}}$
    \State $r_{\text{max}} \gets \max(d_r, d_g, d_b)$
    \State $I \gets \begin{bmatrix} 0.0 & 0.0 & 0.0 \end{bmatrix}^T$
    \State $K \gets \begin{bmatrix} 0.0 & 0.0 & 0.0 \end{bmatrix}^T$
    \Statex
    \LineComment{\textbf{Step 1: Target View-Space 3D Position Reconstruction}}
    \State $x \gets \text{Unproject}(u, D(u), M_{\text{proj}}^{-1})$
    \Statex
    \LineComment{\textbf{Step 2: Generate noise pattern to avoid $\theta_i$ periodicity}}
    \State $\text{jitter} \gets 2.0 \times \pi \times \text{Noise}(u)$
    \Statex
    \For{$i \gets 0$ \textbf{to} $N - 1$}
        \LineComment{\textbf{Step 3: Fibonacci \& Burley Profile Importance Sampling}}
        \State $\theta_i \gets \text{samples}[i].x + \text{jitter}$
        \State $r_i \gets \text{samples}[i].y \times r_{\text{max}}$
        \State $\text{offset} \gets \begin{bmatrix} r_i \cos\theta_i & r_i \sin\theta_i & 0.0 \end{bmatrix}^T$
        \Statex
        \LineComment{\textbf{Step 4: Neighbor Screen Projection}}
        \State $u_i \gets \text{Project}(x + \text{offset}, M_{\text{proj}})$
        \Statex
        \LineComment{\textbf{Step 5: Neighbor View-Space Position Reconstruction}}
        \State $D_i \gets D(u_i)$
        \State $x_i \gets \text{Unproject}(u_i, D_i, M_{\text{proj}}^{-1})$
        \Statex
        \LineComment{\textbf{Step 6: Subsurface Diffusion Weights (Eq.~3)}}
        \State $s_i \gets \max(\|x - x_i\|_2, \epsilon)$
        \State $rRr \gets s_i \times R(s_i, d)$
        \State $pr \gets r_i \times R(r_i, r_{\text{max}})$
        \State $k_i \gets rRr / pr$
        \Statex
        \LineComment{\textbf{Step 7: Irradiance Gathering}}
        \State $L_i \gets L(u_i).rgb$
        \State $I \gets I + (k_i \times L_i)$
        \State $K \gets K + k_i$
    \EndFor
    \Statex
    \LineComment{\textbf{Step 8: Normalized weighted Integration (Eq.~4)}}
    \State $I \gets A(u).rgb \times \left( I / \max(K, \epsilon) \right)$
    \State \Return $I$
\EndProcedure
\end{algorithmic}
\end{algorithm}

\begin{algorithm}
\caption{Burley Multiple Scattering Tensorized}
\label{alg:burley_tensorized}
\footnotesize
\begin{algorithmic}[1]
\Require 
    \Statex $\mathbf{U, V}$ : Grid coordinate tensors. $(H, W)$ in $[-1, 1]$
    \Statex $\mathbf{S}_T$ : Homogeneous scattering distance tensor. $(H, W, 3)$
    \Statex $\mathbf{D}_T$ : Depth buffer tensor map. $(H, W)$
    \Statex $\mathbf{A}_T$ : Surface albedo tensor map. $(H, W, 3)$
    \Statex $\mathbf{L}_T$ : Diffuse irradiance tensor map. $(1, 3, H, W)$
    \Statex $N$ : Sample count $\in [2, 128]$
    \Statex $\mathbf{samples}_{\theta}$ : Fibonacci angles tensor. $(N)$
    \Statex $\mathbf{samples}_r$ : Stratified Burley Inverse CCDF radii tensor. $(N)$ (Eq.~2)
    \Statex $\boldsymbol{\Phi}_c$ : Camera intrinsic parameters $\{z_{\text{near}}, z_{\text{far}}, w_{\text{sensor}}, f_{\text{focal}}\}$
\Ensure 
    \Statex $\mathbf{I}_T$ : Scattered diffuse radiance tensor. $(H, W, 3)$

\smallskip
\Procedure{EvaluateMultipleScattering}{}
    \LineComment{\textbf{Subsurface Scattering Setup}}
    \State $w_{\text{scale}} \gets 0.01, \quad \epsilon \gets 10^{-6}$
    \State $d \gets S_T \times w_{\text{scale}}$ \Comment{$(H, W, 3)$}
    \State $r_{\text{max}} \gets d.\max()$
    \Statex
    \LineComment{\textbf{Step 1: Target View-Space 3D Position Reconstruction}}
    \State $z \gets \text{LinearizeDepth}(D_T, \Phi_c)$ \Comment{$(H, W)$ (Eq.~5)}
    \State $x_v \gets \text{Unproject}(U, z, \Phi_c)$ \Comment{$(H, W)$ (Eq.~6)}
    \State $y_v \gets \text{Unproject}(V, z, \Phi_c)$ \Comment{$(H, W)$ (Eq.~6)}
    \State $P_v \gets \begin{bmatrix} x_v & y_v & -z \end{bmatrix}^T$ \Comment{$(H, W, 3)$}
    \Statex
    \LineComment{\textbf{Step 2: Generate noise pattern to avoid $\theta$ periodicity}}
    \State $\text{jitter} \gets 2.0 \times \pi \times \text{Random}(1, H, W)$ \Comment{$(1, H, W)$}
    \Statex
    \LineComment{\textbf{Step 3: Fibonacci \& Burley Profile Importance Sampling}}
    \State $\theta \gets samples_{\theta}.\text{view}(N, 1, 1) + \text{jitter}$ \Comment{$(N, H, W)$}
    \State $r \gets samples_r.\text{view}(N, 1, 1) \times r_{\text{max}}$ \Comment{$(N, 1, 1)$}
    \State $x_o \gets \cos\theta \odot r$ \Comment{$(N, H, W)$}
    \State $y_o \gets \sin\theta \odot r$ \Comment{$(N, H, W)$}
    \Statex
    \LineComment{\textbf{Step 4: Neighbor Screen Projection}}
    \State $U_s \gets \text{Project}(x_v + x_o, z, \Phi_c)$ \Comment{$(N, H, W)$ (Eq.~7)}
    \State $V_s \gets \text{Project}(y_v + y_o, z, \Phi_c)$ \Comment{$(N, H, W)$ (Eq.~7)}
    \State $G \gets \text{Stack}(U_s, V_s, \text{dim}=-1)$ \Comment{$(N, H, W, 2)$}
    \Statex
    \LineComment{\textbf{Step 5: Neighbor View-Space Position Reconstruction}}
    \State $D_s \gets \text{GridSample}(D_T, G)$ \Comment{$(N, H, W)$}
    \State $z_s \gets \text{LinearizeDepth}(D_s, \Phi_c)$ \Comment{$(N, H, W)$ (Eq.~5)}
    \State $x_s \gets \text{Unproject}(U_s, z_s, \Phi_c)$ \Comment{$(H, W)$ (Eq.~6)}
    \State $y_s \gets \text{Unproject}(V_s, z_s, \Phi_c)$ \Comment{$(H, W)$ (Eq.~6)}
    \State $P_s \gets \begin{bmatrix} x_s & y_s & -z_s \end{bmatrix}^T$ \Comment{$(N, H, W, 3)$}
    \Statex
    \LineComment{\textbf{Step 6: Subsurface Diffusion Weights (Eq.~3)}}
    \State $s \gets \max(\|P_v - P_s\|_2, \epsilon)$ \Comment{$(N, H, W, 1)$}
    \State $rRr \gets s \odot R(s, d)$ \Comment{$(N, H, W, 3)$}
    \State $pr \gets r \odot R(r, r_{\text{max}})$ \Comment{$(N, 1, 1)$}
    \State $k \gets rRr \oslash pr$ \Comment{$(N, H, W, 3)$}
    \Statex
    \LineComment{\textbf{Step 7: Irradiance Gathering \& Dimension Collapse}}
    \State $L_T \gets L_T.\text{expand}(N)$ \Comment{$(N, 3, H, W)$}
    \State $L_s \gets \text{GridSample}(L_T, G)$ \Comment{$(N, 3, H, W)$}
    \State $L_s \gets L_s.\text{permute}(0, 2, 3, 1)$ \Comment{$(N, H, W, 3)$}
    \State $I_T \gets \sum_{\text{dim}=0} (k \odot L_s)$ \Comment{$(H, W, 3)$}
    \State $K_T \gets \sum_{\text{dim}=0} k$ \Comment{$(H, W, 3)$}
    \Statex
    \LineComment{\textbf{Step 8: Normalized weighted Integration (Eq.~4)}}
    \State $I_T \gets A_T \odot \left( I_T \oslash \max(K_T, \epsilon) \right)$ \Comment{$(H, W, 3)$}
    \State \Return $I_T$
\EndProcedure
\end{algorithmic}
\end{algorithm}

Initializing the albedo from a flat-lit, diffuse-only baseline is critical for downstream parameter decoupling. Arbitrary or global-mean initialization risk optimization collapse into local minima due to facial features, hair, or micro-shadow contamination. If specular highlights contaminate the initial albedo estimate, the optimization struggles to isolate independent light transport components. Securing a clean, purely diffuse baseline guarantees that high-frequency specular details are absorbed into the residual specular map rather than getting baked into the estimated albedo or roughness maps.

We perform our initialization strategy in two stages (Algorithm~\ref{alg:albedo_initialization}):

\clearpage

\begin{enumerate}
    \item \textbf{Adaptive Skin Masking:} The input is converted to CIE $L^*a^*b^*$ space. Chromatic channels ($a^*, b^*$) are clustered via $k$-means ($k=2$) after filtering low-luminance background pixels ($L^* \le 20$). The skin cluster is identified by the maximum $a^*$ centroid. High-frequency specularity and deep shadow artifacts are eliminated by removing the top $10\%$ and bottom $5\%$ luminance percentiles within the raw mask.
    \item \textbf{Statistical Sigma Clipping:} Active skin pixels are mapped back to linear RGB space. For each spectral channel, a $2\sigma$ Sigma Clipping filter eliminates remaining outliers (e.g., edge pixels, macro-shadow boundary artifacts) from the distribution. The robust mean of the clipped pixels forms the final initialization vector $\mathbf{c}_{\text{init}}$.
\end{enumerate}

\subsection{GPU Shader Implementation}
\label{subsec:shader_impl}
Algorithm~\ref{alg:burley_shader} details our screen-space multiple scattering approximation within the fragment shader, mapping directly to the mathematical formulations described in the main manuscript. All equation references within the pseudocode correspond to the main text.

\subsection{Burley Multiple Scattering Differentiable Tensorized Implementation}
\label{subsec:tensorized_impl}
While forward rendering executes via the GPU fragment shader, inverse optimization requires an auto-differentiable SSSS model to backpropagate loss gradients. We translate the screen-space diffusion pass into vectorized PyTorch operations by broadcasting inputs across the $N$ importance sample coordinates into a unified high-dimensional tensor. This architecture enables simultaneous 3D view-space reconstruction, projection, and differentiable bilinear texture lookups, removing sequential per-pixel execution loops. The tensorized execution logic is formalized in Algorithm~\ref{alg:burley_tensorized}, where tuples within the inline comments denote multi-dimensional tensor shapes, and equation references correspond to the formulations established in the main text.